\documentclass[jair,twoside,11pt,theapa]{article}
\usepackage{jair, theapa, rawfonts}

\usepackage[utf8]{inputenc}
\usepackage{url}
\usepackage{amsmath,amssymb,amsthm}
\usepackage{csquotes}
\usepackage{subcaption}
\usepackage{cases} 
\usepackage[inline]{enumitem} 

\usepackage[vlined,algoruled,rightnl,nofillcomment]{algorithm2e}
\usepackage{graphicx}
\usepackage{verbatim}
\graphicspath{{./figures/}}

\usepackage{mathtools}
\usepackage[nounderscore]{syntax} 

\usepackage{tikz}
\definecolor{darkgreen}{RGB}{68,180,46}
\definecolor{darkgray}{RGB}{80,80,80}
 
\usepackage{tabularx}
\usepackage{multirow}

 \newcommand{\rev}[2]{{\color{red}#1}{\color{blue}#2}} 
 \renewcommand{\rev}[2]{{\color{blue}#2}}            
 \renewcommand{\rev}[2]{#2}                          
\newcommand\nop[1]{}        
\usepackage[capitalize]{cleveref}
\crefname{figure}{Figure}{Figures}
\Crefname{figure}{Figure}{Figures}
\crefname{equation}{}{}
\Crefname{equation}{}{}

\newtagform{angles}[\textit]{$\langle$}{$\rangle$}
\usetagform{angles}
\newcommand{\eqrefformat}[1]{$\langle$\textit{#1}$\rangle$}
\creflabelformat{equation}{#2\eqrefformat{#1}#3}

\newcommand{\mi}[1]{\ensuremath{\mathit{#1}}}

\newcommand{\slv}[1]{\textsc{#1}} 
\newcommand{\alphaslv}{\slv{Alpha}}

\newcommand{\func}[1]{\ensuremath{\mathsf{#1}}}

\newcommand{\variablesymbols}{\ensuremath{\mathcal{V}}}
\newcommand{\constantsymbols}{\ensuremath{\mathcal{C}}}
\newcommand{\functionsymbols}{\ensuremath{\mathcal{F}}}
\newcommand{\predicatesymbols}{\ensuremath{\mathcal{P}}}
\newcommand{\folanguage}{\ensuremath{\langle \variablesymbols, \constantsymbols, \functionsymbols, \predicatesymbols \rangle}}
\newcommand{\prog}{\ensuremath{P}}
\newcommand{\universe}{\ensuremath{U}}
\newcommand{\base}{\ensuremath{B}}
\newcommand{\interpretation}{\ensuremath{I}}
\newcommand{\assignment}{\ensuremath{A}}
\newcommand{\assignedtruth}{\func{truth}}
\newcommand{\assignedtruthof}[2][\assignment]{\assignedtruth_{#1}(#2)}
\newcommand{\prop}{\func{prop}}
\newcommand{\nafsymbol}{\ensuremath{not}}
\newcommand{\naf}[1]{\ensuremath{\mathrm{\nafsymbol}~#1}}
\newcommand{\head}{\func{head}}
\newcommand{\body}{\func{body}}
\newcommand{\bodyp}{\func{body^+}}
\newcommand{\bodyn}{\func{body^-}}
\newcommand{\ground}{\ensuremath{\func{grd}}}
\newcommand{\sigT}{\ensuremath{\mathbf{T}}}
\newcommand{\sigF}{\ensuremath{\mathbf{F}}}
\newcommand{\sigM}{\ensuremath{\mathbf{M}}}
\newcommand{\sigU}{\ensuremath{\mathbf{U}}}
\newcommand{\ucap}{\ensuremath{\mathit{UCAP}}}
\newcommand{\iucap}{\ensuremath{\mathit{IUCAP}}}
\newcommand{\condsymbol}{\ensuremath{\func{cond}}}
\newcommand{\cond}[1]{\ensuremath{\func{\condsymbol}(#1)}}
\newcommand{\condpos}[1]{\ensuremath{\func{\condsymbol^+}(#1)}}
\newcommand{\condneg}[1]{\ensuremath{\func{\condsymbol^-}(#1)}}
\newcommand{\heudirstmt}{\mathtt{\#heuristic}}
\newcommand{\sumagg}{\mathtt{\#sum}}
\newcommand{\heusign}{\ensuremath{s}}
\newcommand{\heuat}{\ensuremath{\mi{ha}}}
\newcommand{\heuhead}{\ensuremath{\mi{ha}}}
\newcommand{\heubody}{\ensuremath{\mi{hB}}}
\newcommand{\fheuat}{\ensuremath{\func{at\rev{}{o}m}}}
\newcommand{\fsignset}{\ensuremath{\func{signs}}}
\newcommand{\heusignT}{\sigT}
\newcommand{\heusignF}{\sigF}
\newcommand{\heusignM}{\sigM}
\newcommand{\heusignAny}{\ensuremath{\heusignF\heusignM\heusignT}}
\newcommand{\heusignFT}{\ensuremath{\heusignF\heusignT}}
\newcommand{\heusignFM}{\ensuremath{\heusignF\heusignM}}
\newcommand{\heudir}{\ensuremath{d}}
\newcommand{\heudirs}{\ensuremath{D}}

\newcommand{\ChoiceOn}{\ensuremath{\mathrm{ChoiceOn}}}
\newcommand{\ChoiceOff}{\ensuremath{\mathrm{ChoiceOff}}}
\newcommand{\HeuOnrm}[1]{\ensuremath{\mathrm{HeuOn#1}}}
\newcommand{\HeuOffrm}[1]{\ensuremath{\mathrm{HeuOff#1}}}

\newcommand{\filterbysigns}[2]{\ensuremath{#1|_{#2}}}
\newcommand{\vars}{\ensuremath{\func{vars}}}

\newcommand{\substitution}{\ensuremath{\sigma}}
\newcommand{\choiceatom}[2]{\ensuremath{\beta(#1,#2)}}
\newcommand{\lb}{\mi{lb}}
\newcommand{\ub}{\mi{ub}}
\newcommand{\splitpos}{\func{split}^+}
\newcommand{\splitneg}{\func{split}^-}
\newcommand{\splittable}{\mi{s}}
\newcommand{\Splittable}{\mathrm{S_{split}}}
\newcommand{\modcond}{\func{heuristic}}
\newcommand{\transform}[1]{\func{tf{#1}}}

\newcommand{\astarproblem}{\ensuremath{\mathit{P}}}
\newcommand{\astarfrontier}{\ensuremath{\mathit{F}}}
\newcommand{\astarexplored}{\ensuremath{\mathit{E}}}

\newtheorem{definition}{Definition}
\newtheorem{example}{Example}
\newtheorem{theorem}{Theorem}
\newtheorem{lemma}{Lemma}

\jairheading{TBD}{TBD}{TBD}{TBD}{TBD}

\ShortHeadings{Domain-Specific Heuristics in Answer Set Programming}
{Comploi-Taupe, Friedrich, Schekotihin, \& Weinzierl}
\firstpageno{1}

\begin{document}

\title{Specifying and Exploiting Non-Monotonic Domain-Specific Declarative Heuristics  in Answer Set Programming\thanks{This work significantly extends a conference paper by \fullciteA{DBLP:journals/corr/abs-1909-08231} by more examples, rethought language syntax, new encodings, a more detailed evaluation, and an extensive study of state-space search with A* implemented in the suggested approach.}}

\author{%
	\name Richard Comploi-Taupe\thanks{Richard \rev{}{Comploi-}Taupe \rev{}{(formerly known as Richard Taupe)} is the main author of this paper; other authors contributed equally and are listed in the alphabetical order.}
		\email richard.taupe@siemens.com, rtaupe@edu.aau.at \\
		\addr Siemens AG Österreich and Alpen-Adria-Universität Klagenfurt, Austria
	\AND
	\name Gerhard Friedrich
		\email gerhard.friedrich@aau.at \\
		\addr Alpen-Adria-Universität Klagenfurt, Austria
	\AND
		\name Konstantin Schekotihin
		\email konstantin.schekotihin@aau.at \\
		\addr Alpen-Adria-Universität Klagenfurt, Austria
	\AND
		\name Antonius Weinzierl
		\email weinzierl@kr.tuwien.ac.at \\
		\addr TU Wien, Vienna, Austria
}

\maketitle

\begin{abstract}
	
	Domain-specific heuristics are an essential technique for solving combinatorial problems efficiently.
	Current approaches to integrate domain-specific heuristics with Answer Set Programming (ASP) are unsatisfactory when dealing with heuristics that are specified non-monotonically on the basis of partial assignments.
	Such heuristics frequently occur in practice,
	for example, when picking an item that has not yet been placed in bin packing.
	Therefore, we present novel syntax and semantics for declarative specifications of domain-specific heuristics in ASP.
	Our approach supports heuristic statements that depend on the partial assignment maintained during solving, which has not been possible before.
	We provide an implementation in \alphaslv\ that makes \alphaslv\ the first lazy-grounding ASP system to support declaratively specified domain-specific heuristics.
	\rev{Three }{Two} practical example domains are used to demonstrate the benefits of our proposal.
	\rev{One of these domains is general}{Additionally, we use our approach to implement informed}  search with A*\!, which is tackled \rev{with}{within} ASP for the first time.
	\rev{}{A* is applied to two further search problems.}
	The experiments confirm that combining lazy-grounding ASP solving and our novel heuristics can be vital for solving industrial-size problems.
\end{abstract}

\section{Introduction}
Answer Set Programming
(ASP) \shortcite{aspbook-baral,Gebser.2012,aspbook-gelfond,aspbook-lifschitz}
is a declarative knowledge representation formalism
that has been applied successfully in a variety of industrial and scientific applications
\shortcite{DBLP:journals/aim/ErdemGL16,DBLP:journals/ki/FalknerFSTT18}\rev{.}{\ such as configuration \shortcite{DBLP:journals/aim/FalknerFHSS16}, team building \shortcite{DBLP:journals/tplp/RiccaGAMLIL12}, routing \shortcite{DBLP:journals/tplp/GebserOSR18}, or scheduling \shortcite{DBLP:conf/ruleml/DodaroGKMP19}.}

The \emph{ground-and-solve} approach is the predominant method employed by state-of-the-art ASP solvers.
In ground-and-solve, solvers first instantiate the given non-ground program and then apply various strategies to find answer sets of the obtained ground program \shortcite{DBLP:conf/ijcai/GebserLMPRS18}.
Ground-and-solve is applied by systems
such as \slv{clingo} \shortcite{DBLP:journals/tplp/GebserKKS19}, \slv{dlv} \shortcite{DBLP:journals/ki/AdrianACCDFFLMP18,DBLP:journals/tocl/LeonePFEGPS06}, or \slv{dlv2} \shortcite{DBLP:conf/lpnmr/AlvianoCDFLPRVZ17}.

However, modern applications manifested two issues with the ground-and-solve approach.
The first issue is the so-called \emph{grounding bottleneck}: Large problem instances in industrial applications often cannot be grounded by modern grounders like \slv{gringo} \shortcite{DBLP:conf/lpnmr/GebserKKS11} or \slv{I-DLV} \shortcite{DBLP:journals/ia/CalimeriFPZ17} in acceptable time and space \shortcite{DBLP:journals/amai/EiterFFW07}.
The second issue is that, even if the problem can be grounded, computation of answer sets might take considerable time, as indicated by \rev{the}{} ASP Competition\rev{s}{} \rev{}{reports} \shortcite{DBLP:journals/ai/CalimeriGMR16,DBLP:journals/jair/GebserMR17,aspcomp2017}.

\rev{}{Both issues were recently adressed by many researchers.} 
First, to overcome the grounding bottleneck, lazy-grounding ASP systems interleave grounding and solving to instantiate and store only relevant parts of the ground program in memory.
Lazy grounding is implemented by systems such as \slv{gasp} \shortcite{DBLP:journals/fuin/PaluDPR09}, \slv{ASPeRiX} \shortcite{DBLP:journals/tplp/LefevreBSG17}, \slv{OMiGA} \shortcite{DBLP:conf/jelia/Dao-TranEFWW12}, and \alphaslv\ \shortcite{DBLP:conf/lpnmr/Weinzierl17}.
\rev{The runtime performance issue is tackled by modern solvers using various techniques, including domain-specific heuristics.
Seamless, declarative integration of heuristics into
ASP encodings has been proposed by \shortciteA{DBLP:conf/aaai/GebserKROSW13}. On the other hand, procedural heuristics have been proposed by \shortciteA{DBLP:journals/tplp/DodaroGLMRS16}. This procedural approach allows for heuristics that interact directly with the internal decision-making procedures.}
{The second performance-related issue is tackled by modern solvers using various techniques among which domain-specific heuristics play a central role. Examples of such approaches include both declarative and procedural approaches.
A seamless and declarative integration of heuristics into ASP encodings has been proposed by \shortciteA{DBLP:conf/aaai/GebserKROSW13}, whereas an approach using procedural heuristics that interact directly with the internal decision-making procedures is discussed by \shortciteA{DBLP:journals/tplp/DodaroGLMRS16}.}

\rev{Dynamically evaluating heuristics w.r.t.\ a partial assignment can be an essential feature for practical domains.
For example, heuristics in product configuration or for bin packing may need to compute the amount of space left after placing a component or an item.
So far, such dynamic reasoning has been supported by the existing procedural approach, but not by the declarative one.

Both existing approaches to integrate domain-specific heuristics with ASP solving are unsatisfactory:
Procedural heuristics counteract the declarative nature of ASP, and the existing declarative approach does not permit dynamic heuristics reasoning about partial assignments.
Moreover, previous research has not addressed declarative heuristics for the lazy-grounding case.
}
{
However, there is no ASP system that addresses both issues simultaneously. That is, previous research has not addressed declarative heuristics for the lazy-grounding case.
In addition, the existing integrations of domain-specific heuristics with ASP solving are unsatisfactory:
Procedural heuristics counteract the declarative nature of ASP, and the existing declarative approach makes it quite tedious and counter-intuitive to model \emph{dynamic heuristics} for reasoning about partial assignments, which is an essential feature for practical applications.
For example, heuristics in product configuration or for bin packing may need to compute the amount of space left after placing a component or an item.
So far, such dynamic reasoning has been supported by the existing procedural approach, but not by the declarative one.
}

We tackle the challenge of finding a satisfying solution \rev{to these issues. }{addressing both the grounding bottleneck and the runtime performance issue.}
To this end, we extend the existing declarative approach by \shortciteA{DBLP:conf/aaai/GebserKROSW13} \rev{. Our extension supports}{with} dynamic heuristics while at the same time keeping the language simple and easy to use.
\rev{We}{Second, we} integrate our approach into a lazy-grounding system, which requires non-trivial adaptations due to the different solving mechanisms in effect.

\subsection{Contributions}

In this work, we present a novel approach to dynamic declarative domain-specific heuristics for ASP. We combine this approach with lazy grounding to facilitate the solving of large and complex problems.
In summary, our work provides the following contributions:
\begin{itemize}
	\item \rev{we }{We} present novel semantics that makes declarative specifications of domain-specific heuristics more intuitive by allowing heuristics to depend non-monotonically on the partial assignment maintained during solving (i.e., heuristics may be applicable in one partial assignment and cease to be applicable at a later stage)\rev{;}{.}
	\item \rev{we }{We} propose a language for declarative specifications of domain-specific heuristics within answer-set programs that can be seen as a variant of the one introduced by \citeA{DBLP:conf/aaai/GebserKROSW13}, and we formally define the language by an EBNF grammar\rev{;}{.}
	\item \rev{we }{We} show how to integrate our language into a lazy-grounding ASP system and provide a reference implementation within the well-known lazy-grounding system \alphaslv\rev{;}{.}
	\item \rev{we }{We} demonstrate how to use our approach to model domain-specific heuristics for \rev{three }{two} practical example domains: the House Reconfiguration Problem (HRP) \rev{, }{ and} the Partner Units Problem (PUP)\rev{, and }{. Additionally, we use our approach to implement} state-space search with A*\!, which is tackled \rev{with}{within} ASP for the first time.\rev{}{\footnote{\rev{}{\shortciteA{DBLP:conf/epia/CabalarRV19} present an A* algorithm which employs ASP through repeated calls to an ASP solver for computing states and their successors.}}} \rev{}{Integrating A* and ASP allows, on the one hand, the declarative specification of states and their successor states employing ASP and, on the other hand, the application of informed search strategies exploiting heuristic functions. We apply A* to two further search problems.}
	\item \rev{and }{And} finally, we demonstrate how solving performance can profit from our approach by presenting experimental results in these practical domains.
\end{itemize}

\subsection{Organization}

Preliminaries are covered by \cref{sec-preliminaries}, before
\Cref{sec-domspec-soa} outlines \rev{}{the} state of the art in domain-specific heuristics.
On this basis, we introduce our novel semantics in \cref{sec-domspec-novel}.
\Cref{sec-domspec-lazygrounding} explains how to integrate our approach into a lazy-grounding ASP system.
Applications and experimental results are presented in \cref{sec-applications-and-experiments}, and
\Cref{sec-conclusion} concludes the article.

\section{Preliminaries}
\label{sec-preliminaries}

Answer Set Programming (ASP) \shortcite{aspbook-baral,Gebser.2012,aspbook-gelfond,aspbook-lifschitz} is an approach to declarative programming.
Instead of stating how to solve a problem, the programmer formulates the problem as a logic program specifying the search space and the properties of valid solutions.
An ASP solver then finds models (so-called \textit{answer sets}) for this logic program, which correspond to solutions for the original problem.

\subsection{Syntax}
\label{sec-preliminaries-syntax}

ASP offers a rich input language, of which we introduce only the core concepts needed \rev{to formalise the concepts of }{in} this paper.
For a comprehensive definition of ASP's syntax and semantics, we refer to \shortciteA{DBLP:journals/tplp/CalimeriFGIKKLM20}.

Let $\folanguage$ define a first-order language, where $\variablesymbols$ is a set of variable symbols, $\constantsymbols$ is a set of constant symbols, $\functionsymbols$ is a set of function symbols, and $\predicatesymbols$ is a set of predicate symbols.

A classical \emph{atom} is of the form $p(t_1,\dots,t_n)$, where $p \in \predicatesymbols$ is a predicate symbol and $t_1,\dots,t_n$ are \emph{terms}.
Each variable $v \in \variablesymbols$ and each constant $c \in \constantsymbols$ is a term.
Furthermore, for $f \in \functionsymbols$, $f(t_1,\dots,t_n)$ is a function term.

An answer-set program \prog\ is a finite set of (disjunctive) rules of the form
\begin{equation}
\label{eqRule}
h_1 \vee \dots \vee h_k \leftarrow b_1,~\ldots~,b_m,~\naf{b_{m+1}},~\ldots,~\naf{b_n}.
\end{equation}
where $h_1,\dots,h_k$ and $b_1,\dots,b_n$ are atoms and $\mathit{\nafsymbol}$ is negation as failure \rev{}{(a.k.a.\ default negation)}, which refers to the absence of information, i.e., an atom is assumed to be false as long as it is not derived by \rev{another }{some} rule.
A \emph{literal} is either an atom $a$ or its negation $\naf{a}$.
Given a rule $r$ of the form \eqref{eqRule}, $\head(r)=\{ h_1, \dots, h_k \}$ is called the \textit{head} of $r$, and
$\body(r) = \{ b_1,\dots,b_m,$ $\naf{b_{m+1}},$
$\ldots,\naf{b_n} \}$ is called the \emph{body} of $r$.
By $\bodyp(r) = \{ b_1,\dots,b_m \}$ and $\bodyn(r) = \{b_{m+1}, \ldots, b_n\}$ we denote the positive and negative atoms in the body of $r$, respectively.
A rule $r$ where $\head(r) = \emptyset$, e.g., $\leftarrow \mathrm{b}.$, is called \textit{integrity constraint}, or simply \textit{constraint}.
A rule $r$ where $\body(r) = \emptyset$, e.g., $\mathrm{h} \leftarrow.$, is called \textit{fact}. 
A rule is ground if all its atoms are variable-free. A ground program comprises only ground rules.

\subsection{Semantics}
\label{sec-preliminaries-semantics}

Given a program $\prog$, the \emph{Herbrand universe} of $\prog$, denoted by $\universe_\prog$, consists of all \rev{}{integers and of all} ground terms constructible from constant symbols and function symbols appearing in $\prog$.
The \emph{Herbrand base} of $\prog$, denoted by $\base_\prog$, is the set of all ground classical atoms that can be built by combining predicates appearing in $\prog$ with terms from $\universe_\prog$ as arguments \shortcite{DBLP:journals/tplp/CalimeriFGIKKLM20}.

A substitution $\sigma$ is a mapping from variables $\variablesymbols$ to elements of the Herbrand universe $\universe_\prog$ of a program $\prog$.
Let $O$ be a rule, an atom, or a literal, then by $O\sigma$ we denote a \rev{ground }{}rule, atom, or literal obtained by replacing each variable $v \in \vars(O)$ by $\sigma(v)$. The function $\vars$ maps any rule, atom, literal, or any other object containing variables to the set of variables it contains. For instance, $\vars(\mathrm{a(X)}) = \{ \mathrm{X} \}$ and for a rule $r_1: \mathrm{a(X)} \leftarrow \mathrm{b(X,Y)}.$, $\vars(r_1) = \{ \mathrm{X}, \mathrm{Y} \}$.

\rev{}{As usual, we assume rules to be \emph{safe}, which is the case for a rule $r$ if $\vars(r) \subseteq \bigcup_{a \in \bodyp(r)}\vars(a)$.}

The (ground) instantiation \rev{$\ground(r)$}{} of a rule $r$ equals $r\sigma$ for some substitution $\sigma$, which maps all variables in $r$ to ground terms.
The (ground) instantiation $\ground(\prog)$ of a program $\prog$ is the set of all possible instantiations of the rules in $\prog$ \shortcite{DBLP:journals/tplp/CalimeriFGIKKLM20,DBLP:journals/ai/FaberPL11}.
Function symbols may cause the Herbrand base and the full grounding of a program to be infinite \shortcite{alvi-etal-2011-nonmonat30}.
By restricted usage of function symbols, answer-set programs can be designed in a way that reasoning is decidable.

\rev{Every non-ground answer set program $\prog$ is semantically equivalent to its instantiation $\ground(\prog)$ if all its rules are safe.
A rule $r$ is \emph{safe} if $\vars(r) \subseteq \bigcup_{a \in \bodyp(r)}\vars(a)$. Therefore, semantics are defined only for ground programs.}{}

A\rev{}{n} \emph{Herbrand interpretation} for a program $\prog$ is a set of ground classical atoms $\interpretation \subseteq \base_\prog$.
A ground classical atom $a$ is true w.r.t.\ an interpretation $\interpretation$, denoted $\interpretation \models a$, iff $a \in \interpretation$.
A ground literal $\naf{a}$ is true w.r.t.\ an interpretation $\interpretation$, denoted $\interpretation \models \naf{a}$, iff $\interpretation \nvDash a$.
A rule $r$ is \emph{satisfied} w.r.t.\ $\interpretation$, denoted $\interpretation \models r$, if some head atom is true w.r.t.\ $\interpretation$ ($\exists h \in \head(r): \interpretation \models h$) whenever all body literals are true w.r.t.\ $\interpretation$ ($\forall b \in \body(r): \interpretation \models b$).
An interpretation $\interpretation$ is a model of $\prog$, denoted $\interpretation \models \prog$, if $\interpretation \models r$ for all rules $r \in \ground(\prog)$ \shortcite{DBLP:journals/ai/FaberPL11}.

Given a ground program $\prog$ and an interpretation $\interpretation$, let $\prog^\interpretation$ denote the transformed program obtained from $\prog$ by deleting rules in which a body literal is false w.r.t.\ $\interpretation$:
$\prog^\interpretation = \{ r \mid r \in \prog, \forall b \in \body(r): \interpretation \models b \}$ \shortcite{DBLP:journals/ai/FaberPL11}.

An interpretation $\interpretation$ of a program $\prog$ is an \emph{answer set} of $\prog$ if it is a subset-minimal model of $\ground(\prog)^\interpretation$, i.e., $\interpretation$ is a model of $\ground(\prog)^\interpretation$ and there exists no $\interpretation' \subsetneq \interpretation$ that is a model of $\ground(\prog)^\interpretation$ \shortcite{DBLP:journals/ai/FaberPL11}.

\subsection{Notation}
\label{sec-preliminaries-notation}

In this section, we introduce some notation that will be used later in the article.

An \emph{assignment} $\assignment$ over $\base_\prog$ is a set of signed literals $\sigT~a$, $\sigF~a$, or $\sigM~a$, where $\sigT~a$ and $\sigF~a$ express that an atom $a$ is true and false, respectively, and $\sigM~a$ indicates that $a$ \enquote{must-be-true}.
\sigM\ \rev{signifies }{means} that an atom must eventually become true by derivation in a correct solution extending the current partial assignment, but no derivation has yet been found that would make the atom true.
\rev{}{Intuitively, $\sigT~b \in \assignment$ means that $b$ is true and justified, i.e., derived by a rule that fires under $\assignment$, while $\sigM~b \in \assignment$ only indicates that $b$ is true but potentially not derived.}
Let $\assignment_s = \{ a \mid s\ a \in \assignment \}$ for $s \in \{ \sigF, \sigM, \sigT \}$ denote the set of atoms occurring with a specific sign in assignment $\assignment$.
We assume assignments to be \emph{consistent}, i.e., no negative literal may also occur positively ($\assignment_\sigF \cap (\assignment_\sigM \cup \assignment_\sigT) = \emptyset$), and every positive literal must also occur with must-be-true ($\assignment_\sigT \subseteq \assignment_\sigM$).
\rev{}{The latter condition ensures that assignments are monotonically growing (w.r.t.\ set inclusion) in case an atom that was must-be-true becomes justified by a rule deriving it and hence changes to true.}

An assignment $\assignment$ is \emph{complete} if every atom in the Herbrand base is assigned true or false ($\forall a \in \base_\prog: a \in \assignment_\sigF \cup \assignment_\sigT$).
An assignment that is not complete is \emph{partial}.

The function $\assignedtruth_\assignment: \base_\prog \to \{ \sigT, \sigM, \sigF, \sigU \}$ for a (partial) assignment $\assignment$ maps an atom to the truth value that the atom is currently assigned in the given assignment, or to $\sigU$ if the atom is currently unassigned:
\[
	\assignedtruthof{a} = \begin{cases}
		\sigF\ &a \in \assignment_\sigF,	\\
		\sigM\ &a \in \assignment_\sigM \setminus \assignment_\sigT,	\\
		\sigT\ &a \in \assignment_\sigT,	\\
		\sigU\ &\text{otherwise.}
	\end{cases}
\]
\rev{}{For example, if $\assignment = \{ \sigM~\mathrm{a}, \sigT~\mathrm{a}, \sigM~\mathrm{b} \}$, then $\assignment_\sigF = \emptyset$, $\assignment_\sigM = \{ \mathrm{a}, \mathrm{b} \}$, $\assignment_\sigT = \{ \mathrm{a} \}$, $\assignedtruthof{\mathrm{a}} = \sigT$, and $\assignedtruthof{\mathrm{b}} = \sigM$.}

\rev{}{
  Note that historically, the concept of a truth value \enquote{must-be-true} was introduced to ASP solving by the DLV solver for efficiency \shortcite{DBLP:journals/ai/EiterFLPP03,DBLP:conf/lpnmr/FaberLP99}. The same efficiency was later realized without needing \enquote{must-be-true} by use of so-called source pointers, together with unfounded-set checks, and completion \shortcite{DBLP:journals/ai/GebserKS12}. These techniques ensure that in a final assignment $\assignment$, all atoms assigned true are derived by a rule that fires in $\assignment$.
  The distinction between an atom being true or must-be-true, however, is still present in most modern ASP solvers. The only difference to our approach is that other solvers represent must-be-true implicitly via the internal state of source-pointers and the result of unfounded-set propagation, while we chose to represent \enquote{must-be-true} explicitly as a truth value.
  
  Expressing \enquote{must-be-true} explicitly also allows for more fine-grained domain-specific heuristics, as they may now distinguish between whether a goal $g$ has been reached in an assignment $\assignment$ by being assigned true (i.e., $\assignedtruthof{g} = \sigT$) or still must be reached (i.e., $\assignedtruthof{g} = \sigM$). The former case may indicate that a rule checking the goal condition fired and derived $g$, while the latter may arise from a constraint $\leftarrow \naf g.$, which states that the goal must be met in any answer set. Our approach to domain-specific heuristics allows heuristics to distinguish both cases.
}

Many useful language constructs have been introduced to extend the basic language of ASP defined in \cref{sec-preliminaries-syntax,sec-preliminaries-semantics}.
We discuss such extensions only briefly and refer to \shortciteA{DBLP:journals/tplp/CalimeriFGIKKLM20} and \shortciteA{potasscoguide} for full details.

A \textit{cardinality atom} is of the form
\rev{
$
\lb ~ \{a_1:l_{1_1},\dots,l_{1_m};\ldots; a_n:l_{n_1},\dots,l_{n_o}\} ~ \ub
$,
}
{
$
\lb ~ \{a_1:l_{1_1},\dots,l_{m_1};\ldots; a_n:l_{1_n},\dots,l_{m_n}\} ~ \ub
$,
}
where\rev{}{, for $1 \leq i \leq n$,}
\rev{$a_i:l_{i_1},\dots,l_{i_m}$ }{$a_i:l_{1_i},\dots,l_{m_i}$} represent\rev{}{s a} \textit{conditional literal\rev{s}{}} in which $a_i$ (the head of the conditional literal) is a classical atom\rev{,}{} \rev{}{and} all \rev{$l_{i_j}$ }{$l_{j_i}$} are literals, and
$\lb$ and $\ub$ are integer terms indicating a lower and an upper bound, respectively. If one or both of the bounds are not given, their defaults are used, \rev{}{i.e.,} $0$ for $\lb$ and $\infty$ for $\ub$.
A cardinality atom is satisfied if $\lb \leq |C| \leq \ub$ holds, where $C$ is the set of head atoms in the cardinality atom that are satisfied together with their conditions (e.g., \rev{$l_{i_1},\dots,l_{i_m}$ }{$l_{1_i},\dots,l_{m_i}$} for $a_i$).

A \emph{choice rule} is a rule $r$ with $\head(r)$ consisting of a cardinality atom.
For example, consider the choice rule $1 ~\{ \mathrm{a} ; \mathrm{b} \}~ 2 \leftarrow$.
A program consisting only of this rule has three answer sets: $\{ \mathrm{a} \}$, $\{ \mathrm{b} \}$, and $\{ \mathrm{a}, \mathrm{b} \}$.
Subset minimality usually required by answer sets is here circumvented by a rewriting; for details, see \shortciteA{DBLP:journals/tplp/CalimeriFGIKKLM20}.

As an extension of cardinality atoms, ASP also supports aggregate atoms that apply aggregate functions like $\mi{max}$, $\mi{min}$ or $\mi{sum}$ to sets of literals.
An aggregate atom is satisfied if the value computed by the aggregate function respects the given bounds,
e.g., $1 = \sumagg \{ 1 : \mathrm{a}; 2 : \mathrm{b} \}$ is satisfied if $\mathrm{a}$ but not $\mathrm{b}$ is true.

Built-in predicates with a fixed meaning, such as $=, \neq, \leq, \geq, <, >$,
and arithmetic operations such as $+, -, \times, /$, and $\backslash$ (modulo) are also used in infix notation,
and $|t|$ denotes the absolute value of an integer term $t$. 

\subsection{Lazy Grounding}
\label{sec-preliminaries-lazygrounding}

The grounding bottleneck is a well-known issue of traditional ASP solving, where the input program is grounded first before the ground\rev{ed}{}, i.e., variable-free, program is solved.
There have been many attempts to mitigate this issue with more \rev{optimized }{optimised} and clever grounding procedures \shortcite<see, for example,>{DBLP:journals/aim/KaufmannLPS16,DBLP:journals/ia/CalimeriFPZ17,DBLP:conf/lpnmr/GebserKKS11,DBLP:conf/birthday/FaberLP12} or through formulations of answer-sets in formalisms that do not need grounding per se, e.g.\rev{}{,} query-driven evaluation \shortcite<cf.>{Cat2015,DBLP:journals/corr/abs-1709-00501,DBLP:journals/tplp/AriasCSM18} or circumscription\rev{}{s} \cite<cf.>{DBLP:journals/ai/FerrarisLL11}.
\rev{The former approaches usually work well for some input programs, while the latter may result in formalisms that are themselves undecidable.}{}

Lazy grounding is an approach that interleaves the solving and grounding phases, such that computations are guaranteed to yield all answer sets. The foundation for lazy grounding is known as the computation sequence and has been developed by \shortciteA{DBLP:conf/iclp/LiuPST07}. A computation sequence $\mathbf{S} = \langle S_0, S_1, \ldots S_n \rangle$ is a sequence of partial assignments that is monotonically growing (w.r.t.\ set inclusion).
Every element $S_i$ of the sequence represents the state of the computation at step $i$.
The first element of the sequence is empty ($S_0 = \emptyset$), and every other element $S_i$ contains the signed literals that can be derived from the preceding partial assignment $S_{i-1}$ in the program $P$.

Since each element of a computation sequence is a partial assignment containing signed literals, and the sequence is monotonically growing, each $S_i$ contains atoms assigned $\sigT$ that will remain true in all extensions of $S_i$, and atoms assigned $\sigF$ that will definitely remain false in all extensions of $S_i$.

A rule $r$ is said to be \emph{applicable} in $S_i$ if $\{ \sigT~a \mid a \in \bodyp(r) \} \subseteq S_i$ and $\{ \sigM~a \mid a \in \bodyn(r) \} \cap S_i = \emptyset$, i.e., if the positive body is satisfied and $S_i$ does not contradict the negative body.
An applicable rule is said to \emph{fire} when its negative body is chosen to be false by the solver's search procedure.
Firing a rule allows deriving its head.
For each applicable rule the computation sequence may split in two, one $S_{i+1}$ where the rule is assumed to fire and one $S_{i+1}'$  where the rule does not fire. This property allows capturing the guessing mechanism of ASP.

Based on the fact that the computation sequence only needs to know those ground rules that are applicable, lazy-grounding ASP solvers can ground a rule lazily whenever the ground instance becomes applicable. Thus, only those rules are grounded, whose positive body holds in the current partial assignment.

The first lazy-grounding ASP solvers based on the computation sequence principle were \slv{gasp} by \shortciteA{DBLP:journals/fuin/PaluDPR09} and \slv{ASPeRiX} by \shortciteA{DBLP:conf/lpnmr/LefevreN09,DBLP:journals/tplp/LefevreBSG17}. Later \shortciteA{DBLP:conf/jelia/Dao-TranEFWW12} built the \slv{OMiGA} solver that uses a RETE network for efficient grounding. Unfortunately, all these solvers suffered from a lack of efficient solving techniques widely used in traditional ground-and-solve systems.

\rev{}{Efficient ground-and-solve systems for ASP usually employ a technique called conflict-driven clause-learning (CDCL) or a closely related variant called conflict-driven nogood-learning (CDNL). This is employed in most SAT solvers \shortcite{cdcl,cdcl2} and modern ASP solvers like \slv{clingo} \shortcite{DBLP:journals/ai/GebserKS12} or \slv{dlv2} \shortcite{DBLP:conf/lpnmr/AlvianoCDFLPRVZ17}. As CDCL works on clauses, a given ASP program is first transformed into an equivalent set of clauses. Then the models of these clauses are computed using a DPLL-style algorithm, which at its core gains insight into the given problem by learning new clauses from conflicts that are encountered during search. Note that CDNL only differs from CDCL by using the dual form of clauses, which are called nogoods. More details about CDNL and nogood representations are given in \cref{sec-domspec-lazygrounding}, where the implementation of our domain-specific heuristics is presented.}

The \alphaslv\ system by \shortciteA{DBLP:conf/lpnmr/Weinzierl17} \rev{}{has} combined the most important technique for efficient solving, \rev{conflict-driven nogood learning}{CDNL}, with lazy grounding. It is the most recent lazy-grounding ASP system available.

\rev{Note that computation }{Computation} sequences require a normal logic program as input (i.e., rules of the form \eqref{eqRule} without disjunction, cardinality atoms, and aggregate atoms\rev{}{, cf.\ \shortciteA{DBLP:conf/iclp/LiuPST07,DBLP:journals/tplp/LefevreBSG17,DBLP:conf/lpnmr/Weinzierl17}}). Hence lazy-grounding systems usually only accept normal logic programs or, in the case of \alphaslv, rewrite enhanced ASP constructs like aggregates or choice rules into normal rules.
Disjunctive heads can only be rewritten if they are head-cycle-free \cite{DBLP:journals/amai/Ben-EliyahuD94}.
Therefore, in our examples, each rule contains at most one head atom.
\rev{}{Although this appears to be rather limiting, \alphaslv{} is close to supporting the full range of the ASP-Core-2 language through program transformations, including aggregates, choice rules, and integer arithmetics, which are very useful constructs in practical applications.
  In many cases, program transformations result in first-order encodings, which a lazy-grounding system only needs to instantiate for those ground instances actually encountered during search. So these transformations enable significant performance benefits in some cases \cite{Bomanson2019}.

  Furthermore, restricting the input to head-cycle-free programs is not really limiting, as programs containing disjunction with head-cycles are usually employed only when problems on the second level of the polynomial hierarchy, i.e.,~in $\mathbf{\Sigma^P_2}$, are to be solved. The worst-case complexity of normal logic programs without disjunction but with variables, however, is already well beyond $\mathbf{\Sigma^P_2}$ and $\mathbf{PSPACE}$, in fact it is at least $\mathbf{NEXPTIME}$-hard in the worst-case as shown by \citeA{DBLP:conf/coco/DantsinEGV97}. Hence, problems in $\mathbf{\Sigma^P_2}$ can be tackled with \alphaslv{}.}

\section{State of the Art in Domain-Specific Heuristics}
\label{sec-domspec-soa}

State-of-the-art ASP solvers are well suited to solve a wide range of problems, as shown in ASP competitions, experiments, and (industrial) applications reported in the literature \cite{aspcomp2017,DBLP:journals/aim/ErdemGL16,DBLP:journals/ki/FalknerFSTT18,DBLP:conf/rweb/LeoneR15}.
However, applying general ASP solvers to large instances of industrial problems often requires sophisticated encodings or solver tuning methods, e.g., portfolio solvers like \slv{claspfolio} \shortcite{DBLP:journals/tplp/HoosLS14} or \slv{me-asp} \shortcite{DBLP:conf/aiia/MarateaPR15}, to achieve satisfactory performance.

Depending on the problem and instances, tuning of search parameters and encodings \rev{is not }{may not be} sufficient to meet runtime requirements.
Domain-specific heuristics \rev{are }{were} needed to achieve breakthroughs in solving industrial configuration problems with ASP.
Several approaches have implemented embedding heuristic knowledge into the ASP solving process.

\slv{hwasp} \cite{DBLP:journals/tplp/DodaroGLMRS16} \rev{facilitates }{extends \slv{wasp} \shortcite{DBLP:conf/lpnmr/AlvianoADLMR19} by facilitating} external procedural heuristics that are consulted at specific points during the solving process via an API. As a result, \slv{hwasp} can find solutions for all published instances of the Partner Units Problem (PUP) by exploiting external heuristics formulated in C++. 

A declarative approach to formulating domain-specific heuristics in ASP was suggested by \citeA{DBLP:conf/aaai/GebserKROSW13}. The \slv{clingo} system supports $\heudirstmt$ directives described in detail by \citeA{potasscoguide}.
\rev{Heuristics extend the ASP language to allow for declarative specification of atom weights and signs for the solver's internal heuristics. }{%
Heuristic directives extend the ASP language to enable declarative specification of weights determining atom and sign orders in the corresponding internal decision heuristics of a solver.%
} %
An atom's weight influences the order in which atoms are considered by the solver when making a decision. A sign modifier instructs whether the selected atom must be assigned true or false. Atoms with a higher weight are assigned a value before atoms with a lower weight.

The following (non-ground) meta-statement defines domain-specific heuristics in \slv{clingo}, where $\heuhead$ is an atom, $\heubody$ is a conjunction of literals representing the heuristic body, and $w$, $p$, and $m$ are terms \cite{potasscoguide}.
\begin{align}
	\label{eq-clingo-heuristic-directive}
	&\heudirstmt ~~ \heuhead : \heubody. \qquad [w@p,m]&
\end{align}
The optional term $p$ gives a preference between heuristic values for the same atom (preferring those with higher $p$).
The term $m$ specifies the type of heuristic information and can take the following values: \texttt{sign}, \texttt{level}, \texttt{true}, \texttt{false}, \texttt{init} and \texttt{factor}.
For instance, heuristics for $m{=}$\texttt{init} and $m{=}$\texttt{factor} allow modifying initial and actual atom scores evaluated by the solver's decision heuristics (e.g., VSIDS).
The $m{=}$\texttt{sign} modifier forces the decision heuristics to assign an atom $\heuhead$ a required sign, i.e., $\sigT$ or $\sigF$, and $m{=}$\texttt{level} allows for the definition of an order in which the atoms are assigned---the larger the value of $w$, the earlier an atom must be assigned. 
Finally, $m{=}$\texttt{true} specifies that $a$ should be mapped to $\sigT$, i.e., $\assignedtruthof{\heuhead} = \sigT$, with weight $w$ if $\heubody$ is satisfied, $m{=}$\texttt{false} is the analogue heuristics that maps $a$ to $\sigF$.

\rev{Apparently, \citeA{DBLP:conf/aaai/GebserKROSW13,potasscoguide} do not provide a formal specification of which conditions must be fulfilled for a heuristic to take effect. However, the following observation can be made empirically }{
	\citeA{DBLP:conf/aaai/GebserKROSW13} provide a formal specification of a heuristic predicate and the effects of heuristic atoms on the selection of unassigned atoms and their truth assignment in the search process.
	\citeA{potasscoguide} present syntax and semantics for heuristic directives.
	However, the implementation of heuristic directives in \slv{clingo} (v.\ 5.3) might result in a counter-intuitive interpretation of the default negation in heuristic directives since  
}%
for an assignment $\assignment$, a heuristic statement $h$ of the form \cref{eq-clingo-heuristic-directive} is applicable and affects solving iff $\body(h) = \{ b_1,~\ldots~,b_m,~\naf{b_{m+1}},~\ldots,~\naf{b_n} \}$ is assigned true, i.e., $\assignedtruthof{b_i} = \sigT$ for $1 \leq i \leq m$ and $\assignedtruthof{b_j} = \sigF$ for $m+1 \leq j \leq n$.
\rev{Note, }{That is,} the literal $\naf b_i$ is not evaluated to true in the absence of a truth assignment to $b_i$ (as one would expect for default negation). Instead, the literal is evaluated to a truth-value iff $b_i$ is assigned to a truth-value. We illustrate this issue in the following example: 

\begin{example}
	\label{sec-domspec-soa-ex1}
	Consider the following program containing two heuristic directives:
	\begin{align*}
		&\{ ~ \mathrm{a(2)~;~a(4)~;~a(6)~;~a(8)~;~a(5)} ~ \} \leftarrow . \\ &\leftarrow \sumagg ~ \{ ~ \mathrm{X} : \mathrm{a(X)} ~ \} = \mathrm{S}, ~~ \mathrm{S} \backslash 2 \neq 0.
		&\\
		&\heudirstmt ~~ \mathrm{a(5)}.   &[1, \mathrm{true}] \\
		&\heudirstmt ~~ \mathrm{a(4)} ~~ : ~~ \naf{\mathrm{a(5)}}. &[2, \mathrm{true}] \end{align*} The program guesses a subset of $\{ 2, 4, 6, 8, 5 \}$, the sum of which must be even, i.e., $\mathrm{a(5)}$ must not be chosen.
	The heuristic statements specify
	that $\mathrm{a(5)}$ shall be set to true with weight 1, and that $\mathrm{a(4)}$ shall be set to true if $\naf{\mathrm{a(5)}}$ is true with weight 2.
\end{example}

In solving the program in the preceding example, \slv{clingo} (v.\ 5.3) first assigns $\mathrm{a(5)}$ to true in our experiments, although $\mathrm{a(4)}$ has a higher weight and $\mathrm{a(5)}$ is not known to be true in the beginning.
Next,
$\mathrm{a(8)}$ is chosen to be false,\footnote{Choices not determined by the heuristic directives may vary from one implementation to another.} the solver backtracks and only $\mathrm{a(5)}$ stays assigned.
Finally, $\mathrm{a(8)}$ is chosen to be true, and a conflict is learned that makes $\mathrm{a(5)}$ false after backtracking.
Now that $\naf{\mathrm{a(5)}}$ is satisfied, the second heuristic chooses $\mathrm{a(4)}$ to be true, and we obtain the answer set $\{ \mathrm{a(4), a(8)} \}$ after a few more guesses on the yet unassigned atoms.
The second heuristic becomes active only late\rev{}{r} because $\naf{\mathrm{a(5)}}$ is evaluated as true only if $\mathrm{a(5)}$ is assigned false.

\rev{\slv{clingo}'s semantics for heuristic directives seems to introduce some limitations for formulating heuristics that require to reason about the absence of truth-values for atoms, e.g. the absence of decisions in a search state. For example, in configuring technical systems, we might prefer to assign, in the current search state, the most relevant yet unplaced electronic component to a free slot of a motherboard. In scheduling, we might prefer the assignment of lots to machines that are not assigned in the current search state and whose delivery deadline is most urgent. }{Heuristics evaluated over partial assignments appear quite often in practice. For example, in configuring technical systems, we might prefer to assign, in the current search state, the most relevant yet unplaced electronic component to a free slot of a motherboard. In scheduling, we might prefer the assignment of lots to machines that are not assigned in the current search state and whose delivery deadline is most urgent. \slv{clingo}'s semantics for heuristic directives allows one to model heuristics over partial assignments as in \cref{sec-domspec-soa-ex1} using priorities. However, modelling of prioritized directives can be tedious and an automatised translation might require introducing unnecessary auxiliary variables.}

To overcome this issue
we propose, in the following section, to evaluate negation as failure (i.e., $\nafsymbol$) in heuristic statements \emph{w.r.t.\ the current partial assignment} in the solver. This partial assignment represents the search state. 
As a consequence, $\naf{X}$ is true if $X$ is false \emph{or unassigned} during the search. The following example shows the application of default negation to formulate a well known-heuristic for pathfinding.

\begin{example}
	\label{ex-pathfinding-clingo}
	Let \emph{Pathfinding} be the problem of finding a path from one square to another on a rectangular grid by moving horizontally and vertically and avoiding obstacles.
	For example, in \cref{fig-pathfinding}, there are two shortest paths from the start (S) to the goal (G), both of length 4.
	\begin{figure}
		\centering
		\begin{tikzpicture}[scale=0.55,node distance=0.4cm,>=latex]
	
	\foreach \i in {0,...,5} 
	{
		\draw [-] (0,\i) -- (6,\i);
	}

	\foreach \i in {0,...,6} 
	{
		\draw [-] (\i,0) -- (\i,5);
	}
	
	\foreach \i in {0,...,5} 
	{
		\node[anchor=west] at (\i+.1,-.5) {\i} ;
	}

	\foreach \i in {0,...,4} 
	{
		\node[anchor=south] at (-.5,\i) {\i} ;
	}

	\node[anchor=center] at (4.5,2.5) {S};
	\node[anchor=center] at (1.5,3.5) {G};
	
	\path[draw, fill=black] (2,1) rectangle (3,2) ; 
	\path[draw, fill=black] (3,3) rectangle (4,4) ; 
\end{tikzpicture}
		\caption{A sample Pathfinding instance}
		\label{fig-pathfinding}
	\end{figure}
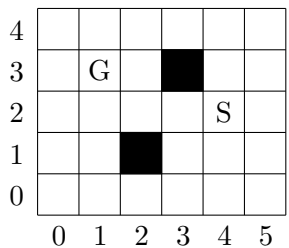
	
	This instance is specified by the following self-explanatory facts:
	\[
	\begin{array}{llll}
		\mathrm{xmin}(0).& \mathrm{ymin}(0).& \mathrm{start}(4,2).& \mathrm{obstacle}(3,3).\\ \mathrm{xmax}(5).& \mathrm{ymax}(4).& \mathrm{goal}(1,3).& \mathrm{obstacle}(2,1).
	\end{array}
	\]
	
	Assume adequate definitions of the following predicates:
	\rev{$\mathrm{at}(X,Y,T)$ means that the agent's position is $(X,Y)$ at time $T$; $\mathrm{neighbour}(\mi{X1},\mi{Y1},\mi{X2},\mi{Y2})$ signifies that the agent can move from $(\mi{X1},\mi{Y1})$ to $(\mi{X2},\mi{Y2})$ and vice versa; and $\mathrm{max\_time}(\mi{MaxT})$ gives the time horizon. }{
	\begin{itemize}
		\item $\mathrm{at(X,Y,T)}$ means the agent's position is $(X,Y)$ at time $T$;
		\item $\mathrm{neighbour}(\mathrm{X1}$, $\mathrm{Y1}$, $\mathrm{X2}$, $\mathrm{Y2})$ signifies that the agent can move from $(\mi{X1},\mi{Y1})$ to $(\mi{X2},\mi{Y2})$ and vice versa;
		\item and $\mathrm{max\_time}(\mathrm{MaxT})$ gives the time horizon.
	\end{itemize}
	}

	Then\rev{}{,} the following program encodes the problem:
	\[\begin{array}{rcl}
		\mathrm{at(X,Y,0)} &\leftarrow &\mathrm{start(X,Y)}. \\
		\mathrm{reached\_goal(T)} &\leftarrow &\mathrm{goal(X,Y)}, \mathrm{at(X,Y,T)}. \\
		\multicolumn{3}{l}{1\;\{\;\mathrm{move(X,Y,T)}\;:\;\mathrm{neighbour}(\mathrm{AtX},\mathrm{AtY},\mathrm{X},\mathrm{Y}), \naf{\mathrm{obstacle(X,Y)}} \;\}\;1}\\
			&\leftarrow &\mathrm{at}(\mathrm{AtX},\mathrm{AtY},\mathrm{T}), \naf{\mathrm{reached\_goal(T)}},\\
			&&\mathrm{max\_time}(\mathrm{MaxT}), \mathrm{T} < \mathrm{MaxT}.\\
		\mathrm{at(X,Y,T+1)}  &\leftarrow &\mathrm{move(X,Y,T)}.\\
		\mathrm{moved(T)}  &\leftarrow &\mathrm{move(X,Y,T)}. \\
		\mathrm{visited(X,Y)}  &\leftarrow &\mathrm{at(X,Y,T)}.
	\end{array}\]
	
	A heuristic preferring to move to squares with the least Manhattan distance to the goal \cite<\enquote{greedy best-first search}, cf.>{aima,DBLP:books/daglib/0068933} can be encoded by the following heuristic directive, following \slv{clingo}'s syntax given in \cref{eq-clingo-heuristic-directive}:
	\begin{align*}
		\heudirstmt ~~ \mathrm{move(X,Y,T)} ~:~ &\mathrm{at(AtX,AtY,T)}, \mathrm{neighbour(AtX,AtY,X,Y)},\\
		&\naf{\mathrm{obstacle(X,Y)}}, \mathrm{goal(GoalX,GoalY)}, \mathrm{max\_time(MaxT)}.\\ &[\mathrm{MaxT} - (|\mathrm{X}-\mathrm{GoalX}| + |\mathrm{Y}-\mathrm{GoalY}|),\mathrm{true}] \end{align*}
	
	But how to restrict suggestions by this heuristic to move only at time $T$ if the agent has not already moved at time $T$, or to move only to squares not already visited?
	We would expect to achieve such a restriction by adding one or both of the following literals to the heuristic's body:
	\[
	\begin{array}{cc}
		\naf \mathrm{moved(T)},&\naf \mathrm{visited(X,Y)}
	\end{array}
	\]
	However, the heuristic is rendered unusable if either of the two literals is added.
	The reason is that \slv{clingo} does not interpret $\mathrm{\nafsymbol}$ within bodies of heuristics w.r.t.\ the current partial assignment, and thus the heuristic is not applicable unless all atoms within negative literals are assigned false.
\end{example}

\section{A Novel Semantics for Declarative Domain-Specific Heuristics}
\label{sec-domspec-novel}

Declaratively specifying domain-specific heuristics in ASP plays a vital role in enabling ASP to solve large-scale industrial problems.
\slv{clingo} has been the only ASP system to support such heuristics so far.
Although language and semantics of heuristic directives in \slv{clingo} have shown to be beneficial in many cases, dynamic aspects of negation as failure in heuristic conditions have not been addressed satisfactorily.

We present novel syntax and semantics for heuristic directives in ASP that improve this situation.
We assume that the underlying solver can assign one of three values to any atom: \emph{true} (denoted with $\sigT$), \emph{false} ($\sigF$), and \emph{must-be-true} ($\sigM$) \cite<cf.>{DBLP:conf/lpnmr/Weinzierl17}.
The following definitions can be used without modification for solvers that do not use the third truth value $\sigM$. The set of atoms assigned must-be-true will be empty in this case.
\begin{definition}[Heuristic Directive]
	\label{def-heuristic-directive}
	A \emph{heuristic directive} is of the form \cref{eq-alpha-heuristic-directive}, where
	$ha_i$ ($0\leq i \leq n$) are heuristic atoms of the form $\heusign_i~a_i$, in which $\heusign_0 \in \{ \{ \heusignF \}, \{ \heusignT \} \}$ and $\heusign_i \subseteq \{ \heusignF, \heusignM, \heusignT~\}$ are sets of sign symbols and $a_i$ is an atom, and $w$ and $l$ are integer terms.
	\begin{align}
	&\heudirstmt ~~ \heuat_0 : \heuat_1, \dots, \heuat_k, \naf{\heuat_{k+1}}, \dots, \naf{\heuat_n}. \qquad [w@l]&
	\label{eq-alpha-heuristic-directive}
	\end{align}
	The heuristics' head is given by $\heuat_0$ and its condition by $\{ \heuat_1, \dots, \heuat_k$, $\naf{\heuat_{k+1}}$, \dots, $\naf{\heuat_n} \}$, which is similar to a rule body.
\end{definition}
Where the meaning is clear from the context, we may omit all symbols except sign symbols themselves in a set of sign symbols, e.g., we write $\heusignT\heusignM$ instead of $\{ \heusignT, \heusignM \}$.

\rev{}{
	Similarly to a rule (cf.\ \cref{sec-preliminaries-semantics}), a heuristic directive can contain variables and must be \emph{safe}.
	Since the exact definition of safety of a heuristic directive depends on implementation matters, we will address it in more detail in \cref{sec-domspec-lazygrounding-grounding}.
}

The textual syntax of a heuristic directive to be used in answer-set programs is defined by the following EBNF grammar, where $\langle\mathit{classical\_atom}\rangle$ stands for a classical atom,
and $\langle\mathit{term}\rangle$ stands for a term as defined in \cref{sec-preliminaries-syntax}:

\setlength{\grammarindent}{12em}
\begin{grammar}
	<heuristic\_directive> ::= ‘\#heuristic’ <head\_atom> [ <body> ] ‘.’ [ <annotation> ]
	
	<head\_atom> ::= [ <head\_sign> ] <classical\_atom>
	
	<head\_sign> ::= ‘\heusignT’ | ‘\heusignF’
	
	<body> ::= ‘:’ <body\_literals\_list>
	
	<body\_literals\_list> ::= <body\_literal> [ ‘,’ <body\_literals\_list> ]
	
	<body\_literal> ::= [ ‘not’ ] <body\_atom>
		
	<body\_atom> ::= ([ <body\_sign\_list> ] <classical\_atom>)
	
	<body\_sign\_list> ::= [ ‘\heusignT’ ] [ ‘\heusignM’ ] [ ‘\heusignF’ ]
		
	<annotation> ::= ‘[’ <weight> [ ‘@’ <level> ] ‘]’
	
	<weight> ::= <term>
	
	<level> ::= <term>
\end{grammar}
Note that the definition of $\langle\mathit{body\_sign\_list}\rangle$ imposes a specific order in which heuristic signs have to appear (first $\heusignT$, then $\heusignM$, and finally $\heusignF$). However, we assume that an implementation allows heuristic signs to appear in any order (e.g., $\heusignF\heusignT$).

\begin{example}
\label{ex:directive1}
Consider the following heuristic directive $\heudir$:
\begin{align*}
&\heudirstmt\ \heusignF~\mathrm{a} : \heusignT\heusignM~\mathrm{b}, \heusignT~\mathrm{c}, \naf{\heusignT\heusignM\heusignF~\mathrm{\rev{d }{e}}}.
\end{align*}
This directive means that the atom $\mathrm{a}$ shall be assigned $\sigF$ if $\mathrm{b}$ is assigned $\sigT$ or $\sigM$, $\mathrm{c}$ is assigned $\sigT$, and $\mathrm{\rev{d }{e}}$ is not assigned.
\end{example}

We now introduce some notation that will be used in further definitions.
The function $\fheuat$ maps a heuristic atom $\heuat_i$ \rev{}{of the form $\heusign_i~a_i$} to $a_i$ by removing the sign, and a set of heuristic atoms to the set of atoms occurring in them (e.g., $\fheuat(\sigM~\mathrm{a}) = \mathrm{a}$, $\fheuat(\{ \sigM~\mathrm{a}, \sigT~\mathrm{b} \}) = \{ \mathrm{a}, \mathrm{b} \}$).
The function $\fsignset$ maps a heuristic atom $\heuat_i$ to $s_i$ by removing the atom (e.g., $\fsignset(\sigM~\mathrm{a}) = \{ \sigM \}$).

The \emph{head} of a heuristic directive $\heudir$ of the form \cref{eq-alpha-heuristic-directive} is denoted by $\head(\heudir) = \heuat_0$, its \emph{weight} by $\func{weight}(\heudir) = w$ if given, else 0, and its \emph{level} by $\func{level}(\heudir) = l$ if given, else 0.
The \emph{(heuristic) condition} of a heuristic directive $\heudir$ is denoted by $\cond{\heudir} := \{ \heuat_1, \dots, \heuat_k,$ $\naf{\heuat_{k+1}}$, $\dots$, $\naf{\heuat_n} \}$,
the \emph{positive condition} is $\condpos{\heudir} := \{ \heuat_1, \dots, \heuat_k \}$ and the \emph{negative condition} is $\condneg{\heudir}$ $:=$ $\{ \heuat_{k+1}$, $\dots$, $\heuat_{n} \}$.

Let $\mi{HA}$ be a set of heuristic atoms.
Then, for $\heusign \subseteq \{ \heusignF, \heusignM, \heusignT~\}$, $\filterbysigns{\mi{HA}}{\heusign} = \{ a \mid \heusign~a \in \mi{HA} \}$ denotes the set of atoms in $\mi{HA}$ whose set of sign symbols equals $\heusign$.

\begin{example}
\label{ex:directive2}
Consider the heuristic directive $\heudir$ from \cref{ex:directive1} again:
\begin{align*}
&\heudirstmt\ \heusignF~\mathrm{a} : \heusignT\heusignM~\mathrm{b}, \heusignT~\mathrm{c}, \naf{\heusignT\heusignM\heusignF~\mathrm{\rev{d }{e}}}.
\end{align*}
Here, $\filterbysigns{\condpos{\heudir}}{\heusignM\heusignT} = \{ \mathrm{b} \}$;
$\filterbysigns{\condpos{\heudir}}{\heusignT} = \{ \mathrm{c} \}$;
and $\filterbysigns{\condpos{\heudir}}{\heusignF} = \emptyset$.
\\Furthermore, $\filterbysigns{\condneg{\heudir}}{\heusignAny} = \{ \mathrm{\rev{d }{e}} \}$; note that the order of sign symbols does not matter due to set semantics.
\end{example}

\rev{
A heuristic directive has one more syntactical restriction---it must be \emph{safe}.
Safety is a concept also used for rules in ASP.
Intuitively, all variables in a rule must occur in the rule's positive body such that the grounder can \emph{bind} all variables to constant values.
This binding would not be possible if a variable occurred only in a rule's head or negative body.
Since the exact definition of safety depends on implementation matters, we will address it in more detail in \cref{sec-domspec-lazygrounding-grounding}.
}{}

Our proposal differs from \slv{clingo}'s in the following ways, apart from
the syntactic\rev{al}{} differences between \cref{eq-clingo-heuristic-directive} and \cref{eq-alpha-heuristic-directive}:
\begin{itemize}
	\item Each heuristic atom \rev{in the condition}{} contains a set of sign symbols.
	Each sign symbol represents one of the truth values $\heusignF$ (\emph{false}), $\heusignT$ (\emph{true}), and $\heusignM$ (\emph{must-be-true}).
	\item In the condition, sign symbols provide a richer way of controlling when the condition is satisfied.
	A positive literal in the condition is satisfied if the truth value currently assigned to its atom is contained in its set of sign symbols (which is $\{ \heusignM, \heusignT~\}$ by default if not explicitly given).
	A negative literal in the condition is satisfied if the truth value currently assigned to its atom is \emph{not} contained in its set of sign symbols or if its atom is currently not assigned any truth value.
	\item In the heuristic head, the sign symbol is used to determine the truth value to be chosen by the heuristic\rev{s}{}.
	If $\heusign_0$ is $\heusignT$ or empty, the heuristics makes the solver guess $a_0$ to be true; if $\heusign_0$ is $\heusignF$, $a_0$ will be made false.\footnote{In the head, we only support truth values \sigT\ and \sigF\ because, from a user's point of view, it does not make sense to assign \sigM\ to an atom heuristically.}
	We do not use the modifier $m$.
	\item Instead of weight $w$ and tie-breaking priority $p$, we use terms $w$ and $l$ denoting weight and level as familiar from \rev{optimize-statements }{optimise-statements} in ASP-Core-2 \cite{DBLP:journals/tplp/CalimeriFGIKKLM20} or weak constraints in DLV \cite{DBLP:journals/tocl/LeonePFEGPS06}.
	The level is more important than the weight; both default to $0$, and together they are called priority.
\end{itemize}

\rev{Heuristics under \slv{clingo}'s semantics can easily be represented in our framework by replacing \enquote{\nafsymbol} by \enquote{$\heusignF$} and replacing weight and priority with appropriate values for weight and level.
The converse is presumed not to hold.}{}

We now describe our semantics more formally, beginning with the condition under which a heuristic atom is satisfied.
\begin{definition}[Satisfying a Heuristic Atom]
	\label{def-heuat-satisfied}
	Given a ground heuristic atom $\heuat$ and a partial assignment $\assignment$,
	$\heuat$ is \emph{satisfied} w.r.t.\ $\assignment$ iff:
	$\assignedtruthof{\fheuat(\heuat)} \in \fsignset(\heuat)$, i.e., if its atom is assigned a truth value that is included in the heuristic atom's sign set.\footnote{Note that the function $\assignedtruth$ maps to only one truth value even though $\sigT~a \in \assignment$ implies $\sigM~a \in \assignment$, so $\assignedtruthof{a} = \sigM$ iff $\sigM~a \in \assignment$ and $\sigT~a \notin \assignment$.}
\end{definition}
Whether a heuristic directive is satisfied depends on whether the atoms occurring in the directive are satisfied.
\begin{definition}[Satisfying a Heuristic Directive]
	\label{def-condition-satisfied}
	Given a ground heuristic directive $\heudir$ and a partial assignment $\assignment$
	, $\cond{\heudir}$ is \emph{satisfied} w.r.t.\ $\assignment$ iff:
	every $\heuat \in \condpos{\heudir}$ is satisfied and no \rev{$\heuat \in \condneg{\heuat}$ }{$\heuat \in \condneg{\heudir}$} is satisfied.
\end{definition}
Intuitively, a heuristic condition is satisfied iff its positive part is fully satisfied and none of its default-negated literals is contradicted.
\begin{definition}[Applicability of a Heuristic Directive -- Semantics]
	\label{def-directive-applicable}
	A ground heuristic directive $\heudir$ is \emph{applicable} w.r.t.\ a partial assignment $\assignment$ \rev{}{and a ground program $\prog$} iff:
	$
	\cond{\heudir}$ is satisfied, 
	\rev{}{$\exists r \in \prog$ s.t.\ $\head(r) = \fheuat(\head(\heudir))$ and $\{ \sigT~a \mid a \in \bodyp(r) \} \subseteq \assignment$ and $\{ \sigM~a \mid a \in \bodyn(r) \} \cap \assignment = \emptyset$,}
	and $\assignedtruthof{\fheuat(\head(\heudir))} \in \{ \sigU, \sigM \}$.
\end{definition}
Intuitively, a heuristic directive is applicable iff its condition is satisfied\rev{}{, there exists a currently applicable rule that can derive the atom in the heuristic directive's head,} and the atom in its head is assigned neither $\sigT$ nor $\sigF$.
If the atom in the head is assigned $\sigM$, the heuristic directive may still be applicable, because any atom with the non-final truth value $\sigM$ must be either $\sigT$ or $\sigF$ in any answer set.

\Cref{def-heuat-satisfied,def-condition-satisfied,def-directive-applicable} reveal the main difference between the semantics proposed here and the one implemented by \slv{clingo}:
In our approach, heuristic signs composed of truth values $\sigT$, $\sigM$, and $\sigF$ can be used in heuristic conditions to reason about atoms that are already assigned specific truth values in a partial assignment. Furthermore, default negation can be used to reason about atoms that are assigned \emph{or still unassigned}.
Our semantics truly means default negation in the current partial assignment, while the one implemented by \slv{clingo} amounts to strong negation in the current search state \rev{}{and a sophisticated modelling approach with priorities must be used to implement required heuristics}.
This difference is crucial since reasoning about incomplete information is essential in many cases.
An example is a heuristic for a configuration problem that only applies to components not yet placed.

What remains to be defined is the semantics of weight and level.
Given a set of applicable heuristic directives, one directive with the highest weight will be chosen from the highest level.
Suppose there are several \rev{}{heuristic directives} with the same maximum priority (i.e., weight and level). In that case, the solver can use domain-independent heuristics like VSIDS \cite{DBLP:conf/dac/MoskewiczMZZM01} as a fallback to break the tie.
\begin{definition}[The Subset of Heuristics Eligible for Immediate Choice]
	\label{def-maxpriority}
	Given a set $\heudirs$ of applicable ground heuristic directives, the \emph{subset eligible for immediate choice} is defined as $\func{maxpriority}(\heudirs)$ in two steps:
	\begin{align*}
	\func{maxlevel}(\heudirs) &:= \{ \heudir \mid \heudir \in \heudirs ~\mathrm{and}~ \func{level}(\heudir) = \max_{\heudir \in \heudirs}~\func{level}(\heudir) \} \\
	\func{maxpriority}(\heudirs) &:= \{ \heudir \mid \heudir \in \func{maxlevel}(\heudirs) ~\mathrm{and}~ \func{weight}(\heudir) = \max_{\heudir \in \func{maxlevel}(\heudirs)} \func{weight}(\heudir) \}
	\end{align*}
\end{definition}

After choosing a heuristic using $\func{maxpriority}$, a
solver makes a decision on the directive's head atom.
\rev{Note that heuristics only choose between atoms derivable by currently applicable rules.}{}%
Other solving procedures, e.g., deterministic propagation, are unaffected by processing heuristics.
\rev{}{In case no heuristic directive is applicable, the solver's default heuristic (e.g., VSIDS) makes a choice as usual.}

\begin{example}
	Consider the program given in \cref{sec-domspec-soa-ex1}.
	When converted to the syntax proposed in \cref{def-heuristic-directive}, its heuristic directives look like directives \cref{dir1,dir2} in the following program.
	Consider also the newly introduced directives \cref{dir3,dir4} in this program.
	\begin{align}
		&\heudirstmt ~~ \mathrm{a(5)}.~~[1]									\label{dir1}\\
		&\heudirstmt ~~ \mathrm{a(4)} ~~ : ~~ \naf{\mathrm{a(5)}}.~~[2]	\label{dir2}\\
		&\heudirstmt ~~ \heusignF~\mathrm{a(5)} ~~ : ~~ \mathrm{a}(4).~~[2]			\label{dir3}\\
		&\heudirstmt ~~ \mathrm{a(6)} ~~ : ~~ \heusignF~\mathrm{a}(5), \heusignT~\mathrm{a}(4).~~[2]			\label{dir4}
	\end{align}

	Intuitively, directive \cref{dir1} unconditionally prefers to make $\mathrm{a(5)}$ \rev{true }{\sigT} with weight 1.
	All other directives have a higher weight, 2, but they become applicable at different times.
	Directive \cref{dir2} prefers to make $\mathrm{a(4)}$ \rev{true }{\sigT} if $\mathrm{a(5)}$ is neither \rev{true }{\sigT} nor \rev{must-be-true }{\sigM}, directive \cref{dir3} prefers to make $\mathrm{a(5)}$ \rev{false }{\sigF} if $\mathrm{a(4)}$ is \rev{true }{\sigT} or \rev{must-be-true }{\sigM}, and \cref{dir4} prefers to make $\mathrm{a(6)}$ \rev{true }{\sigT} if $\mathrm{a(5)}$ is \rev{false }{\sigF} and $\mathrm{a(4)}$ is \rev{true }{\sigT}.
	
	Let $\assignment_0 = \emptyset$ be the empty partial assignment before any decision has been made.
	W.r.t.\ $\assignment_0$, \cref{dir1} is applicable because its condition is empty and its head is still unassigned.
	Directive \cref{dir2} is also applicable because $\mathrm{a}(5)$ is still unassigned.
	Directives \cref{dir3,dir4} are not applicable w.r.t.\ $\assignment_0$.
	Directive \cref{dir2} is chosen because it has the highest priority among applicable directives.
	Thus, $\mathrm{a}(4)$ is assigned $\sigT$, updating our assignment to $\assignment_1 = \{ \sigM~\mathrm{a(4)}, \sigT~\mathrm{a(4)} \}$.
	This makes \cref{dir3} applicable, $\mathrm{a}(5)$ is assigned $\sigF$ and our assignment is $\assignment_2 = \{ \sigM~\mathrm{a(4)}, \sigT~\mathrm{a(4)}, \sigF~\mathrm{a(5)} \}$.
	Note that the condition of \cref{dir2} was still satisfied at this point, but it was not applicable because its head was already assigned.
	Now, \cref{dir1} is also not applicable anymore, and the only directive that remains is \cref{dir4}.
	Since \cref{dir4} is applicable, $\mathrm{a(6)}$ is made \rev{true }{\sigT} and added to the assignment.
	Next, the atoms that remained unassigned are guessed by the default heuristics until an answer set is found.
\end{example}

\begin{example}
	\label{ex-pathfinding-alpha}
	Continuing from \cref{ex-pathfinding-clingo}, now using heuristic directives of the form \cref{eq-alpha-heuristic-directive}, it becomes possible to restrict the condition of the Pathfinding heuristics as desired.
	\begin{align*}
		\heudirstmt ~~ \mathrm{move(X,Y,T)} ~:~ &\mathrm{at(AtX,AtY,T)}, \mathrm{neighbour(AtX,AtY,X,Y)},\\
		&\naf \mathrm{obstacle(X,Y)}, \mathrm{goal(GoalX,GoalY)}, \mathrm{max\_time(MaxT)},\\
		&\naf \heusignT~\mathrm{visited(X,Y)}, \naf \heusignT~\mathrm{moved(T)}.\\
		&[(|\mathrm{X}-\mathrm{GoalX}| + |\mathrm{Y}-\mathrm{GoalY}|) * -1]
	\end{align*}
	Compared to \cref{ex-pathfinding-clingo}, now we can use the additional conditions $\naf \heusignT~\mathrm{visited(X,Y)}$ and $\naf \heusignT~\mathrm{moved(T)}$ that relate to the current partial assignment.
	Through these conditions, the heuristic only suggests moves that do not visit squares repeatedly, and it does not suggest to move to several squares at the same time.
	\rev{Here, it is essential to use the $\sigT$ sign to avoid that the condition is accidentally switched off in case $\mathrm{moved(T)}$ or $\mathrm{visited(X,Y)}$ is propagated to \sigM.
        }{}
	
	Note that the annotation looks different:
	Since negative weights make sense in our semantics, we can use $(|\mathrm{X}-\mathrm{GoalX}| + |\mathrm{Y}-\mathrm{GoalY}|) * -1$ instead of $\mathrm{MaxT} - (|\mathrm{X}-\mathrm{GoalX}| + |\mathrm{Y}-\mathrm{GoalY}|)$.
	Thus, the higher the Manhattan score of a square, the lower (i.e., nearer to $-\infty$) is the heuristic weight, and the less attractive it will be to move to that square.
	Furthermore, the modifier \enquote{true} has been dropped because we directly encode the heuristic's polarity in the directive's head.
\end{example}

\rev{}{
	Distinguishing between $\sigT\sigM$ and $\sigT$ facilitates fine-grained control of when a heuristic is active.
	$\sigT$ is used when it is essential that the $\sigT$ assignment is already justified (or not yet justified, in case default negation is used), and $\sigT\sigM$ is used when this is not important, as long as the atom must be true in a valid answer set.
	In \cref{ex-pathfinding-alpha}, $\naf \heusignT~\mathrm{visited(X,Y)}$ is used to switch off the heuristic when the agent \emph{has moved} to $(X,Y)$.
	The condition $\naf \heusignT\heusignM~\mathrm{visited(X,Y)}$, instead, would cause the heuristic to be switched off also when the solver only knows that the agent \emph{needs to move} to $(X,Y)$ at some point.
	This situation can be caused by a constraint forcing the agent to move to a specific square $(X,Y)$ at some point, because then the solver would propagate $\sigM$ to $\mathrm{visited(X,Y)}$.
	On the other hand, in many cases we want the condition to cover both $\sigT$ and $\sigM$.
	\Cref{sec-applications-and-experiments} will include many cases where the sign set $\sigT\sigM$ is used (recall that this is the default sign set used whenever none is given).
}

\section{Integration into a Lazy-Grounding ASP Solver}
\label{sec-domspec-lazygrounding}

This section presents how the above domain-specific heuristics can be realised within a lazy-grounding ASP system.
We chose \alphaslv\ as the basis for the implementation as it currently is the
most efficient lazy-grounding ASP system.\footnote{\alphaslv\ sources and binaries can be found on \url{https://github.com/alpha-asp/Alpha}. Features described in this section have been implemented on the \texttt{domspec\_heuristics\_extended} branch\rev{~and will soon be merged to \texttt{master}}{}.}
The presentation, therefore, contains
parts specific to \rev{lazy-grounding systems and}{} \alphaslv{}.
Our declarative domain-specific heuristics may also be realised in traditional ground-and-solve systems.
Note, however, that traditional ground-and-solve systems\rev{~have to ground all rules in the potential search space. Lazy grounding avoids this; hence domain-specific heuristics may}{, in contrast to lazy-grounding systems, will have to instantiate all heuristic directives. This additional overhead may cause domain-specific heuristics to} perform worse if implemented on top of a ground-and-solve system.

\subsection{Transforming Sign Sets}
\label{sec-domspec-lazygrounding-splitting}

In order to simplify the implementation, heuristic atoms with certain sign sets are transformed in a preprocessing step.
Every heuristic directive of the form \cref{eq-alpha-heuristic-directive} is equivalent to a set of heuristic directives containing only sign sets $\sigF$, $\sigT$, \rev{}{and} $\sigM\sigT$.
Note that every heuristic directive is also equivalent to a set of heuristic directives containing only singleton sign sets $\sigF$, $\sigM$, and $\sigT$;
however, we are including $\sigM\sigT$ instead of $\sigM$ to allow for a more efficient implementation.
Using \sigM\sigT\ instead of \sigM\ is necessary for re-using \alphaslv's nogood propagation capabilities, which will be introduced in \cref{sec-domspec-lazygrounding-alpha,sec-domspec-lazygrounding-grounding}.

Since heuristic directives can be reduced to directives using only sign sets $\sigF$, $\sigT$, $\sigM\sigT$, all other sign sets ($\heusignM$, $\heusignFM$, $\heusignFT$, and $\heusignAny$) can be transformed while retaining the semantics of the whole set of heuristic directives in the program.
\rev{
We call sign sets $\heusignFM$, $\heusignFT$, and $\heusignAny$ \emph{splittable} in the following.
Let $\Splittable = \{ \heusignFM, \heusignFT, \heusignAny \}$ denote the set of splittable sign sets.
Functions $\splitpos$ and $\splitneg$ are used to denote results of splitting:
$\splitpos(\heusignFM) = \heusignM$, $\splitpos(\heusignFT) = \heusignT$, $\splitpos(\heusignAny) = \heusignM\heusignT$, and $\splitneg(\splittable) = \heusignF$ for all $\splittable \in \Splittable$.

We distinguish four cases in which sign sets are transformed:
\begin{enumerate*}[label=(\arabic*)]
	\item splittable sign set in a positive condition; \label{transform-case-1}
	\item splittable sign set in a negative condition; \label{transform-case-2}
	\item \sigM\ in a positive condition; \label{transform-case-3}
	\item \sigM\ in a negative condition. \label{transform-case-4}
\end{enumerate*}

A directive $\heudir$ containing a splittable sign set in its positive condition \ref{transform-case-1} is transformed by picking one heuristic atom $\heuat$ with a splittable sign set, i.e., $\heuat \in \condpos{\heudir}$ where $\fsignset(\heuat) \in \Splittable$, and replacing $\heudir$ by two new directives:
one in which $\heuat = \splittable\ a$ is replaced by $\splitpos(\splittable)\ a$, and one in which $\heuat$ is replaced by $\splitneg(\splittable)\ a$.

A directive $\heudir$ containing a splittable sign set in its negative condition \ref{transform-case-2} is transformed by picking one heuristic atom $\heuat \in \condneg{\heudir}$ where $\fsignset(\heuat) \in \Splittable$, and replacing $\heuat = \splittable\ a$ inside $\heudir$ by two new heuristic atoms:
one in which $\splittable$ is replaced by $\splitpos(\splittable)$, and one in which $\splittable$ is replaced by $\splitneg(\splittable)$.

A directive $\heudir$ containing a sign set \sigM\ in its positive condition \ref{transform-case-3} is transformed by picking one heuristic atom $\sigM~a \in \condpos{\heudir}$, replacing $\sigM~a$ inside $\condpos{\heudir}$ by $\sigM\sigT~a$, and adding $\sigT~a$ to the negative condition $\condneg{\heudir}$.
Intuitively, $\sigM~a$ is satisfied if $a$ is assigned $\sigM$ or $\sigT$, but not $\sigT$.

A directive $\heudir$ containing a sign set \sigM\ in its negative condition \ref{transform-case-4} is transformed by picking one heuristic atom $\sigM~a \in \condneg{\heudir}$, and replacing $\heudir$ by two new directives $\heudir'$ and $\heudir''$, where $\condpos{\heudir'} = \condpos{\heudir} \cup \{ \sigF\sigT~a \}$, $\condneg{\heudir'} = \condneg{\heudir} \setminus \{ \sigM~a \}$, $\condpos{\heudir''} = \condpos{\heudir}$, and $\condneg{\heudir''} = \condneg{\heudir} \cup \{ \heusignAny~a \} \setminus \{ \sigM~a \}$.
Intuitively, $\naf{\sigM~a}$ is satisfied if $a$ is assigned $\sigF$ or $\sigT$, or $a$ is unassigned.

A program $\prog$ is transformed in preprocessing by applying this procedure to the heuristic directives in $\prog$ repeatedly until only sign sets $\sigF$, $\sigT$, and $\sigM\sigT$ remain.
}
{
We define a helper function $\modcond$ that constructs a heuristic directive by replacing the positive and negative condition within a given heuristic directive:
$\modcond(\heudir, c^+, c^-) \mapsto \heudir'$, s.t.\ $\head(\heudir') = \head(\heudir), \condpos{\heudir'} = c^+, \condneg{\heudir'} = c^-, \func{weight}(\heudir') = \func{weight}(\heudir)$, and $\func{level}(\heudir') = \func{level}(\heudir)$.

We then define the function $\transform{}$ that transforms one heuristic directive into a set of equivalent directives.
The function non-deterministically picks and transforms one heuristic atom with a sign set $\heusign \in \{ \heusignM, \heusignFM, \heusignFT, \heusignAny \}$, or returns the unmodified input $d$ if no such atom exists.
\begin{subnumcases}{\label{eq:transform} \transform{}(\heudir) =}
	\transform{1}(\heudir, \heusign, a) & for some $\heusign~a \in \condpos{\heudir}$ and $\heusign \in \{ \heusignFM, \heusignFT, \heusignAny \}$ \label{eq:transform:1} \\
	\transform{2}(\heudir, \heusign, a) & for some $\heusign~a \in \condneg{\heudir}$ and $\heusign \in \{ \heusignFM, \heusignFT, \heusignAny \}$ \label{eq:transform:2} \\
	\transform{3}(\heudir, a) & for some $\heusignM~a \in \condpos{\heudir}$ \label{eq:transform:3} \\
	\transform{4}(\heudir, a) & for some $\heusignM~a \in \condneg{\heudir}$ \label{eq:transform:4} \\
	\{ \heudir \} & if $\nexists~\heusign~a \in \condpos{\heudir} \cup \condneg{\heudir}$ s.t.\ $\heusign \in \{ \heusignM, \heusignFM, \heusignFT, \heusignAny \}$. \label{eq:transform:else}
\end{subnumcases}
\begin{align*}
	\transform{1}(\heudir, \heusign, a) = \{	& \modcond(\heudir, \condpos{\heudir} \cup \{ \heusign {\setminus} \heusignF~a \} \setminus \{ \heusign~a \}, \condneg{\heudir}), \\
								& \modcond(\heudir, \condpos{\heudir} \cup \{ \heusignF~a \} \setminus \{ \heusign~a \}, \condneg{\heudir}) \} \\[0.5em]
	\transform{2}(\heudir, \heusign, a) = \{ & \modcond(\heudir, \condpos{\heudir}, \condneg{\heudir} \cup \{ \heusign {\setminus} \heusignF~a, \heusignF~a \} \setminus \{ \heusign~a \}) \} \\[0.5em]
	\transform{3}(\heudir, a) = \{ & \modcond(\heudir, \condpos{\heudir} \cup \{ \heusignM\heusignT~a \} \setminus \{ \heusignM~a \}, \condneg{\heudir} \cup \{ \heusignT~a \}\})\} \\[0.5em]
	\transform{4}(\heudir, a) = \{ & \modcond(\heudir, \condpos{\heudir} \cup \{ \heusignFT~a \}, \condneg{\heudir} \setminus \{ \heusignM~a \}), \\
								& \modcond(\heudir, \condpos{\heudir}, \condneg{\heudir} \cup \{ \heusignAny~a \} \setminus \{ \heusignM~a \}) \}
\end{align*}
Recall that $\heusignF$ is an abbreviation for $\{ \heusignF \}$, thus $\heusign {\setminus} \heusignF$ is the same as $\heusign {\setminus} \{ \heusignF \}$.

Intuitively, the transformations work as follows:

\eqrefformat{\ref{eq:transform:1}} transforms one directive into two new directives.
In this process, one sign set containing $\heusignF$ is split into two partitions, $\heusignF$ and the remaining signs.
The original directive is satisfied iff one of the new directives is satisfied.

\eqrefformat{\ref{eq:transform:2}} transforms one directive into one new directive.
This time, one heuristic atom in the negative condition whose sign set contains $\heusignF$ is split into two.
The original heuristic atom is satisfied iff one of the two new heuristic atoms is satisfied.

\eqrefformat{\ref{eq:transform:3}} represents the informal equivalence that \enquote{$\heusignM$} is the same as \enquote{$\heusignM$ or $\heusignT$, but not $\heusignT$}, while
\eqrefformat{\ref{eq:transform:4}} represents that \enquote{$\naf{\heusignM}$} is the same as \enquote{$\heusignF$ or $\heusignT$ or unassigned}.

Finally, due to \eqrefformat{\ref{eq:transform:else}}, a heuristic directive is unaffected by this transformation if it contains only the sign sets $\sigF$, $\sigT$, and $\sigM\sigT$.

\Cref{fig:visualproof} visualises all possible transformation steps done by $\transform{}$.
Every possible sign set is represented by two nodes---one for positive and one for negative literals.
Sign sets unaffected by the transformation are shown as rectangular nodes, the others as ellipses.
Each arc corresponds to a transformation step and is labelled with a symbol corresponding to one of the cases in \cref{eq:transform}.
\begin{figure}
	\centering
	\includegraphics[width=\textwidth]{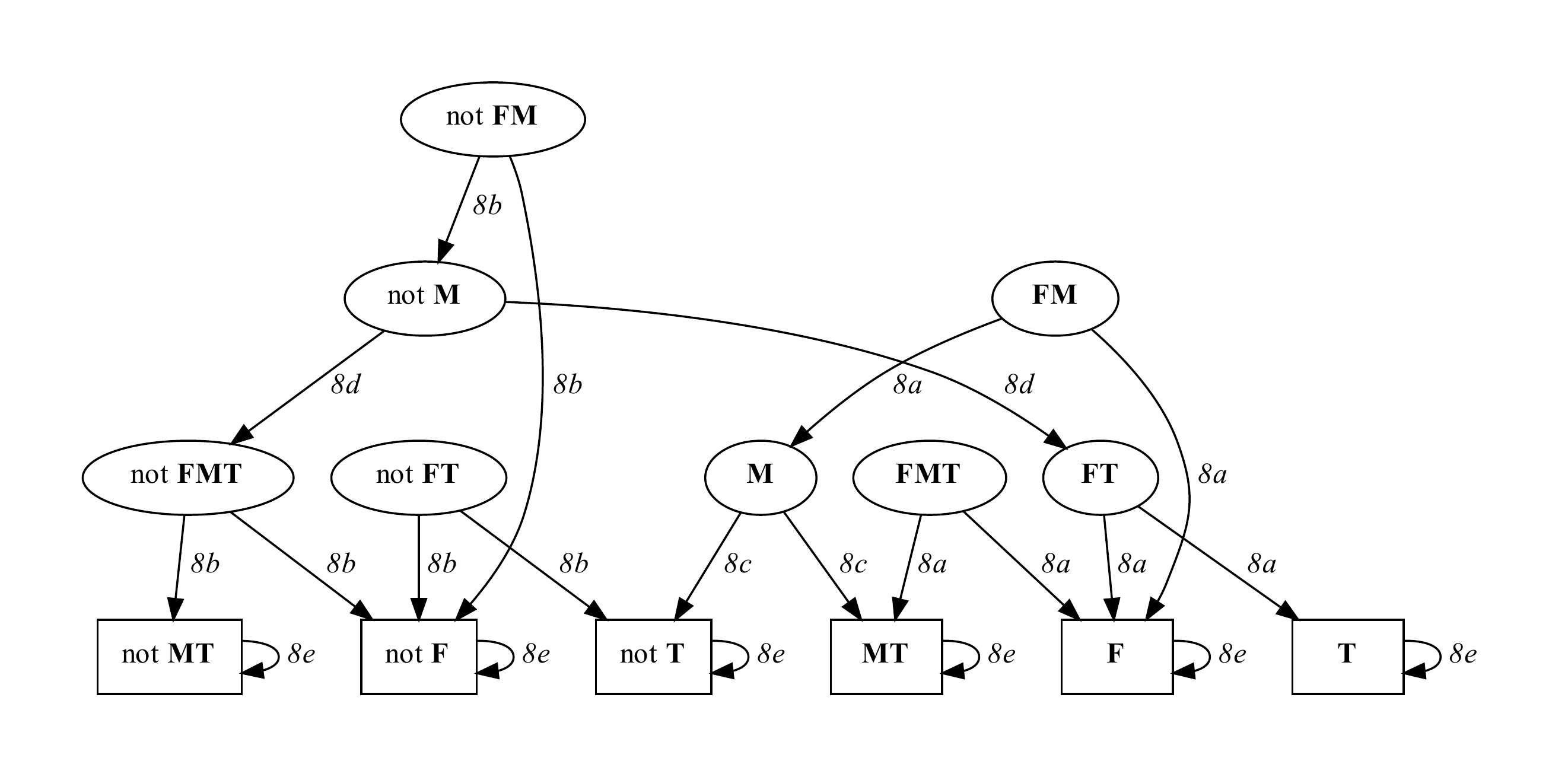}
	\caption{\rev{}{Visualisation of the transformation steps done by $\transform{}$ (see also \cref{lemma:transformation1})}}
	\label{fig:visualproof}
\end{figure}
}

\begin{example}
	Consider the following heuristic directive:
	\begin{align*}
		&\heudirstmt~ \mathrm{h} : \heusignAny~\mathrm{a}, \naf{\heusignFT~\mathrm{b}}.
	\end{align*}
	\rev{By splitting the splittable sign set in its negative condition, the heuristic directive is replaced by the following one: }
	{Transformation \eqrefformat{\ref{eq:transform:2}} replaces it by the following directive, removing $\heusignFT$ in the negative condition:}
	\begin{align*}
		&\heudirstmt~ \mathrm{h} : \heusignAny~\mathrm{a}, \naf{\heusignF~\mathrm{b}}, \naf{\heusignT~\mathrm{b}}.
	\end{align*}
	\rev{This directive is then replaced by two new ones, splitting the remaining splittable sign set in its positive condition: }
	{Then, transformation \eqrefformat{\ref{eq:transform:1}} removes $\heusignAny$ in the positive condition and thus produces two new directives:}
	\begin{align*}
		&\heudirstmt~ \mathrm{h} : \heusignF~\mathrm{a}, \naf{\heusignF~\mathrm{b}}, \naf{\heusignT~\mathrm{b}}.\\
		&\heudirstmt~ \mathrm{h} : \heusignM\heusignT~\mathrm{a}, \naf{\heusignF~\mathrm{b}}, \naf{\heusignT~\mathrm{b}}.
	\end{align*}
	These two directives together are semantically equivalent to the original one.
\end{example}

\begin{example}
	Consider the following heuristic directive containing \sigM\ in its positive and negative condition:
	\begin{align*}
		&\heudirstmt~ \mathrm{h} : \heusignM~\mathrm{a}, \naf{\heusignM~\mathrm{b}}.
	\end{align*}
	Transforming $\sigM~\mathrm{a}$ \rev{}{by \eqrefformat{\ref{eq:transform:3}}} in the positive condition results in the following directive:
	\begin{align*}
		&\heudirstmt~ \mathrm{h} : \heusignM\heusignT~\mathrm{a}, \naf{\heusignT~\mathrm{a}}, \naf{\heusignM~\mathrm{b}}.
	\end{align*}
	Transforming $\sigM~\mathrm{b}$ in the negative condition of the new directive, employing transformation \eqrefformat{\ref{eq:transform:4}}, yields two further directives:
	\begin{align*}
		&\heudirstmt~ \mathrm{h} : \heusignM\heusignT~\mathrm{a}, \heusignF\heusignT~\mathrm{b}, \naf{\heusignT~\mathrm{a}}.\\
		&\heudirstmt~ \mathrm{h} : \heusignM\heusignT~\mathrm{a}, \naf{\heusignT~\mathrm{a}}, \naf{\heusignAny~\mathrm{b}}.
	\end{align*}
	Both these conditions \rev{now contain splittable sign sets. When these are transformed, we obtain }{need to be transformed further (by \eqrefformat{\ref{eq:transform:1}} and \eqrefformat{\ref{eq:transform:2}}), thus obtaining} three directives in total:
	\begin{align*}
	&\heudirstmt~ \mathrm{h} : \heusignM\heusignT~\mathrm{a}, \heusignF~\mathrm{b}, \naf{\heusignT~\mathrm{a}}.\\
	&\heudirstmt~ \mathrm{h} : \heusignM\heusignT~\mathrm{a}, \heusignT~\mathrm{b}, \naf{\heusignT~\mathrm{a}}.\\
	&\heudirstmt~ \mathrm{h} : \heusignM\heusignT~\mathrm{a}, \naf{\heusignT~\mathrm{a}}, \naf{\heusignF~\mathrm{b}}, \naf{\heusignM\heusignT~\mathrm{b}}.
	\end{align*}
	These three directives together are semantically equivalent to the original one.
\end{example}

\rev{}{
Let $\heudirs$ be a set of heuristic directives, and let $\heudirs^1 = \bigcup_{\heudir \in \heudirs} \transform{}(\heudir)$ denote the result of replacing each heuristic directive by the result of heuristic directive transformation.
Furthermore, let $\heudirs^{n+1} = \bigcup_{\heudir \in \heudirs^n} \transform{}(\heudir)$ and $\heudirs^\ast = \lim_{n \to \infty} \heudirs^n$ denote the fixpoint of this operation.

\begin{lemma}
	\label{lemma:transformation1}
For any set of heuristic directives $\heudirs$, $\heudirs^\ast$ contains only sign sets $\sigF$, $\sigT$, and $\sigM\sigT$.
\end{lemma}
One can show that, no matter with which sign set and polarity the transformation starts, it always ends up with on of the sign sets $\sigF$, $\sigT$, and $\sigM\sigT$.
This is obvious from the visualisation of the transformation steps in \cref{fig:visualproof}.

\begin{lemma}
	\label{lemma:transformation2}
For any heuristic directive $\heudir$, it holds that $\heudir$ is applicable w.r.t.\ a partial assignment $\assignment$ iff $\transform{}(\heudir)$ contains a heuristic directive $\heudir'$ that is applicable w.r.t.\ $\assignment$ and whose head is the same as that of $\heudir$.
\end{lemma}
A proof sketch for \cref{lemma:transformation2} can be found in \cref{sec-proofs}.
}

\rev{}{
\subsection{Enhancing Heuristic Directives}
\label{sec-domspec-lazygrounding-enhancing}

Recall from \cref{def-directive-applicable} that for a heuristic directive to be applicable it is necessary that there exists an applicable rule that derives the atom in the heuristic directive's head, which enforces that atoms chosen by the heuristics have support.
This condition can be implemented by an additional transformation that works as follows:

The condition of every heuristic directive is enhanced by adding the body of the corresponding rule.
If there are several rules deriving the same head, one copy of the directive is created for each of them.
To ensure consistency of variable names, directive and rule are first standardised apart (so that they do not share any variable names) and the unifier of the two heads is applied to both the directive and the rule.

Since a rule is applicable if every positive body literal is assigned $\sigT$ and no atom in the negative body is assigned $\sigM$ (cf.\ \cref{sec-preliminaries-lazygrounding}), sign set $\sigT$ is used when copying a positive literal and sign set $\sigM\sigT$ is used when copying a negative literal.
Duplicate literals are not added to the heuristic condition.

\begin{example}
	Consider the following program:
	\begin{align*}
		& \mathrm{h(X)} \leftarrow \mathrm{a(X)}, \mathrm{b(X)}, \naf{\mathrm{c(X)}}.\\
		& \mathrm{h(X)} \leftarrow \mathrm{a(X)}, \mathrm{b(X)}, \naf{\mathrm{d(X)}}.\\
		&\heudirstmt~ \mathrm{h(N)} : \sigT~\mathrm{a(N)}.
	\end{align*}
	The transformation described above results in the following modified program:
	\begin{align*}
		& \mathrm{h(X)} \leftarrow \mathrm{a(X)}, \mathrm{b(X)}, \naf{\mathrm{c(X)}}.\\
		& \mathrm{h(X)} \leftarrow \mathrm{a(X)}, \mathrm{b(X)}, \naf{\mathrm{d(X)}}.\\
		&\heudirstmt~ \mathrm{h(N)} : \sigT~\mathrm{a(N)}, \sigT~\mathrm{b(N)}, \naf{\sigM\sigT~\mathrm{c(N)}}.\\
		&\heudirstmt~ \mathrm{h(N)} : \sigT~\mathrm{a(N)}, \sigT~\mathrm{b(N)}, \naf{\sigM\sigT~\mathrm{d(N)}}.
	\end{align*}
	In this example, $\mathrm{h(N)}$ (the head of the heuristic directive) has been unified with $\mathrm{h(X)}$ (the head of the deriving rules).
	The resulting unifier $\{ \mathrm{X} \mapsto \mathrm{N} \}$ has been applied to the rules before adding sign sets to their body literals and copying the resulting heuristic literals to the heuristic conditions.
	Duplicate literals (in this case, a duplicate of $\heusignT\ \mathrm{a(N)}$) have been omitted during this process.
\end{example}

}

\subsection{Lazy Grounding in the \alphaslv{} System}
\label{sec-domspec-lazygrounding-alpha}

To understand how domain-specific heuristics are integrated into the \alphaslv{} system, we recapitulate some of its inner workings first.
The \alphaslv{} system at its core contains a search loop similar to conflict-driven clause-learning (CDCL) employed in most SAT solvers \shortcite{cdcl,cdcl2} and modern ASP solvers like \slv{clingo} \shortcite{DBLP:journals/ai/GebserKS12} or \slv{dlv2} \shortcite{DBLP:conf/lpnmr/AlvianoCDFLPRVZ17}. In SAT, the encoding of the problem usually is given by clauses (a clause is a  set of ground literals where at least one must be satisfied). ASP uses nogoods instead (a nogood is a set of ground literals where all together must never be satisfied). Note that nogoods are dual to clauses, and they can represent precisely the same information.

In CDCL solvers, the primary mode of propagation is unit propagation which works as follows: \rev{given }{Given} a partial assignment $A$ and a nogood $g = \{l_1, \ldots, l_n\}$ where each $l_i$, $1 \leq i \leq n$ is a signed literal $\sigT~a$ or $\sigF~a$ with $a$ being a ground atom\rev{. Then}{}, $g$ is \emph{unit} w.r.t.\ $A$ if there exists $1 \leq i \leq n$ such that $l_i = s\ b$ with $\assignedtruthof{b} = \sigU$ and $g \setminus \{l_i\} \subseteq A$. Intuitively, a nogood $g$ is unit w.r.t.~$A$ if it is already violated except for one remaining literal $l_i$. To avoid violation of the nogood $g$, the atom $b$ of the remaining literal $l_i = s\ b$ must be assigned to the opposite truth value of $l_i$ in the end. So, whenever $g$ is unit, one can extend $A$ with the negation of $l_i$\rev{, i.e., with $\overline{l_i}$ where $\overline{\sigT~b} = \sigF~b$ and $\overline{\sigF~b} = \sigT~b$}{}.

In addition to unit propagation, many ASP solvers also employ some form of unfounded-set propagation and source pointers, which ensure that the assignment constructed by the solver is also well-founded (that is, free of self-founding positive cycles). \alphaslv{} does not use such propagation because \alphaslv's computation sequence, like in other lazy-grounding ASP solvers, already guarantees well-foundedness. 
Specifically, the solver can only guess on an applicable rule whether it fires or not (cf.\ \cref{sec-preliminaries-lazygrounding}), and not guess the truth value of an arbitrary unassigned atom. So, in \alphaslv, an atom $b$ is only assigned $\sigT$ if there is a well-founded rule firing whose head atom is $b$.

Moreover, the \alphaslv{} system uses a third truth value $\sigM$, called \emph{must-be-true}, which \alphaslv{} can assign to atoms considered true but not yet justified \cite<cf.>{DBLP:conf/ijcai/BogaertsW18}\rev{}{.}
This value allows further propagation in many circumstances and thus shrinks the search space. For example, the constraint $\leftarrow \naf \mathrm{d}.$ states that any answer set must assign true to the atom $\mathrm{d}$, but it gives no justification why $\mathrm{d}$ should hold, i.e., $\mathrm{d}$ must-be-true.
The solver needs to distinguish between a fact and a constraint because a fact makes $\mathrm{d}$ true and justifies it, while a constraint requires $\mathrm{d}$ to be true but does not justify it. To achieve this distinction, the notion of a nogood is slightly enhanced in \alphaslv{}: One literal in a nogood may be indicated as the nogood’s head.
\rev{
  Propagation in \alphaslv{} on a given assignment $\assignment$ and a nogood $g$ that is unit with $l_i = \sigF~b$ being the unassigned atom then works as follows: propagation assigns the atom $b$ the truth value $\sigT$ iff $l_i$ is the head of $g$ and all other literals $l_j \in g$, $j \neq i$ with $l_j = \sigT~c_j$ are assigned to true in $\assignment$, i.e., $\assignedtruthof{c_j} = \sigT$. Otherwise $b$ is assigned $\sigM$.
}{
  Given an assignment $\assignment$ and a nogood $g = \{ l_1, \ldots, l_n \}$, we say that $g$ is
  \begin{itemize}
  \item \emph{$\sigF$-unit} on $l_i$ w.r.t.\ $\assignment$ if $1 \leq i \leq n$, $l_i = \sigT b$, $\assignedtruthof{b} = \sigU$, and
    for all $l_j \in g$ with $j \neq i$ holds
    $\sigF c \in \assignment$ if $l_j = \sigF c$ and $\sigM c \in \assignment$ if $l_j = \sigT c$.
  \item \emph{$\sigM$-unit} on $l_i$ w.r.t.\ $\assignment$ if $1 \leq i \leq n$, $l_i = \sigF b$, $\assignedtruthof{b} = \sigU$, and
    for all $l_j \in g$ with $j \neq i$ holds
    $\sigF c \in \assignment$ if $l_j = \sigF c$ and $\sigM c \in \assignment$ if $l_j = \sigT c$.
  \item \emph{$\sigT$-unit} on $l_i$ w.r.t.\ $\assignment$ if $1 \leq i \leq n$, $l_i = \sigF b$ is the indicated head of $g$, $\assignedtruthof{b} = \sigU$ or $\assignedtruthof{b} = \sigM$ and
    for all $l_j \in g$ with $j \neq i$ holds
    $\sigF c \in \assignment$ if $l_j = \sigF c$ and $\sigT c \in \assignment$ if $l_j = \sigT c$.
  \end{itemize}
}
\rev{}{
  Given an assignment $\assignment$ and a nogood $g = \{ l_1, \ldots, l_n \}$, we denote by $\prop(g,\assignment)$ the truth value assigned by unit propagation, that is:
  \begin{center}$
  \prop(g,\assignment) = \begin{cases}
                            \{ \sigF b \} & \text{if $g$ is $\sigF$-unit on $l_i$ w.r.t.\ $\assignment$ and $l_i=\sigT b$,}\\
                            \{ \sigM b \} & \text{if $g$ is $\sigM$-unit but not $\sigT$-unit on $l_i$ w.r.t.\ $\assignment$ and $l_i=\sigF b$,}\\
                            \{ \sigM b, \sigT b \} & \text{if $g$ is $\sigT$-unit on $l_i$ w.r.t.\ $\assignment$ and $l_i=\sigF b$,}\\
                            \{ \} & \text{otherwise.}
                         \end{cases}$
  \end{center}
  Finally, given a set of nogoods $\Delta$ and an assignment $\assignment$, we denote by $\assignment^\Delta = \assignment \cup \bigcup_{g \in \Delta} \prop(g,\assignment)$ the assignment extended with all truth values that follow by unit propagation.
}

\begin{example}
  Consider the constraint $\leftarrow \naf \mathrm{d}.$, which is translated into the nogood $g_1 = \{ \sigF~\mathrm{d} \}$, and compare it with the fact $\mathrm{d}.$, which is translated into the nogood $g_2 = \{\underline{\sigF~\mathrm{d}}\}$. Here, $g_2$ is a nogood where the first and only literal is indicated as the head, so if $g_2$ is unit (which it always is unless $\mathrm{d}$ is already assigned), then $g_2$ will propagate the atom $\mathrm{d}$ to truth value $\sigT$. On the other hand, $g_1$ does not have a head, so if $g_1$ is unit, $\mathrm{d}$ is propagated only to $\sigM$ as $g_1$ does not justify that $\mathrm{d}$ is true but merely requires $\mathrm{d}$ to be derived by some \rev{other }{}rule.
\end{example}

While a constraint can be represented as one nogood, rules are represented using multiple nogoods. Given a non-ground rule $r$ and a grounding substitution $\sigma$ for $r$\rev{}{\ (i.e., a substitution that covers all variables in $\vars(r)$)}, let
\[
	r\sigma = h\sigma \leftarrow b_1\sigma, \ldots, b_m\sigma, \naf b_{m+1}\sigma, \ldots, \naf b_n\sigma\rev{}{.}
\]
be the rule where $\sigma$ is applied to every atom. Similarly to \slv{clingo}, the body of every ground rule in \alphaslv{} is represented with a fresh atom, denoted $\beta(r,\sigma)$. The set of nogoods representing $r\sigma$ then is:
\begin{align}
	\label{eq:nogood-beta-to-full-body}&\{ \underline{\sigF~\choiceatom{r}{\substitution}}, \sigT~b_1 \substitution, \dots, \sigT~b_m \substitution, \sigF~b_{m+1} \substitution, \dots, \sigF~b_n \substitution \}\\
	\label{eq:nogood-neg-head-to-beta} &\{ \underline{\sigF~h\substitution}, \sigT~\choiceatom{r}{\substitution}\}\\
	\label{eq:nogood-beta-to-posbodylit} &\{ \sigT~\choiceatom{r}{\substitution}, \sigF~b_1 \substitution \}, \dots, \{ \sigT~\choiceatom{r}{\substitution}, \sigF~b_m \substitution \}\\
	\label{eq:nogood-beta-to-negbodylit} &\{ \sigT~\choiceatom{r}{\substitution}, \sigT~b_{m+1} \substitution \}, \dots, \{ \sigT~\choiceatom{r}{\substitution}, \sigT~b_n \substitution \}
\end{align}
Notice that \eqref{eq:nogood-beta-to-full-body} ensures that the atom $\beta(r,\sigma)$ becomes true by unit propagation if the full body of $r\sigma$ is satisfied. If $\beta(r,\sigma)$ holds, then the rule's head also becomes true due to unit propagation on \eqref{eq:nogood-neg-head-to-beta}. Finally, the nogoods of \eqref{eq:nogood-beta-to-posbodylit} and \eqref{eq:nogood-beta-to-negbodylit} ensure that the atom representing the body of the rule, i.e., $\beta(r,\sigma)$ is false whenever one literal in the body is not satisfied.

According to the notion of a computation sequence, any rule whose negative body is non-empty offers a choice point when becoming applicable. To aid in detecting this applicability condition, such rules also yield the following nogoods:
\begin{align}
  \label{eq:nogood-choice-on} &\{ \underline{\sigF~\ChoiceOn(r,\sigma)}, \sigT~b_{1}, \dots, \sigT~b_m\}\\
  \label{eq:nogood-choice-off} &\{ \underline{\sigF~\ChoiceOff(r,\sigma)}, \sigT~b_{m+1}\}, \dots, \{ \underline{\sigF~\ChoiceOff(r,\sigma)}, \sigT~b_{n}\}
\end{align}
Intuitively, the nogood \eqref{eq:nogood-choice-on} makes $\ChoiceOn(r,\sigma)$ true whenever the positive body of $r\sigma$ is satisfied, and $\ChoiceOff(r,\sigma)$ is true whenever the current assignment contradicts the negative body of $r\sigma$ due to nogoods of \eqref{eq:nogood-choice-off}. Whenever $\ChoiceOn(r,\sigma)$ is true, and $\ChoiceOff(r,\sigma)$ is not true, the ground rule $r\sigma$ is applicable. Using these nogoods, \alphaslv{} can detect which ground rules are applicable, i.e., which choice points are allowed by the computation sequence.

For a more detailed description of nogoods and propagation using nogoods with heads in \alphaslv, we refer to \citeA{DBLP:conf/lpnmr/Weinzierl17} and \citeA{DBLP:conf/inap/LeutgebW17}.

\newcommand{\mustbetrue}{\textit{must-be-true}\xspace}

\subsection{Generating Nogoods for Heuristic Directives}
\label{sec-domspec-lazygrounding-grounding}

Similarly to normal rules, heuristic directives are represented using nogoods. To obtain ground instances of heuristic directives, each heuristic directive is transformed into a specifically marked rule such that the lazy-grounding procedures of \alphaslv{} generate all relevant ground instances of the marked rule. Since the rule is marked specifically, each ground instance is not translated into nogoods like an ordinary rule but translated as follows.

To evaluate sign symbols correctly, the following fact is exploited:
\alphaslv\ uses unit propagation to assign a truth value to the head of the nogood---the head is always a negative literal---when all other literals in the nogood are already satisfied.
The head atom will be assigned $\sigT$ if all positive literals in the nogood are assigned $\sigT$, and $\sigM$ if some positive literals are assigned $\sigM$\rev{}{\ (cf.\ \cref{sec-domspec-lazygrounding-alpha} for details)}.

Six solver-internal atoms are created for each ground heuristic directive.
Let $\heudir$ be a ground heuristic directive.
Then, the following atoms are created to be used in nogoods as described below:
$\HeuOnrm{T}(\heudir)$, $\HeuOnrm{MT}(\heudir)$, $\HeuOnrm{F}(\heudir)$, $\HeuOffrm{T}(\heudir)$, $\HeuOffrm{MT}(\heudir)$, and $\HeuOffrm{F}(\heudir)$.
These atoms are used similarly to the atoms $\ChoiceOn(r,\sigma)$ and $\ChoiceOff(r,\sigma)$, which are used to detect when an ordinary rule is applicable. Here, they serve the purpose of detecting when a heuristic directive becomes active. 

The following nogoods are generated:\footnote{\rev{}{Recall from \cref{sec-domspec-lazygrounding-splitting} that we can assume heuristic directives to contain only sign sets $\sigF$, $\sigT$, $\sigM\sigT$.}}
\begin{align}
&\text{For \dots}			&\text{this nogood is generated.} \nonumber \\
&\{a_1, \dots, a_n\} = \fheuat(\filterbysigns{\condpos{\heudir}}{\heusignT})				&\{ \underline{\sigF~\HeuOnrm{T}(\heudir)}, \sigT~a_1, \dots, \sigT~a_n \} \label{nogood-example1} \\
&\{a_{n+1}, \dots, a_m\} = \fheuat(\filterbysigns{\condpos{\heudir}}{\heusignM\heusignT}) 	&\{ \underline{\sigF~\HeuOnrm{MT}(\heudir)}, \sigT~a_{n+1}, \dots, \sigT~a_m \} \label{nogood-generated-2}\\
&\{a_{l+1}, \dots, a_k\} = \fheuat(\filterbysigns{\condpos{\heudir}}{\heusignF})			&\{ \underline{\sigF~\HeuOnrm{F}(\heudir)}, \sigF~a_{l+1}, \dots, \sigF~a_k \} \label{nogood-generated-3}
\end{align}
\begin{align}
&\text{And for every \dots}			&\text{this nogood is generated.} \nonumber \\
&\heusignT~a \in \condneg{\heudir}	&\{ \underline{\sigF~\HeuOffrm{T}(\heudir)}, \sigT~a \} \label{nogood-example2} \\
&\heusignM\heusignT~a \in \condneg{\heudir}	&\{ \underline{\sigF~\HeuOffrm{MT}(\heudir)}, \sigT~a \} \label{nogood-generated-5}\\
&\heusignF~a \in \condneg{\heudir}	&\{ \underline{\sigF~\HeuOffrm{F}(\heudir)}, \sigF~a \} \label{nogood-example-last}
\end{align}

Propagation treats these nogoods just as any other nogood.
For example, nogood \cref{nogood-example1} has the effect that $\HeuOnrm{T}(\heudir)$ will propagate to $\sigT$ if all $\{ a_1, \dots, ~a_n \}$ are assigned $\sigT$ because a heuristic is potentially switched on if all atoms in positive literals with sign set $\heusignT$ are true.

Likewise, nogood \cref{nogood-example2} has the effect that $\HeuOffrm{T}(\heudir)$ will propagate to $\sigT$ if $a$ is assigned $\sigT$ because a heuristic is switched off if an atom occurring negatively with $\heusignT$ is true.

\rev{}{
  \begin{lemma}
    \label{lemma-heu-iff-atom}
    Given a ground heuristic directive $\heudir$, an assignment $\assignment$, 
    and an atom\\
    $h \in \{\HeuOnrm{T}(d), \HeuOnrm{MT}(d), \HeuOnrm{F}(d), \HeuOffrm{T}(d), \HeuOffrm{MT}(d), \HeuOffrm{F}(d)\}$.\\
    Then, $\assignedtruthof{h} = \sigT$, respectively $\assignedtruthof{h} = \sigM$, iff there is one nogood $ng$ of the form \eqref{nogood-example1}-\eqref{nogood-example-last} and $ng$ is $\sigT$-unit, respectively $\sigM$-unit, on $\sigF h$ w.r.t.\ $\assignment' = \assignment \setminus \{\sigT h, \sigM h\}$, i.e., $\prop(ng,\assignment') = \{\sigT h, \sigM h\}$, respectively $\prop(ng,\assignment') = \{\sigM h\}$.
  \end{lemma}
  \Cref{lemma-heu-iff-atom} is proven in \cref{sec-proofs}.
}

To be able to re-use existing grounding strategies \cite<cf.>{DBLP:conf/lpnmr/TaupeWF19}, we require every heuristic directive to be \emph{safe}.
\begin{definition}[Safety of a Heuristic Directive]
	\label{def-heuristic-directive-safety}
	A heuristic directive of the form \cref{eq-alpha-heuristic-directive} is \emph{safe} if every variable occurring in it also occurs in heuristic atoms inside its positive condition whose set of sign symbols is either \heusignT\ or \heusignM\heusignT.
	More formally, a heuristic directive $\heudir$ is \emph{safe} if $\vars(\heudir) \subseteq \vars(\filterbysigns{\condpos{\heudir}}{\heusignT} \cup \filterbysigns{\condpos{\heudir}}{\heusignM\heusignT})$.
\end{definition}
Since every variable must appear in a positive body literal that does not use sign $\heusignF$, the safety of a heuristic directive is more restrictive than the safety of a rule \cite<cf.>{DBLP:journals/tocl/LeonePFEGPS06}.
The reason for this restriction is that \alphaslv\ grounds heuristic directives using the same techniques as when grounding normal rules.
A rule must be grounded no later than when all atoms in its positive body are assigned \sigT\ or \sigM\ because then the rule is potentially applicable.\footnote{A rule may also be grounded earlier \cite<cf.>{DBLP:conf/lpnmr/TaupeWF19}.}
Therefore, grounding of a rule is triggered when one of its positive body literals becomes satisfied.
Since \alphaslv\ does not trigger grounding when an atom is assigned false, only literals without \sigF\ can be used to bind variables.
Systems that do not impose such a restriction on when a rule is grounded may define heuristic directives' safety more liberally.
A heuristic directive is grounded, at the latest, when all atoms in its positive body whose sign set is \heusignT\ or \heusignM\heusignT\ are assigned \sigT\ or \sigM.
This is necessary for information on heuristic conditions to become known to the solver in time.

\subsection{Evaluating Heuristics During Solving}
\label{sec-domspec-lazygrounding-solving}

To process heuristics, the solver needs to recognise when a heuristic $\heudir$ is applicable in the current partial assignment $\assignment$.
To do this, the solver inspects the truth values of the $\HeuOnrm{}$ and $\HeuOffrm{}$ atoms that are propagated using the nogoods introduced in \cref{sec-domspec-lazygrounding-grounding}.

\begin{definition}[Satisfying a Heuristic Directive in Solving]
	A ground heuristic directive $\heudir$ is \rev{}{\emph{solving-satisfied}} in a partial assignment $A$ iff all of the following conditions hold:
	\[
		\begin{array}{ll}
			\assignedtruthof{\HeuOnrm{T}(\heudir)} = \sigT,					& \assignedtruthof{\HeuOffrm{T}(\heudir)} \neq \sigT, \\
			\assignedtruthof{\HeuOnrm{MT}(\heudir)} \in \{ \sigM, \sigT \},	& \assignedtruthof{\HeuOffrm{MT}(\heudir)} \notin \{ \sigM, \sigT \}, \\
			\assignedtruthof{\HeuOnrm{F}(\heudir)} = \sigT,					& \assignedtruthof{\HeuOffrm{F}(\heudir)} \neq \sigT.
		\end{array}
	\]
\end{definition}
Intuitively, a heuristic \rev{}{directive} is satisfied if it is enabled by all $\HeuOnrm{}$ atoms and not disabled by any $\HeuOffrm{}$ atom.
\rev{}{
  For the following correctness statements, we need the notion of an assignment that is deductively closed and consistent with a given program:
  Let $\prog$ be a ground program, an assignment $\assignment$ is \emph{deductively consistent} with $\prog$ if for every rule $r \in \prog$ such that the body of $r$ is satisfied w.r.t.\ $\assignment$ it holds that the head of $r$ is satisfied w.r.t.\ $\assignment$. Note that this also implies that $\assignment$ does not violate any constraint of $\prog$.
  \begin{lemma}
    \label{lemma-solving-satisfied-iff-cond}
    Given a ground program $\prog$ and its nogood representation $\Delta_\prog$, let $\heudir$ be a ground heuristic directive of $\prog$ and $\assignment$ be an assignment deductively consistent with $\prog$. Then, $\cond{\heudir}$ is satisfied w.r.t.~$\assignment$ and $\prog$ iff $\heudir$ is solving-satisfied w.r.t.~$\assignment^{\Delta_\prog}$, the assignment extended with all truth values that follow by unit propagation from $\Delta_\prog$.
  \end{lemma}
  \Cref{lemma-solving-satisfied-iff-cond} is proven in \cref{sec-proofs}.
}
\begin{definition}[Applicability of a Heuristic Directive in Solving]
  \label{def:solving-applicable}
A ground heuristic directive $\heudir$ is \rev{}{\emph{solving-applicable} w.r.t.~an assignment $\assignment$} iff $\heudir$ is \rev{}{solving-satisfied} and\\ $\assignedtruthof{\fheuat(\head(\heudir))}$ $\in$ $\{ \sigU, \sigM \}$.
\end{definition}
Intuitively, a heuristic directive is applicable if it is satisfied and the atom in its head is not yet assigned $\sigT$ or $\sigF$.
Note the similarity to \cref{def-condition-satisfied,def-directive-applicable}, where the semantics of a heuristic directive are defined in terms of a directive being satisfied and applicable.
\rev{}{The existence of a currently applicable rule that derives the atom in the heuristic directive's head is not required here because this condition is already enforced by the transformation described in \cref{sec-domspec-lazygrounding-enhancing}.}

\rev{}{
  \begin{theorem}
    Given a ground program $\prog$ and its nogood representation $\Delta_\prog$. Let $\heudir$ be a ground heuristic directive of $\prog$ and $\assignment$ be an assignment deductively consistent with $\prog$. Then,  $\heudir$ is applicable w.r.t.~$\assignment$ and $\prog$ iff $\heudir$ is solving-applicable w.r.t.~$\assignment^{\Delta_\prog}$.
  \end{theorem}

  \begin{proof}
    ``$\Rightarrow$'': Let $\heudir$ be applicable w.r.t.~$\assignment$, i.e.,
    \begin{enumerate*}[label=(\roman*)]
    \item $\cond{\heudir}$ is satisfied,
    \item $\exists r \in \prog$ s.t.\ $\head(r) = \fheuat(\head(\heudir))$ and $\{ \sigT~a \mid a \in \bodyp(r) \} \subseteq \assignment$ and $\{ \sigM~a \mid a \in \bodyn(r) \} \cap \assignment = \emptyset$, and
    \item $\assignedtruthof{\fheuat(\head(\heudir))} \in \{ \sigU, \sigM \}$.
    \end{enumerate*}
    From $(i)$ and $(ii)$ together with Lemma~\ref{lemma-solving-satisfied-iff-cond} we directly conclude that $\heudir$ is solving-satisfied. Since $\assignedtruthof{\fheuat(\head(\heudir))} \in \{ \sigU, \sigM \}$ holds due to $(iii)$, it follows from \cref{def:solving-applicable} that $\heudir$ is solving-applicable w.r.t.~$\assignment^{\Delta_\prog}$.

    ``$\Leftarrow$'': Let $\heudir$ be solving-applicable w.r.t.~$\assignment^{\Delta_\prog}$, i.e., $\heudir$ is solving-satisfied and\\ $\assignedtruthof{\fheuat(\head(\heudir))}$ $\in$ $\{ \sigU, \sigM \}$.
    Being solving-satisfied implies by Lemma~\ref{lemma-solving-satisfied-iff-cond} that $\cond{\heudir}$ is satisfied w.r.t.~$\assignment$ and $\prog$.
    Since the transformation described in~\cref{sec-domspec-lazygrounding-enhancing} ensured that
    $\exists r \in \prog$ s.t.\ $\head(r) = \fheuat(\head(\heudir))$ and $\{ \sigT a \mid a \in \bodyp(r) \} \subseteq \assignment$ and $\{ \sigM a \mid a \in \bodyn(r) \} \cap \assignment = \emptyset$ all hold,
    it follows from~\cref{def-directive-applicable} that $\heudir$ is \emph{applicable} w.r.t.~$\assignment$ and $\prog$.
  \end{proof}
}

Recall that the solver determines information on weight, level, and head of a heuristic directive $d$ from $\func{weight}(d)$, $\func{level}(d)$, and $\func{head}(d)$, respectively.

Finding the applicable \rev{heuristics }{heuristic directives} with the highest priority is aided by efficient data structures like a heap.
When an applicable heuristic \rev{}{directive} exists, the one with the highest priority is \emph{fired}, i.e., the solver uses it to make a choice.
However, the atom in the head of the heuristic \rev{}{directive} cannot be directly chosen since \alphaslv\ cannot choose arbitrary atoms but only guess applicable rules due to its reliance on computation sequences.
Therefore, an applicable rule that derives the atom in the head of the heuristic \rev{}{directive} must be identified. This rule is then guessed to fire or set to not fire, depending on the heuristic sign.

Notice that there may be more than one applicable rule deriving the head of the heuristic \rev{}{directive} that was fired, so it may be ambiguous which rule should be picked. Instead of picking a rule arbitrarily, we decided for our implementation to terminate with an error to disallow such occurrences where a firing heuristic directive finds multiple applicable rules deriving the same head. Other solutions are also possible, including ones where, through some rewriting, always at most one applicable rule is available. However, the occurrence of this issue likely hints at some underlying problems with the heuristics; therefore, we think it may be best to warn the user by raising an error dynamically.

\begin{example}
	Consider the following program $\prog$:
	\begin{align*}
		&\mathrm{x}(1..2).\\
		&\{ \mathrm{a(X)} : \mathrm{x(X)} \}.\\
		&\mathrm{b(X)} \leftarrow \mathrm{x(X)}, \naf \mathrm{c(X)}.\\
		&\mathrm{c(X)} \leftarrow \mathrm{x(X)}, \naf \mathrm{b(X)}.\\
		&\heudirstmt\ \mathrm{b(X)} : \mathrm{x(X)}, \naf \mathrm{a(X)}.~~[\mathrm{X}@2]
	\end{align*}
	Since the positive body of every rule in $\prog$ is satisfied, the full grounding of $\prog$ is immediately produced.
	Under the initial partial assignment consisting just of facts $\assignment_0 = \{ \sigM~\mathrm{x(1)}, \sigT~\mathrm{x(1)}, \sigM~\mathrm{x(2)}, \sigT~\mathrm{x(2)} \}$, both ground heuristic directives are applicable since both their positive bodies are satisfied and neither $\mathrm{a}(1)$ nor $\mathrm{a}(2)$ is assigned yet.
	The directive in which $X$ has been substituted by $2$ has the higher weight, however.
	For this reason, it is chosen, and the solver finds the only rule that can make the heuristic \rev{}{directive}'s head $\mathrm{b}(2)$ true: $\mathrm{b}(2) \leftarrow \mathrm{x}(2), \naf{\mathrm{c}(2)}.$
	The choice point representing the body of this rule is assigned true and, after some propagation, the new partial assignment will contain $\sigT~\mathrm{b}(2)$ (amongst other consequences of propagation).
\end{example}

\section{Applications and Experimental Results}
\label{sec-applications-and-experiments}

We tested our approach to declarative domain-specific heuristics by creating such heuristics for \rev{three }{several} example domains and solving these problem domains using our implementation in the \alphaslv\ system.
\rev{The }{Two concrete} domains under investigation \rev{are }{were} the House Reconfiguration Problem (HRP)\rev{,}{\ and} the Partner Units Problem (PUP)\rev{, and state-space search with A*}{}.
\rev{}{These two configuration problems are abstracted variants of typical configuration problems experienced in more than 25 years of applying AI technology in the area of automated configuration of electronic systems \cite{DBLP:journals/aim/FalknerFHSS16}. For all these applications heuristics exist which allow the efficient generation of satisfying solutions. Furthermore, we realised general state-space search in our approach, using the well-known informed search method A*\!, by which we abstract from concrete domains.  A* allows the incorporation of a heuristic evaluation function. 
By integrating A* into ASP, we can show that, on the one hand, we can exploit the power of informed (heuristic) search methods within ASP. On the other hand, we can employ the knowledge representation capabilities of ASP to specify states, actions, and their successor states declaratively.
Depending on the quality of the heuristic function an optimal solution is generated and the number of generated nodes is reduced. We applied A* to two specific search problems.}
\rev{These three domains }{All these applications} are presented in \cref{sec-applications-hrp,sec-applications-pup,sec-applications-astar} below, along with heuristics in the language proposed in this work.

To put ASP systems under stress, we used problem encodings and instances of varying sizes, where the larger instances were challenging to ground and solve.
More precisely, traditional grounders excessively consumed space or time when grounding these instances, and solving also was infeasible without the aid of domain-specific heuristics.
\rev{}{Additionally, for PUP, instances from the ASP competitions have been used, which are not challenging to ground but hard to solve.}

\subsection{Experimental Setup}
\label{sec-experiments-setup}

Encodings (including heuristics) and instances, and the \alphaslv\ binaries used for our experiments, are available in the online appendix accompanying our article and on our website.\footnote{\url{https://ainf.aau.at/dynacon}}
Details on the sources of the encodings are mentioned in the sections describing the domains.
Optimisation statements were not used since \alphaslv\ does not support them yet.
However, heuristic directives can be written in a way that optimal or near-optimal solutions are preferably found.\footnote{\rev{The optimality of solutions can sometimes be assessed by comparing the value of the objective function. }{}However, for some problems, the optimum is unknown.}

Problem instances were selected by first defining an instance-generating algorithm and then exploring instance sizes to find a set in which all systems could solve some instances under consideration within a time limit of \rev{10 }{15} minutes, and some instances could be solved by none (or very few) of these systems.
\rev{}{For PUP, competition instances were used additionally.}

\citeA{DBLP:conf/lpnmr/TaupeWF19} have introduced \emph{degrees of laziness in grounding}, which vary the conditions under which information about a ground rule is communicated from the grounder to the solver.
Traditionally in lazy grounding, rules are grounded when their positive body is fully satisfied.
In the concept of degrees of laziness, this is called \emph{strict} grounding.
\emph{Permissive} grounding, on the other hand, enables rules to be grounded if their positive body is not fully satisfied, as long as all variables can be bound by positive body literals that are already satisfied.

For example, consider the following non-ground constraint:
\[
	\leftarrow \mathrm{a(X,Y)}, \mathrm{b(X)}.
\]
Under the partial assignment $\assignment = \{ \sigM~\mathrm{a}(1,2),  \sigT~\mathrm{a}(1,2) \}$, the ground constraint
\[
	\leftarrow \mathrm{a}(1,2), \mathrm{b}(1).
\]
will only be produced if permissive grounding of constraints is enabled.

Due to experimental results, \citeA{DBLP:conf/lpnmr/TaupeWF19} suggested grounding constraints permissively and other rules strictly by default.
In our experiments, we investigated the performance of \alphaslv\ with both strict and permissive grounding of constraints.

\rev{}{Furthermore, \alphaslv\ was used without justification analysis \cite{DBLP:conf/ijcai/BogaertsW18} and without support for negative integers in aggregates because we observed these features to deteriorate performance in some cases.}
Apart from \rev{the strict and permissive grounding of constraints }{that}, \alphaslv\ was used in its default configuration.
The JVM running \alphaslv\
was called with command-line parameters \texttt{-Xms1G -Xmx32G}, thus initially allocating 1 GiB for Java's heap and setting the maximum heap size to 32 GiB.

For comparison, \slv{clingo}\footnote{\url{https://potassco.org/clingo/}} \cite{DBLP:journals/tplp/GebserKKS19} was used in version 5.4.0 and \slv{dlv2}\footnote{\url{https://dlv.demacs.unical.it/}} \cite{DBLP:conf/lpnmr/AlvianoCDFLPRVZ17} in version 2.1.0.

Each of the machines used to run the experiments
was equipped with two
Intel\textsuperscript{\textregistered} Xeon\textsuperscript{\textregistered} E5-2650 v4 @ 2.20GHz CPUs with 12 cores.
Furthermore, each machine had 251 GiB of memory and ran Ubuntu 16.04.1 LTS Linux.
Scheduling of benchmarks was done with HTCondor\textsuperscript{\texttrademark} together with the ABC Benchmarking System \cite{DBLP:conf/aiia/Redl16}.\footnote{\url{http://research.cs.wisc.edu/htcondor}, \url{https://github.com/credl/abcbenchmarking}}
Time and memory consumption were measured by \slv{pyrunlim},\footnote{\url{https://alviano.com/software/pyrunlim/}}
which was also used to limit time consumption to \rev{10 }{15} minutes per instance, memory to 40 GiB and swapping to 0.
Care was taken to avoid side effects between CPUs, e.g., by requesting exclusive access to an entire machine for each benchmark from HTCondor.

All solvers were configured to search for the first answer set of each problem instance.
Finding one or only a few solutions is often sufficient in industrial use cases since solving large instances can be challenging \shortcite{DBLP:journals/aim/FalknerFHSS16}.
Therefore, the domain-specific heuristics used in the experiments are designed to help the solver find one answer set that is \enquote{good enough}, even though it may not be optimal.

\subsection{Case Study 1: The House Reconfiguration Problem (HRP)}
\label{sec-applications-hrp}

The House Reconfiguration Problem (HRP) \rev{\cite{DBLP:conf/confws/FriedrichRFHSS11}}{\shortcite{DBLP:conf/confws/FriedrichRFHSS11}} is an abstracted version of industrial (re)configuration problems, e.g., rack configuration.

\subsubsection{Problem Definition}

Formally, HRP is defined as a modification of the House Configuration Problem (HCP).
\begin{definition}[HCP]
	The input for the \emph{House Configuration Problem (HCP)} is given by four sets of constants $P$, $T$, $C$, and $R$ representing persons, things, cabinets, and rooms, respectively, and an ownership relation $\mi{PT} \subseteq P \times T$ between persons and things. 
	
	The task is to find an assignment of things to cabinets $\mi{TC} \subseteq T \times C$ and cabinets to rooms $\mi{CR} \subseteq C \times R$, such that:
	\begin{enumerate*}[label=(\arabic*)]
		\item each thing is stored in a cabinet; 
		\item a cabinet contains at most five things; 
		\item every cabinet is placed in a room; 
		\item a room contains at most four cabinets; and 
		\item a room may only contain cabinets storing things of one person.
	\end{enumerate*}
\end{definition}

\begin{definition}[HRP]
	The input for the \emph{House Reconfiguration Problem (HRP)} is given by an HCP instance $H = \langle P, T, C, R, PT \rangle$, a legacy configuration $\langle \mi{TC}', \mi{CR}' \rangle$, and a set of things $T' \subseteq T$ that are defined as \enquote{long} (all other things are \enquote{short}).
	
	The task is then to find an assignment of things to cabinets $\mi{TC} \subseteq T \times C$ and cabinets to rooms $\mi{CR} \subseteq C \times R$, that satisfies all requirements of HCP as well as the following ones:
	\begin{enumerate*}[label=(\arabic*)]
		\item a cabinet is either small or high;
		\item a long thing can only be put into a high cabinet;
		\item a small cabinet occupies 1 and a high cabinet 2 of 4 slots
		available in a room;
		\item all legacy cabinets are small.
	\end{enumerate*}
\end{definition}

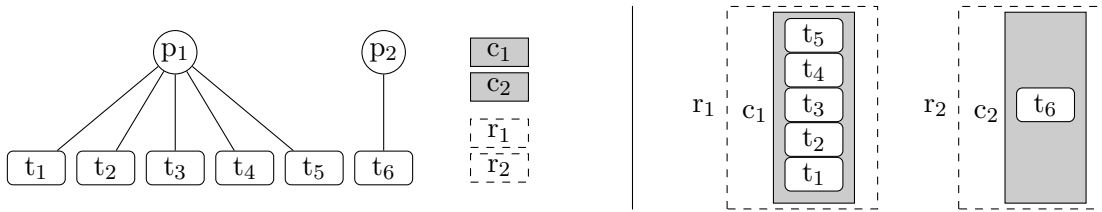
\begin{figure}
	\centering
\tikzstyle{person} = [draw, circle, inner sep=0em, minimum width=1.5em]
\tikzstyle{thing} =   [draw, rectangle, fill=white, inner sep=.2em, minimum width=2em, rounded corners=0.2em]
\tikzstyle{cabinet} =   [draw, rectangle, fill=black!20, inner sep=.2em, minimum width=2em]
\tikzstyle{room} =   [draw, rectangle, dashed, inner sep=.2em, minimum width=2em]

\begin{minipage}{.5\textwidth}
\begin{tikzpicture}[x=2em,y=2em,baseline=0pt]

\node[person] (p1) at (2.4, 1) {$\mathrm{p_1}$};
\node[person] (p2) at (6.0, 1) {$\mathrm{p_2}$};

\node[thing] (t1) at (0.0, -1) {$\mathrm{t_{1}}$};
\node[thing] (t2) at (1.2, -1) {$\mathrm{t_{2}}$};
\node[thing] (t3) at (2.4, -1) {$\mathrm{t_{3}}$};
\node[thing] (t4) at (3.6, -1) {$\mathrm{t_{4}}$};
\node[thing] (t5) at (4.8, -1) {$\mathrm{t_{5}}$};
\node[thing] (t6) at (6.0, -1) {$\mathrm{t_{6}}$};

\node[cabinet] (c1) at (8.0, 1) {$\mathrm{c_{1}}$};
\node[cabinet] (c2) at (8.0, 0.4) {$\mathrm{c_{2}}$};

\node[room] (r1) at (8.0, -0.4) {$\mathrm{r_{1}}$};
\node[room] (r2) at (8.0, -1) {$\mathrm{r_{2}}$};

\draw 
(p1)   -- (t1)
(p1)   -- (t2)
(p1)   -- (t3)
(p1)   -- (t4)
(p1)   -- (t5)
(p2)   -- (t6);
\end{tikzpicture}%
\end{minipage}%
\hfill\vline\hfill
\begin{minipage}{.4\textwidth}
\begin{tikzpicture}[x=2em,y=2em,baseline=0pt]

\draw[room] (-2, -0.6) rectangle node[left, label=left:$\mathrm{r_{1}}~~$]{} ++(2.6,3.5);
\draw[cabinet] (-1.2, -0.5) rectangle node[below, label=left:$\mathrm{c_{1}}$]{} ++(1.4,3.3);

\draw[room] (2, -0.6) rectangle node[left, label=left:$\mathrm{r_{2}}~~$]{} ++(2.6,3.5);
\draw[cabinet] (2.8, -0.5) rectangle node[below, label=left:$\mathrm{c_{2}}$]{} ++(1.4,3.3);

\node[thing] (t1) at (-0.5, 0.0) {$\mathrm{t_{1}}$};
\node[thing] (t2) at (-0.5, 0.6) {$\mathrm{t_{2}}$};
\node[thing] (t3) at (-0.5, 1.2) {$\mathrm{t_{3}}$};
\node[thing] (t4) at (-0.5, 1.8) {$\mathrm{t_{4}}$};
\node[thing] (t5) at (-0.5, 2.4) {$\mathrm{t_{5}}$};

\node[thing] (t6) at (3.5, 1.2) {$\mathrm{t_{6}}$};

\end{tikzpicture}
\end{minipage}
	\caption{Sample HRP instance (left) and one of its solutions (right)}
	\label{fig:hrp_example}
\end{figure}

The sample HRP instance shown in \cref{fig:hrp_example} comprises two cabinets, two rooms, five things that belong to person $\mathrm{p_1}$, and one thing that belongs to person $\mathrm{p_2}$. 
A legacy configuration is empty, and all things are small. 
In a solution, the first person's things are placed in cabinet $\mathrm{c_1}$ in the first room, and the thing of the second person is \rev{the }{in} cabinet $\mathrm{c_2}$ in the second room. 
For this sample instance, a solution of HRP corresponds to a solution of HCP.

\subsubsection{Encodings and Instances}

We adapted the original encoding by \citeA{DBLP:conf/confws/FriedrichRFHSS11} to conform to the current capabilities of \alphaslv\ (i.e., restricted usage of aggregates, no optimisation).

The main two choice rules guessing the assignment of things to cabinets and the assignment of cabinets to rooms look as follows:
\begin{align*}
	\{~ \mathrm{cabinetTOthing(C,T)} ~\} &\leftarrow \mathrm{cabinetDomain(C)}, \mathrm{thing(T)}.	\\
	\{~ \mathrm{roomTOcabinet(R,C)} ~\} &\leftarrow \mathrm{roomDomain(R)}, \mathrm{cabinet(C)}\rev{}{.}
\end{align*}

Instances consist of facts over the following predicates:
\rev{$\mathrm{cabinetDomain}/1$ defines potential cabinets and $\mathrm{roomDomain}/1$ defines potential rooms;
$\mathrm{thingLong}/1$ defines which things are long;
and $\mathrm{legacyConfig}/1$ defines all the other data in the legacy configuration,
e.g., $\mathrm{legacyConfig}(\mathrm{personTOthing}(\mathrm{p1},\mathrm{t1}))$ defines that person $\mathrm{p1}$ owns thing $\mathrm{t1}$, and $\mathrm{legacyConfig}(\mathrm{roomTOcabinet}(\mathrm{r1},\mathrm{c1}))$ specifies one tuple in the legacy assignment of cabinets to rooms. }
{
\begin{itemize}
	\item $\mathrm{cabinetDomain}/1$ defines potential cabinets,
	\item $\mathrm{roomDomain}/1$ defines potential rooms;
	\item $\mathrm{thingLong}/1$ defines which things are long; and
	\item $\mathrm{legacyConfig}/1$ defines all the other data in the legacy configuration, for example:
		\begin{itemize}
			\item  $\mathrm{legacyConfig}(\mathrm{personTOthing}(\mathrm{p1},\mathrm{t1}))$ defines that person $\mathrm{p_1}$ owns thing $\mathrm{t_1}$, and
			\item $\mathrm{legacyConfig}(\mathrm{roomTOcabinet}(\mathrm{r1},\mathrm{c1}))$ specifies one tuple in the legacy assignment of cabinets to rooms.
		\end{itemize} 
\end{itemize}
}

Instances for HRP were generated in the pattern of the original instances by \citeA{DBLP:conf/confws/FriedrichRFHSS11}.
This pattern represents four different reconfiguration scenarios encountered in practice\rev{}{, and the instances are abstracted real-world instances}.
Our instances are considerably larger than the original ones, though (ranging up to 800 things, while the original instances used at most 280 things).

\subsubsection{Heuristics}
\label{sec-applications-hrp-heuristics}

The domain-specific heuristic\rev{s}{} for HRP implemented in our novel approach works by
\begin{enumerate*}[label=(\arabic*)]
	\item first trying to re-use the legacy configuration;	\label{hrp-heu-1}
	\item then filling cabinets with things;				\label{hrp-heu-2}
	\item then filling rooms with cabinets;					\label{hrp-heu-3}
	\item and finally closing remaining choices.			\label{hrp-heu-4}
\end{enumerate*}
Long things are always assigned before short things.

By \enquote{closing remaining choices} we mean assigning \sigF\ to choice points not yet assigned by the heuristic\rev{s}{}.
The purpose of this is to avoid the default heuristics (e.g., VSIDS) from causing conflicts by choosing the wrong truth values.

We now present some selected heuristic directives.
The directives use some intermediate predicates whose meaning should become evident from their names.
The full encoding is available online.\footnote{\rev{}{\url{https://ainf.aau.at/dynacon}}}

The following heuristics re-use the legacy assignment of cabinets to things and of rooms to cabinets \ref{hrp-heu-1}:

\vspace{1em}
\begin{tabularx}{0.95\textwidth}{Xrr}
	$\heudirstmt\ \mathrm{reuse}(\mathrm{cabinetTOthing(C,T)}) :$ & & \\
		\multicolumn{2}{r}{$\mathrm{legacyConfig}(\mathrm{cabinetTOthing(C,T)}), \mathrm{thingLong(T)}.$} 		&$[4@4]$	\\[0.5em]
	$\heudirstmt\ \mathrm{reuse}(\mathrm{cabinetTOthing(C,T)}) :$ & & \\
		\multicolumn{2}{r}{$\mathrm{legacyConfig}(\mathrm{cabinetTOthing(C,T)}), \naf \mathrm{thingLong(T)}.$}	&$[3@4]$	\\[0.5em]
	$\heudirstmt\ \mathrm{reuse}(\mathrm{roomTOcabinet(R,C)}) :$ & & \\
		\multicolumn{2}{r}{$\mathrm{legacyConfig}(\mathrm{roomTOcabinet(R,C)}).$}							 	&$[2@4]$
\end{tabularx}
\vspace{1em}

The following heuristic assigns things to cabinets, preferring long over short things \ref{hrp-heu-2}:

\vspace{1em}
\begin{tabularx}{0.95\textwidth}{Xlr}
	\multicolumn{2}{l}{$\heudirstmt\ \mathrm{cabinetTOthing(C,T)} : $} & \\
		& $\mathrm{cabinetDomain(C)}, \naf \mathrm{fullCabinet(C)},$ & \\
		& $\naf \heusignT\ \mathrm{assignedThing(T)}, \mathrm{personTOthing(P,T)},$ & \\
		& $\naf \mathrm{otherPersonTOcabinet(P,C)},$ & \\
		& $\mathrm{maxCabinet(MC)}, \mathrm{thingLong(T)}.$ 	&$[(\mathrm{MC}{-}\mathrm{C})@3]$
\end{tabularx}

\begin{tabularx}{0.95\textwidth}{Xlr}
	\multicolumn{2}{l}{$\heudirstmt\ \mathrm{cabinetTOthing(C,T)} : $} & \\
		& $\mathrm{cabinetDomain(C)}, \naf \mathrm{fullCabinet(C)},$ & \\
		& $\naf \heusignT\ \mathrm{assignedThing(T)}, \mathrm{personTOthing(P,T)}$ & \\
		& $\naf \mathrm{otherPersonTOcabinet(P,C)},$ & \\
		& $\mathrm{maxCabinet(MC)}, \naf \mathrm{thingLong(T)}.$ 	&$[(\mathrm{MC}{-}\mathrm{C})@2]$
\end{tabularx}
\vspace{1em}

The following heuristic assigns cabinets to rooms \ref{hrp-heu-3}:

\vspace{1em}
\begin{tabularx}{0.95\textwidth}{Xlr}
	\multicolumn{2}{l}{$\heudirstmt\ \mathrm{roomTOcabinet(R,C)} : $} & \\
		& $\mathrm{roomDomain(R)}, \naf \mathrm{fullRoom(R)},$ & \\
		& $\mathrm{cabinet(C)}, \naf \heusignT\ \mathrm{assignedCabinet(C)},$ & \\
		& $\mathrm{personTOcabinet(P,C)}, \naf \mathrm{otherPersonTOroom(P,R)},$ & \\
		& $\mathrm{maxRoom(MR)}.$	 &$[(\mathrm{MR}{-}\mathrm{R})@1]$
\end{tabularx}
\vspace{1em}

Finally, the following heuristics close choice points that are still unassigned \ref{hrp-heu-4}:

\vspace{1em}
\begin{tabularx}{0.95\textwidth}{Xl}
	\multicolumn{2}{l}{$\heudirstmt\ \sigF\ \mathrm{cabinetTOthing(C,T)} :$} \\
		& $\naf \mathrm{cabinetTOthing(C,T)}, \mathrm{cabinetDomain(C)}, \mathrm{thing(T)}.$ 	\\[0.5em]
	\multicolumn{2}{l}{$\heudirstmt\ \sigF\ \mathrm{roomTOcabinet(R,C)} :$} \\
		& $\naf \mathrm{roomTOcabinet(R,C)}, \mathrm{roomDomain(R)}, \mathrm{cabinet(C)}.$
\end{tabularx}
\vspace{1em}

The heuristics we created for \alphaslv\ cannot be used with \slv{clingo} due to the usage of $\heusignT$ and default negation.

An alternative encoding containing heuristic directives for \slv{clingo} has also been created.
This encoding contains heuristic directives that have been faithfully adapted: by
using sign modifiers instead of sign symbols in heuristic heads;
by adding $(l{-}1)$ times the maximum weight from the next lower level to $w$ and omitting $p$;
and by removing all literals from the condition for which it only makes sense to evaluate them w.r.t.\ a partial assignment.

\rev{}{
	Furthermore, an \alphaslv\ encoding with heuristics without our novel features, compliant with those employed by \slv{clingo}, has been created.
}

\subsubsection{Results}
\label{sec-applications-hrp-results}

	\begin{figure}[t]
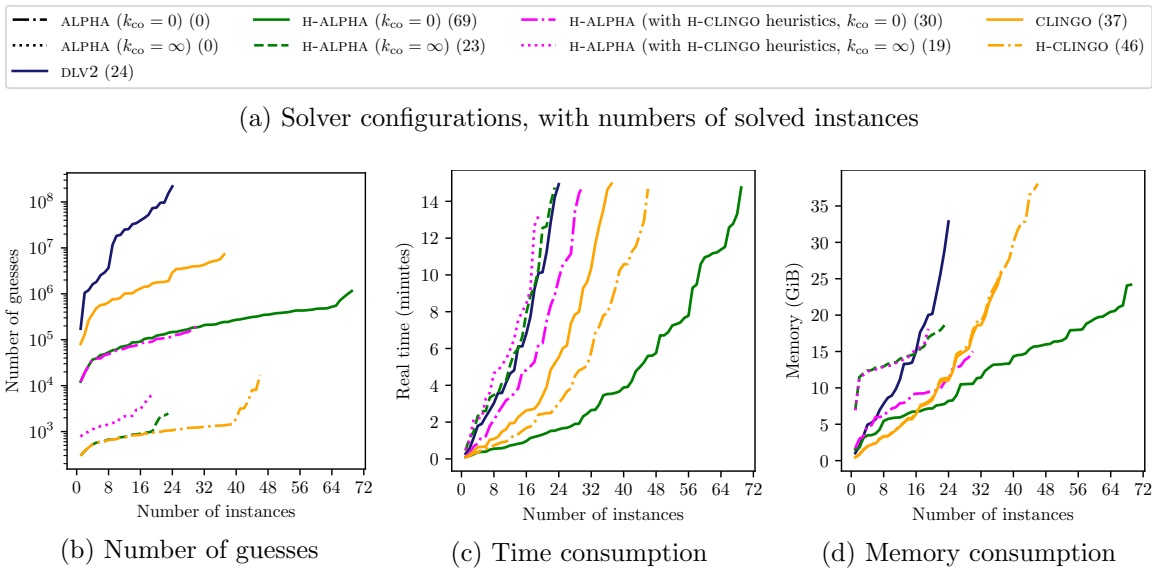

		\centering
		\begin{subfigure}{\textwidth}
			\centering
			\resizebox{\textwidth}{!}{
\begingroup%
\makeatletter%
\begin{pgfpicture}%
\pgfpathrectangle{\pgfpointorigin}{\pgfqpoint{9.290326in}{0.666667in}}%
\pgfusepath{use as bounding box, clip}%
\begin{pgfscope}%
\pgfsetbuttcap%
\pgfsetmiterjoin%
\definecolor{currentfill}{rgb}{1.000000,1.000000,1.000000}%
\pgfsetfillcolor{currentfill}%
\pgfsetlinewidth{0.000000pt}%
\definecolor{currentstroke}{rgb}{1.000000,1.000000,1.000000}%
\pgfsetstrokecolor{currentstroke}%
\pgfsetdash{}{0pt}%
\pgfpathmoveto{\pgfqpoint{0.000000in}{0.000000in}}%
\pgfpathlineto{\pgfqpoint{9.290326in}{0.000000in}}%
\pgfpathlineto{\pgfqpoint{9.290326in}{0.666667in}}%
\pgfpathlineto{\pgfqpoint{0.000000in}{0.666667in}}%
\pgfpathclose%
\pgfusepath{fill}%
\end{pgfscope}%
\begin{pgfscope}%
\pgfsetbuttcap%
\pgfsetmiterjoin%
\definecolor{currentfill}{rgb}{1.000000,1.000000,1.000000}%
\pgfsetfillcolor{currentfill}%
\pgfsetfillopacity{0.800000}%
\pgfsetlinewidth{1.003750pt}%
\definecolor{currentstroke}{rgb}{0.800000,0.800000,0.800000}%
\pgfsetstrokecolor{currentstroke}%
\pgfsetstrokeopacity{0.800000}%
\pgfsetdash{}{0pt}%
\pgfpathmoveto{\pgfqpoint{0.027778in}{0.000000in}}%
\pgfpathlineto{\pgfqpoint{9.262548in}{0.000000in}}%
\pgfpathquadraticcurveto{\pgfqpoint{9.290326in}{0.000000in}}{\pgfqpoint{9.290326in}{0.027778in}}%
\pgfpathlineto{\pgfqpoint{9.290326in}{0.638889in}}%
\pgfpathquadraticcurveto{\pgfqpoint{9.290326in}{0.666667in}}{\pgfqpoint{9.262548in}{0.666667in}}%
\pgfpathlineto{\pgfqpoint{0.027778in}{0.666667in}}%
\pgfpathquadraticcurveto{\pgfqpoint{-0.000000in}{0.666667in}}{\pgfqpoint{-0.000000in}{0.638889in}}%
\pgfpathlineto{\pgfqpoint{-0.000000in}{0.027778in}}%
\pgfpathquadraticcurveto{\pgfqpoint{-0.000000in}{0.000000in}}{\pgfqpoint{0.027778in}{0.000000in}}%
\pgfpathclose%
\pgfusepath{stroke,fill}%
\end{pgfscope}%
\begin{pgfscope}%
\pgfsetbuttcap%
\pgfsetroundjoin%
\pgfsetlinewidth{1.505625pt}%
\definecolor{currentstroke}{rgb}{0.000000,0.000000,0.000000}%
\pgfsetstrokecolor{currentstroke}%
\pgfsetdash{{9.600000pt}{2.400000pt}{1.500000pt}{2.400000pt}}{0.000000pt}%
\pgfpathmoveto{\pgfqpoint{0.055556in}{0.555556in}}%
\pgfpathlineto{\pgfqpoint{0.333333in}{0.555556in}}%
\pgfusepath{stroke}%
\end{pgfscope}%
\begin{pgfscope}%
\definecolor{textcolor}{rgb}{0.000000,0.000000,0.000000}%
\pgfsetstrokecolor{textcolor}%
\pgfsetfillcolor{textcolor}%
\pgftext[x=0.444444in,y=0.506944in,left,base]{\color{textcolor}\fontsize{10.000000}{12.000000}\selectfont \textsc{alpha} (\(\displaystyle k_\mathrm{co}=0\))   (0)}%
\end{pgfscope}%
\begin{pgfscope}%
\pgfsetbuttcap%
\pgfsetroundjoin%
\pgfsetlinewidth{1.505625pt}%
\definecolor{currentstroke}{rgb}{0.000000,0.000000,0.000000}%
\pgfsetstrokecolor{currentstroke}%
\pgfsetdash{{1.500000pt}{2.475000pt}}{0.000000pt}%
\pgfpathmoveto{\pgfqpoint{0.055556in}{0.347222in}}%
\pgfpathlineto{\pgfqpoint{0.333333in}{0.347222in}}%
\pgfusepath{stroke}%
\end{pgfscope}%
\begin{pgfscope}%
\definecolor{textcolor}{rgb}{0.000000,0.000000,0.000000}%
\pgfsetstrokecolor{textcolor}%
\pgfsetfillcolor{textcolor}%
\pgftext[x=0.444444in,y=0.298611in,left,base]{\color{textcolor}\fontsize{10.000000}{12.000000}\selectfont \textsc{alpha} (\(\displaystyle k_\mathrm{co}=\infty\))   (0)}%
\end{pgfscope}%
\begin{pgfscope}%
\pgfsetrectcap%
\pgfsetroundjoin%
\pgfsetlinewidth{1.505625pt}%
\definecolor{currentstroke}{rgb}{0.098039,0.098039,0.439216}%
\pgfsetstrokecolor{currentstroke}%
\pgfsetdash{}{0pt}%
\pgfpathmoveto{\pgfqpoint{0.055556in}{0.138889in}}%
\pgfpathlineto{\pgfqpoint{0.333333in}{0.138889in}}%
\pgfusepath{stroke}%
\end{pgfscope}%
\begin{pgfscope}%
\definecolor{textcolor}{rgb}{0.000000,0.000000,0.000000}%
\pgfsetstrokecolor{textcolor}%
\pgfsetfillcolor{textcolor}%
\pgftext[x=0.444444in,y=0.090278in,left,base]{\color{textcolor}\fontsize{10.000000}{12.000000}\selectfont \textsc{dlv2}   (24)}%
\end{pgfscope}%
\begin{pgfscope}%
\pgfsetrectcap%
\pgfsetroundjoin%
\pgfsetlinewidth{1.505625pt}%
\definecolor{currentstroke}{rgb}{0.000000,0.501961,0.000000}%
\pgfsetstrokecolor{currentstroke}%
\pgfsetdash{}{0pt}%
\pgfpathmoveto{\pgfqpoint{2.012698in}{0.555556in}}%
\pgfpathlineto{\pgfqpoint{2.290476in}{0.555556in}}%
\pgfusepath{stroke}%
\end{pgfscope}%
\begin{pgfscope}%
\definecolor{textcolor}{rgb}{0.000000,0.000000,0.000000}%
\pgfsetstrokecolor{textcolor}%
\pgfsetfillcolor{textcolor}%
\pgftext[x=2.401587in,y=0.506944in,left,base]{\color{textcolor}\fontsize{10.000000}{12.000000}\selectfont \textsc{h-alpha} (\(\displaystyle k_\mathrm{co}=0\))   (69)}%
\end{pgfscope}%
\begin{pgfscope}%
\pgfsetbuttcap%
\pgfsetroundjoin%
\pgfsetlinewidth{1.505625pt}%
\definecolor{currentstroke}{rgb}{0.000000,0.501961,0.000000}%
\pgfsetstrokecolor{currentstroke}%
\pgfsetdash{{5.550000pt}{2.400000pt}}{0.000000pt}%
\pgfpathmoveto{\pgfqpoint{2.012698in}{0.347222in}}%
\pgfpathlineto{\pgfqpoint{2.290476in}{0.347222in}}%
\pgfusepath{stroke}%
\end{pgfscope}%
\begin{pgfscope}%
\definecolor{textcolor}{rgb}{0.000000,0.000000,0.000000}%
\pgfsetstrokecolor{textcolor}%
\pgfsetfillcolor{textcolor}%
\pgftext[x=2.401587in,y=0.298611in,left,base]{\color{textcolor}\fontsize{10.000000}{12.000000}\selectfont \textsc{h-alpha} (\(\displaystyle k_\mathrm{co}=\infty\))   (23)}%
\end{pgfscope}%
\begin{pgfscope}%
\pgfsetbuttcap%
\pgfsetroundjoin%
\pgfsetlinewidth{1.505625pt}%
\definecolor{currentstroke}{rgb}{1.000000,0.000000,1.000000}%
\pgfsetstrokecolor{currentstroke}%
\pgfsetdash{{9.600000pt}{2.400000pt}{1.500000pt}{2.400000pt}}{0.000000pt}%
\pgfpathmoveto{\pgfqpoint{4.176938in}{0.555556in}}%
\pgfpathlineto{\pgfqpoint{4.454716in}{0.555556in}}%
\pgfusepath{stroke}%
\end{pgfscope}%
\begin{pgfscope}%
\definecolor{textcolor}{rgb}{0.000000,0.000000,0.000000}%
\pgfsetstrokecolor{textcolor}%
\pgfsetfillcolor{textcolor}%
\pgftext[x=4.565827in,y=0.506944in,left,base]{\color{textcolor}\fontsize{10.000000}{12.000000}\selectfont \textsc{h-alpha} (with \textsc{h-clingo} heuristics, \(\displaystyle k_\mathrm{co}=0\))   (30)}%
\end{pgfscope}%
\begin{pgfscope}%
\pgfsetbuttcap%
\pgfsetroundjoin%
\pgfsetlinewidth{1.505625pt}%
\definecolor{currentstroke}{rgb}{1.000000,0.000000,1.000000}%
\pgfsetstrokecolor{currentstroke}%
\pgfsetdash{{1.500000pt}{2.475000pt}}{0.000000pt}%
\pgfpathmoveto{\pgfqpoint{4.176938in}{0.347222in}}%
\pgfpathlineto{\pgfqpoint{4.454716in}{0.347222in}}%
\pgfusepath{stroke}%
\end{pgfscope}%
\begin{pgfscope}%
\definecolor{textcolor}{rgb}{0.000000,0.000000,0.000000}%
\pgfsetstrokecolor{textcolor}%
\pgfsetfillcolor{textcolor}%
\pgftext[x=4.565827in,y=0.298611in,left,base]{\color{textcolor}\fontsize{10.000000}{12.000000}\selectfont \textsc{h-alpha} (with \textsc{h-clingo} heuristics, \(\displaystyle k_\mathrm{co}=\infty\))   (19)}%
\end{pgfscope}%
\begin{pgfscope}%
\pgfsetrectcap%
\pgfsetroundjoin%
\pgfsetlinewidth{1.505625pt}%
\definecolor{currentstroke}{rgb}{1.000000,0.647059,0.000000}%
\pgfsetstrokecolor{currentstroke}%
\pgfsetdash{}{0pt}%
\pgfpathmoveto{\pgfqpoint{7.957149in}{0.555556in}}%
\pgfpathlineto{\pgfqpoint{8.234927in}{0.555556in}}%
\pgfusepath{stroke}%
\end{pgfscope}%
\begin{pgfscope}%
\definecolor{textcolor}{rgb}{0.000000,0.000000,0.000000}%
\pgfsetstrokecolor{textcolor}%
\pgfsetfillcolor{textcolor}%
\pgftext[x=8.346038in,y=0.506944in,left,base]{\color{textcolor}\fontsize{10.000000}{12.000000}\selectfont \textsc{clingo}   (37)}%
\end{pgfscope}%
\begin{pgfscope}%
\pgfsetbuttcap%
\pgfsetroundjoin%
\pgfsetlinewidth{1.505625pt}%
\definecolor{currentstroke}{rgb}{1.000000,0.647059,0.000000}%
\pgfsetstrokecolor{currentstroke}%
\pgfsetdash{{9.600000pt}{2.400000pt}{1.500000pt}{2.400000pt}}{0.000000pt}%
\pgfpathmoveto{\pgfqpoint{7.957149in}{0.347222in}}%
\pgfpathlineto{\pgfqpoint{8.234927in}{0.347222in}}%
\pgfusepath{stroke}%
\end{pgfscope}%
\begin{pgfscope}%
\definecolor{textcolor}{rgb}{0.000000,0.000000,0.000000}%
\pgfsetstrokecolor{textcolor}%
\pgfsetfillcolor{textcolor}%
\pgftext[x=8.346038in,y=0.298611in,left,base]{\color{textcolor}\fontsize{10.000000}{12.000000}\selectfont \textsc{h-clingo}   (46)}%
\end{pgfscope}%
\end{pgfpicture}%
\makeatother%
\endgroup%
}
			\caption{Solver configurations, with numbers of solved instances}
			\label{fig-cactus-house-legend}
		\end{subfigure}%
		\vspace{\floatsep}
		\begin{subfigure}{.32\textwidth}
			\centering
			\resizebox{\textwidth}{!}{\input{figures/cactus_house_g.pgf}}
			\caption{\rev{}{Number of guesses}}
			\label{fig-cactus-house-guesses}
		\end{subfigure}%
		\hspace*{\fill}
		\begin{subfigure}{.32\textwidth}
			\centering
			\resizebox{\textwidth}{!}{\input{figures/cactus_house_walltime.pgf}}
			\caption{\rev{Accumulated time }{Time} consumption}
			\label{fig-cactus-house-time}
		\end{subfigure}%
		\hspace*{\fill}
		\begin{subfigure}{.32\textwidth}
			\centering
			\resizebox{\textwidth}{!}{\input{figures/cactus_house_memory.pgf}}
			\caption{Memory consumption}
			\label{fig-cactus-house-memory}
		\end{subfigure}
		\caption{\rev{Time and memory }{Resource} consumption for solving each HRP instance}
		\label{fig-experimental-results-house}
	\end{figure}

\Cref{fig-experimental-results-house} shows performance data for experiments with HRP.
Cactus plots were created in the usual way.
In \cref{fig-cactus-house-time}, the x-axis gives the number of instances solved within real (i.e., wall-clock) time, given on the y-axis.
\rev{Time is accumulated over all solved instances.}{}
\rev{Memory consumption is given on the y-axis of \cref{fig-cactus-house-memory}, where data points are sorted by y-values, which are not accumulated. }{Similarly, \cref{fig-cactus-house-guesses} shows the number of guesses needed and \cref{fig-cactus-house-memory} shows the memory consumed to solve the instances.}
\rev{}{In all three plots, data points are sorted by y-values.}
\Cref{fig-cactus-house-legend} contains a legend with all solver configurations.
The number of instances solved by each system is shown next to its name (in parentheses).

One curve was drawn for each solver configuration:
\alphaslv\ without domain-specific heuristics, with strict ($k_\mathrm{co}=0$) and permissive ($k_\mathrm{co}=\infty$) grounding of constraints;
\alphaslv\ with domain-specific heuristics (\slv{h-alpha}), with strict and permissive grounding of constraints;
\rev{}{\alphaslv\ with \slv{h-clingo}-like domain-specific heuristics, with strict and permissive grounding of constraints;}
\rev{\slv{dlv2};}{}
\slv{clingo} with (\slv{h-clingo}) and without domain-specific heuristics\rev{.}{;}
\rev{}{and \slv{dlv2}.}

Substantial differences can be observed.
The curves for \slv{h-alpha} ($k_\mathrm{co}=0$) reach farthest to the right, meaning that \alphaslv\ with domain-specific heuristics solved the highest number of instances (\rev{59 }{69} out of 94) when grounding constraints strictly.
Surprisingly, with permissive grounding of constraints, \alphaslv\ with domain-specific heuristics exhibited relatively low time and space performance.

No curves are visible at all for \alphaslv\ without domain-specific heuristics because, in this configuration, the system could \rev{solve at most one instance only. }{not solve any instance.}
The other solvers' performance was somewhere in between the \alphaslv\ configurations at both ends of the spectrum.
Notably, \slv{h-clingo} with domain-specific heuristics solved more instances in less time compared to \slv{clingo} without domain-specific heuristics\rev{, but consumed slightly more memory}{}.
The largest instance solved by \slv{h-alpha} contained \rev{600 }{675} things, which is \rev{over 30\% }{almost 50\%} more than the size of the largest instance solved by \slv{h-clingo} \rev{(455)}{(456)}.
Recall that the time limit for solving each instance was \rev{10 }{15} minutes.

\rev{}{
	While \cref{fig-cactus-house-time} only reports overall solving time including grounding efforts, the distribution between grounding and solving time consumed by the ground-and-solve system \slv{clingo} has also been analysed.
	For each of the 37 instances solved by \slv{clingo}, the system spent between 27\% and 84\% of total solving time in grounding, on average 63\%.
	When using domain-specific heuristics, \slv{h-clingo} solved 46 instances and spent between 92\% and 97\% in grounding, on average 96\% of overall time.
	\slv{dlv2} solved 24 instances and spent between 5\% and 22\% of total solving time in grounding, on average 10\%.
}

\subsection{Case Study 2: The Partner Units Problem (PUP)}
\label{sec-applications-pup}

Like HRP, the Partner Units Problem (PUP) \cite{teppan2016solving,DBLP:journals/jcss/TeppanFG16} is an abstracted version of industrial (re)configuration problems. \rev{}{In particular, PUP deals with the configuration of parts of railway safety systems, where the development of domain-specific heuristics was difficult.}

\subsubsection{Problem Definition}

\begin{definition}[PUP]
	\label{def-pup}
	The input to the \emph{Partner Units Problem (PUP)} is given by a set of units $U$ and a bipartite graph $G=(S,Z,E)$, where $S$ is a set of sensors, $Z$ is a set of zones, and $E$ is a relation between $S$ and $Z$.
	
	The task is to find a partition of vertices $v \in S\cup Z$ into bags $u_i \in U$ such that for each bag the following requirements hold:
	\begin{enumerate*}[label=(\arabic*)]
		\item the bag contains at most $\mi{UCAP}$ vertices from $S$ and at most $\mi{UCAP}$
		vertices from $Z$; and
		\item the bag has at most $\mi{IUCAP}$ adjacent bags, where the bags $u_1$ and $u_2$ are adjacent whenever $v_i \in u_1$ and $v_j \in u_2$ for some $(v_i, v_j) \in E$.
	\end{enumerate*}
\end{definition}

\begin{figure}
	\centering
%
%
%
%

\tikzstyle{sensor} = [draw, circle, inner sep=0em, minimum width=1.5em]
\tikzstyle{zone} =   [draw, rectangle, inner sep=.2em, minimum width=2em, rounded corners=0.2em]
\tikzstyle{unit} =   [draw, rectangle, inner sep=.2em, minimum width=2em]

\begin{minipage}{.5\textwidth}
\begin{tikzpicture}[x=1.7em,y=1.7em,baseline=0pt]
\node[sensor] (s1) at (0.0,1.25) {$\mathrm{s_1}$};
\node[sensor] (s2) at (1.5,1.25) {$\mathrm{s_2}$};
\node[sensor] (s3) at (3.0,1.25) {$\mathrm{s_3}$};
\node[sensor] (s4) at (4.5,1.25) {$\mathrm{s_4}$};
\node[sensor] (s5) at (6.0,1.25) {$\mathrm{s_5}$};
\node[sensor] (s6) at (7.5,1.25) {$\mathrm{s_6}$};

\node[zone] (z1)   at (0.0,-1.25) {$\mathrm{z_{1}}$};
\node[zone] (z123) at (1.5,-1.25) {$\mathrm{z_{123}}$};
\node[zone] (z24)  at (3.0,-1.25) {$\mathrm{z_{24}}$};
\node[zone] (z35)  at (4.5,-1.25) {$\mathrm{z_{35}}$};
\node[zone] (z456) at (6.0,-1.25) {$\mathrm{z_{456}}$};
\node[zone] (z6)   at (7.5,-1.25) {$\mathrm{z_{6}}$};

\node[unit] (u1)   at (10,1.5)  {$\mathrm{u_{1}}$};
\node[unit] (u2)   at (10,.75)     {$\mathrm{u_{2}}$};
\node[unit] (u3)   at (10,0) {$\mathrm{u_{3}}$};
\node (u4)   at (10,-.75) {\small$\ucap=2$};
\node (u5)   at (10,-1.5) {\small$\iucap=2$};

\draw 
(z1)   -- (s1)
(z35)  -- (s3)
(z35)  -- (s5)
(z456) -- (s4)
(z456) -- (s5)
(z456) -- (s6)
(z123) -- (s1)
(z123) -- (s2)
(z123) -- (s3)
(z6)   -- (s6)
(z24)  -- (s2)
(z24)  -- (s4);
\end{tikzpicture}%
\end{minipage}%
\hfill\vline\hfill
\begin{minipage}{.4\textwidth}
\begin{tikzpicture}[x=1.7em,y=1.7em,baseline=0pt]
\node[sensor] (s1) at (0.0,1.4) {$\mathrm{s_1}$};
\node[sensor] (s2) at (1.5,1.4) {$\mathrm{s_2}$};
\node[sensor] (s3) at (3.0,1.4) {$\mathrm{s_3}$};
\node[sensor] (s4) at (4.5,1.4) {$\mathrm{s_4}$};
\node[sensor] (s5) at (6.0,1.4) {$\mathrm{s_5}$};
\node[sensor] (s6) at (7.5,1.4) {$\mathrm{s_6}$};

\node[zone] (z1)   at (0.0,-1.4) {$\mathrm{z_{1}}$};
\node[zone] (z123) at (1.5,-1.4) {$\mathrm{z_{123}}$};
\node[zone] (z24)  at (3.0,-1.4) {$\mathrm{z_{24}}$};
\node[zone] (z35)  at (4.5,-1.4) {$\mathrm{z_{35}}$};
\node[zone] (z456) at (6.0,-1.4) {$\mathrm{z_{456}}$};
\node[zone] (z6)   at (7.5,-1.4) {$\mathrm{z_{6}}$};

\node[unit] (u1) at (1.50, 0) {$\mathrm{u_1}$};
\node[unit] (u2) at (3.75, 0) {$\mathrm{u_2}$};
\node[unit] (u3) at (6.00, 0) {$\mathrm{u_3}$};

\draw 
(z1)    -- (u1) -- (s1)
(z123)  -- (u1) -- (s2)
(z24)   -- (u2) -- (s3)
(z35)   -- (u2) -- (s4)
(z6)    -- (u3) -- (s5)
(z456)  -- (u3) -- (s6);

\draw (u1) -- (u2) -- (u3);
\end{tikzpicture}
\end{minipage}
	\caption{Sample PUP instance (left) and one of its solutions (right)}
	\label{fig:pup_example}
\end{figure}
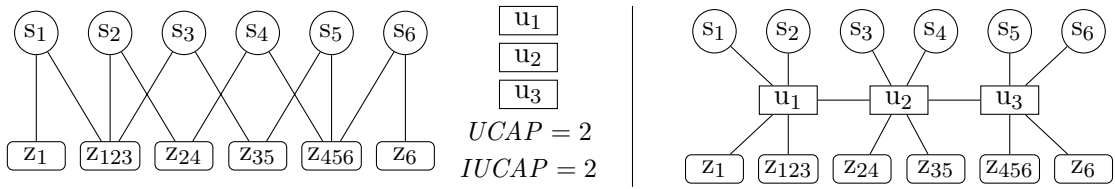

\Cref{fig:pup_example} shows an example of a PUP instance.
The bipartite graph comprises six sensors and six zones.
Each of the three units can be adjacent to at most two other units, and each unit can contain at most two sensors and two zones. 
An assignment of sensors and zones to units that satisfies all PUP requirements is also presented in \cref{fig:pup_example}.

\subsubsection{Encodings and Instances}
\label{sec-pup-encodings}

PUP instances consist of atoms over the predicates
\rev{$\mathrm{comUnit}/1$ (specifying units $U$) and $\mathrm{zone2sensor}/2$ (specifying the zone-to-sensor relation $E$). }
{
\begin{itemize}
	\item $\mathrm{comUnit}/1$ (specifying units $U$) and
	\item $\mathrm{zone2sensor}/2$ (specifying the zone-to-sensor relation $E$).\footnote{In the instances used for our experiments, both $\ucap$ and $\iucap$ are fixed at the value 2.}
\end{itemize}
}

\rev{
Our encoding for PUP is based on encodings from the ASP Competitions \cite{DBLP:journals/tplp/CalimeriIR14,DBLP:journals/ai/CalimeriGMR16}.\footnote{\url{https://www.mat.unical.it/aspcomp2011/files/PartnerUnitsPolynomial/partner_units_polynomial.enc.asp}, \url{https://www.mat.unical.it/aspcomp2014/#Participants.2C_Encodings.2C_Instance_Sets}}
The following rules constitute the main guessing part of the encoding:
\begin{align*}
	\mathrm{elem(z,Z)} &\leftarrow \mathrm{zone2sensor(Z,D)}.	\\
	\mathrm{elem(s,D)} &\leftarrow \mathrm{zone2sensor(Z,D)}.	\\
	\{~ \mathrm{assign(U,T,X)} ~\} &\leftarrow \mathrm{elem(T,X)}, \mathrm{comUnit(U)}.
\end{align*}
}{}

\rev{}{
	We used two different PUP encodings.
	The most efficient ASP encoding available for PUP is the \enquote{new} encoding from the Fifth ASP competition\footnote{\url{https://www.mat.unical.it/aspcomp2014/#Participants.2C_Encodings.2C_Instance_Sets}} \cite{DBLP:journals/ai/CalimeriGMR16}.
	The following rules constitute the main guessing part of the encoding:\footnote{Symbols like $D$ for \enquote{Door} have been replaced by $S$ for \enquote{Sensor} to be consistent with \cref{def-pup}.}
	\begin{align*}
		\mathrm{elem(z,Z)} &\leftarrow \mathrm{zone2sensor(Z,S)}.	\\
		\mathrm{elem(s,S)} &\leftarrow \mathrm{zone2sensor(Z,S)}.	\\
		\{~ \mathrm{gt(A,X,U)} ~\} &\leftarrow \mathrm{elem(A,X)}, \mathrm{comUnit(U)}, \mathrm{comUnit(U1)}, \mathrm{U1}=\mathrm{U}+1, \mathrm{rule(A)}, \mathrm{U} < \mathrm{X}.	\\
		\{~ \mathrm{gt(A,X,U)} ~\} &\leftarrow \mathrm{elem(A,X)}, \mathrm{comUnit(U)}, \mathrm{comUnit(U1)}, \mathrm{U1}=\mathrm{U}+1, \naf \mathrm{rule(A)}.
	\end{align*}
	Other rules are used to derive $\mathrm{unit2zone/2}$ and $\mathrm{unit2sensor/2}$ under very specific conditions from $\mathrm{gt/3}$.

	This encoding is written in such a way that heuristics are encoded in the form of rules and symmetry-breaking constraints, forcing the solver into parts of the search space where a solution can be expected.
	This goes at the cost of readability.
	Thus, the encoding does not lend itself well to extension by heuristic directives interacting with the choice rules.
	
	For this reason, a new encoding has been created for \alphaslv\ and extended by heuristic directives.
	Since also in this encoding parts of the heuristics are encoded as rules, the encoding is introduced in the next section.
}

\rev{}{
	All 18 instances from the ASP competitions \cite{DBLP:journals/tplp/CalimeriIR14,DBLP:journals/ai/CalimeriGMR16} that are satisfiable, belong to the subclass of PUP instances that can be polynomially decided, and in which the input graphs are connected have been used for our experiments.
	Unsatisfiable and non-connected instances have not been used because our heuristics are tailored towards efficiently generating valid configurations for connected instances.
}

\rev{Large }{Additionally, 100} instances were generated that \rev{}{exhibit the grounding bottleneck but} are easy to solve when using dedicated heuristics.
\rev{In size, these instances range from 3 to 300 units.}{%
	Our motivation for using these instances is that problems that humans can solve easily should also be easy for ASP solvers.
	The generated instances share the characteristics that their sensor-zone graphs are acyclic and that each pair of neighbouring zones shares one sensor.
	
	The newly generated instances are significantly larger than the competition instances.
	The largest generated instance contains 300 units, 300 zones, and 597 sensors.
	In contrast, the largest competition instance under consideration contains 40 units, 60 zones, and 79 sensors.
}

\subsubsection{Heuristics}
\label{sec-pup-heuristics}

QuickPup is a heuristic for PUP that successfully solves many hard problem instances \cite{DBLP:conf/iaai/TeppanFF12}.
Our approach supports implementing large parts of the originally procedural algorithm for QuickPup.
Our encoding uses rules by \citeA{teppan2016solving} and \citeA{DBLP:conf/ecai/TeppanF16} to derive a topological order of the zones and sensors.
Heuristic directives subsequently use this topological order.

First, a start zone is determined and denoted by $\mathrm{startZone}/1$.
In our encoding, the start zone is always the first one\rev{ and instances are designed so that solutions can easily be found when starting with this zone}{}.
QuickPup should actually try to use each zone as the start zone one after the other and abort search after a certain amount of time has passed.
This part of the algorithm cannot currently be represented in our framework.

QuickPup assigns zones and sensors to units in a breadth-first-order, called \enquote{topological order} because the graph is traversed level by level.
First, the start zone is assigned, then the sensors connected to the start zone, then the zones connected to those sensors and so on.
A helper predicate $\mathrm{layer}/3$ is introduced to compute the topological order.
In an atom $\mathrm{layer(T,X,L)}$, $T$ denotes the type of element (\enquote{s} for sensor and \enquote{z} for zone), $X$ is the element's identifier, and $L$ is its layer in the computed breadth-first order.

\rev{}{
	While experimenting with different encodings and heuristic directives, we found out that \alphaslv\ greatly profits from encodings where the search space is expanded step-by-step according to this topological order, instead of making all choice points available to the solver at once.
	Therefore, in the encoding used with \alphaslv\ in our final experiments, the topological order is represented by atoms of the $\mathrm{elem\_n/3}$ predicate, e.g., $\mathrm{elem\_n(T,X,N)}$ means that element $(T,X)$ has position $N$ in the topological order.
	Furthermore, $\mathrm{elems\_assigned\_up\_to/1}$ denotes how many elements are already assigned to units, e.g., $\mathrm{elems\_assigned\_up\_to(N)}$ means that the elements with positions 1 up to $N$ in the topological order have already been assigned.
	The choice of units is also restricted:
	Since assigning an element to a previously unused unit does not need to differentiate between individual unused units, each unit only becomes assignable when the unit with the next-lower number is already used.
	\begin{align*}
		\mathrm{elem(z,Z)} \leftarrow\; &\mathrm{zone2sensor(Z,S)}.	\\
		\mathrm{elem(s,S)} \leftarrow\; &\mathrm{zone2sensor(Z,S)}.	\\
		\mathrm{assignable\_unit(U)} \leftarrow\; &\mathrm{comUnit(U)}, \mathrm{used}(\mathrm{U}{-}1).	\\
		\mathrm{assignable(U,T,X)} \leftarrow\; &\mathrm{assignable\_unit(U)}, \mathrm{elem(T,X)}, \mathrm{elem\_n(T,X,N)},	\\
			&\mathrm{elems\_assigned\_up\_to}(\mathrm{N}{-}1).	\\
		\{~ \mathrm{assign(U,T,X)} ~\} \leftarrow\; &\mathrm{assignable(U,T,X)}.
	\end{align*}
}

\rev{
A realisation of QuickPup in our framework requires several heuristic directives.
These directives use the level term in the annotation to process the zones and sensors according to the topological order. The weight term in the annotation is used to assign the units in the right order.
}{
	Due to these rules, elements are processed one after the other and can be assigned to any of the units already used or to the unused unit with the lowest number.
	The selection among these units is guided by heuristic directives, which encode parts of QuickPup.
	We tried many different representations of (parts of) QuickPup as heuristic directives and found out that the overhead caused by a fully faithful representation in \alphaslv\ outweighs its benefits.
	The resulting directives presented below are therefore a compromise between faithfulness to the original heuristics and computational efficiency.
}%
The directives use some intermediate predicates whose meaning should become evident from their names.
The full encoding is available online.\footnote{\rev{}{\url{https://ainf.aau.at/dynacon}}}

\rev{
QuickPup first tries to assign an element to the first unit:
\begin{align*}
	\heudirstmt\ \mathrm{assign(U,T,X)} :\; &\mathrm{comUnit(U)}, \mathrm{elem(T,X)},\\
				&\mathrm{maxLayer(ML)}, \mathrm{layer(T,X,L)},\\
				&\naf \mathrm{full(U,T)}, \naf \mathrm{assigned(T,X)}, \naf \heusignT\ \mathrm{used(U)},\\
				&\mathrm{nUnits(NU)}, \mathrm{U}=1. \; [ \mathrm{NU}@(\mathrm{ML}{-}\mathrm{L}) ]
\end{align*}

If one unit cannot be assigned, QuickPup tries preceding units in decreasing order next:
\begin{align*}
	\heudirstmt\ \mathrm{assign(U,T,X)} :\; &\mathrm{comUnit(U)}, \mathrm{elem(T,X)},\\
	&\mathrm{maxLayer(ML)}, \mathrm{layer(T,X,L)},\\
	&\naf \mathrm{full(U,T)}, \naf \mathrm{assigned(T,X)},\\
	&\heusignT\ \mathrm{used(U)}. \; [ \mathrm{U}@(\mathrm{ML}{-}\mathrm{L}) ]
\end{align*}
}{}

\rev{}{
	First, already used units are tried in decreasing order:
	\begin{align*}
		\heudirstmt\ \mathrm{assign(U,T,X)} :\; &\mathrm{assignable(U,T,X)}, \mathrm{maxLayer(ML)}, \mathrm{layer(T,X,L)},\\
				&\naf \mathrm{full(U,T)}, \naf \heusignT\ \mathrm{assigned(T,X)},\\
				&\heusignT\ \mathrm{used(U)}. \; [ \mathrm{U}@(\mathrm{ML}{-}\mathrm{L}) ]
	\end{align*}
}

A fresh unit is only touched \rev{after all preceding units have been tried: }{if assigning it to the preceding unit has failed:}
\begin{align*}
	\heudirstmt\ \mathrm{assign(U,T,X)} :\; &\rev{\mathrm{comUnit(U)}, \mathrm{elem(T,X)}, }{\mathrm{assignable(U,T,X)},}\\
	&\mathrm{maxLayer(ML)}, \mathrm{layer(T,X,L)},\\
	&\naf \mathrm{full(U,T)}, \naf \rev{}{\heusignT\ } \mathrm{assigned(T,X)},\\
	&\naf \heusignT\ \mathrm{used(U)}, \mathrm{comUnit(U{-}1)}, \heusignT\ \mathrm{used(U{-}1)},\\
	&\heusignF\ \mathrm{assign(U{-}1,T,X)}. \; [ \mathrm{U}@(\mathrm{ML}{-}\mathrm{L}) ]
\end{align*}

Note the condition $\heusignF\ \mathrm{assign(U{-}1,T,X)}$ in the last heuristic directive.
Due to this condition, the heuristic is only applicable if the same element could not be assigned to the preceding unit $U{-}1$.
This situation may be caused by backtracking or by the following heuristic avoiding assignments to units that are already full:
\begin{align*}
	\heudirstmt\ \heusignF\ \mathrm{assign(U,T,X)} :\; &\rev{\mathrm{comUnit(U)}, \mathrm{elem(T,X)}, }{\mathrm{assignable(U,T,X)},}\\
	&\mathrm{maxLayer(ML)}, \mathrm{full(U,T)},\\
	&\naf \mathrm{assign(U,T,X)}. \; [ 1@\mathrm{ML} ]
\end{align*}

Choice points not assigned by any of these heuristics are finally assigned false by a dedicated heuristic directive, similarly as shown for HRP in \cref{sec-applications-hrp-heuristics}.

The heuristics we created for \alphaslv\ cannot be used with \slv{clingo} due to the usage of $\heusignT$, $\heusignF$, and default negation.
An alternative encoding containing heuristic directives for \slv{clingo} has been created
\rev{that is similar to the QuickPup*-like heuristics created for \alphaslv. }{in a similar way as for HRP.}
\rev{}{
	Furthermore, an \alphaslv\ encoding with heuristics without our novel features, compliant with those employed by \slv{clingo}, has been created.
}

\subsubsection{Results}
\label{sec-applications-pup-results}

	\begin{figure}[t]
		\centering
		\begin{subfigure}{\textwidth}
			\centering
			\resizebox{\textwidth}{!}{
\begingroup%
\makeatletter%
\begin{pgfpicture}%
\pgfpathrectangle{\pgfpointorigin}{\pgfqpoint{9.290326in}{0.666667in}}%
\pgfusepath{use as bounding box, clip}%
\begin{pgfscope}%
\pgfsetbuttcap%
\pgfsetmiterjoin%
\definecolor{currentfill}{rgb}{1.000000,1.000000,1.000000}%
\pgfsetfillcolor{currentfill}%
\pgfsetlinewidth{0.000000pt}%
\definecolor{currentstroke}{rgb}{1.000000,1.000000,1.000000}%
\pgfsetstrokecolor{currentstroke}%
\pgfsetdash{}{0pt}%
\pgfpathmoveto{\pgfqpoint{0.000000in}{0.000000in}}%
\pgfpathlineto{\pgfqpoint{9.290326in}{0.000000in}}%
\pgfpathlineto{\pgfqpoint{9.290326in}{0.666667in}}%
\pgfpathlineto{\pgfqpoint{0.000000in}{0.666667in}}%
\pgfpathclose%
\pgfusepath{fill}%
\end{pgfscope}%
\begin{pgfscope}%
\pgfsetbuttcap%
\pgfsetmiterjoin%
\definecolor{currentfill}{rgb}{1.000000,1.000000,1.000000}%
\pgfsetfillcolor{currentfill}%
\pgfsetfillopacity{0.800000}%
\pgfsetlinewidth{1.003750pt}%
\definecolor{currentstroke}{rgb}{0.800000,0.800000,0.800000}%
\pgfsetstrokecolor{currentstroke}%
\pgfsetstrokeopacity{0.800000}%
\pgfsetdash{}{0pt}%
\pgfpathmoveto{\pgfqpoint{0.027778in}{0.000000in}}%
\pgfpathlineto{\pgfqpoint{9.262548in}{0.000000in}}%
\pgfpathquadraticcurveto{\pgfqpoint{9.290326in}{0.000000in}}{\pgfqpoint{9.290326in}{0.027778in}}%
\pgfpathlineto{\pgfqpoint{9.290326in}{0.638889in}}%
\pgfpathquadraticcurveto{\pgfqpoint{9.290326in}{0.666667in}}{\pgfqpoint{9.262548in}{0.666667in}}%
\pgfpathlineto{\pgfqpoint{0.027778in}{0.666667in}}%
\pgfpathquadraticcurveto{\pgfqpoint{-0.000000in}{0.666667in}}{\pgfqpoint{-0.000000in}{0.638889in}}%
\pgfpathlineto{\pgfqpoint{-0.000000in}{0.027778in}}%
\pgfpathquadraticcurveto{\pgfqpoint{-0.000000in}{0.000000in}}{\pgfqpoint{0.027778in}{0.000000in}}%
\pgfpathclose%
\pgfusepath{stroke,fill}%
\end{pgfscope}%
\begin{pgfscope}%
\pgfsetbuttcap%
\pgfsetroundjoin%
\pgfsetlinewidth{1.505625pt}%
\definecolor{currentstroke}{rgb}{0.000000,0.000000,0.000000}%
\pgfsetstrokecolor{currentstroke}%
\pgfsetdash{{9.600000pt}{2.400000pt}{1.500000pt}{2.400000pt}}{0.000000pt}%
\pgfpathmoveto{\pgfqpoint{0.055556in}{0.555556in}}%
\pgfpathlineto{\pgfqpoint{0.333333in}{0.555556in}}%
\pgfusepath{stroke}%
\end{pgfscope}%
\begin{pgfscope}%
\definecolor{textcolor}{rgb}{0.000000,0.000000,0.000000}%
\pgfsetstrokecolor{textcolor}%
\pgfsetfillcolor{textcolor}%
\pgftext[x=0.444444in,y=0.506944in,left,base]{\color{textcolor}\fontsize{10.000000}{12.000000}\selectfont \textsc{alpha} (\(\displaystyle k_\mathrm{co}=0\))   (7)}%
\end{pgfscope}%
\begin{pgfscope}%
\pgfsetbuttcap%
\pgfsetroundjoin%
\pgfsetlinewidth{1.505625pt}%
\definecolor{currentstroke}{rgb}{0.000000,0.000000,0.000000}%
\pgfsetstrokecolor{currentstroke}%
\pgfsetdash{{1.500000pt}{2.475000pt}}{0.000000pt}%
\pgfpathmoveto{\pgfqpoint{0.055556in}{0.347222in}}%
\pgfpathlineto{\pgfqpoint{0.333333in}{0.347222in}}%
\pgfusepath{stroke}%
\end{pgfscope}%
\begin{pgfscope}%
\definecolor{textcolor}{rgb}{0.000000,0.000000,0.000000}%
\pgfsetstrokecolor{textcolor}%
\pgfsetfillcolor{textcolor}%
\pgftext[x=0.444444in,y=0.298611in,left,base]{\color{textcolor}\fontsize{10.000000}{12.000000}\selectfont \textsc{alpha} (\(\displaystyle k_\mathrm{co}=\infty\))   (9)}%
\end{pgfscope}%
\begin{pgfscope}%
\pgfsetrectcap%
\pgfsetroundjoin%
\pgfsetlinewidth{1.505625pt}%
\definecolor{currentstroke}{rgb}{0.098039,0.098039,0.439216}%
\pgfsetstrokecolor{currentstroke}%
\pgfsetdash{}{0pt}%
\pgfpathmoveto{\pgfqpoint{0.055556in}{0.138889in}}%
\pgfpathlineto{\pgfqpoint{0.333333in}{0.138889in}}%
\pgfusepath{stroke}%
\end{pgfscope}%
\begin{pgfscope}%
\definecolor{textcolor}{rgb}{0.000000,0.000000,0.000000}%
\pgfsetstrokecolor{textcolor}%
\pgfsetfillcolor{textcolor}%
\pgftext[x=0.444444in,y=0.090278in,left,base]{\color{textcolor}\fontsize{10.000000}{12.000000}\selectfont \textsc{dlv2}   (10)}%
\end{pgfscope}%
\begin{pgfscope}%
\pgfsetrectcap%
\pgfsetroundjoin%
\pgfsetlinewidth{1.505625pt}%
\definecolor{currentstroke}{rgb}{0.000000,0.501961,0.000000}%
\pgfsetstrokecolor{currentstroke}%
\pgfsetdash{}{0pt}%
\pgfpathmoveto{\pgfqpoint{2.012698in}{0.555556in}}%
\pgfpathlineto{\pgfqpoint{2.290476in}{0.555556in}}%
\pgfusepath{stroke}%
\end{pgfscope}%
\begin{pgfscope}%
\definecolor{textcolor}{rgb}{0.000000,0.000000,0.000000}%
\pgfsetstrokecolor{textcolor}%
\pgfsetfillcolor{textcolor}%
\pgftext[x=2.401587in,y=0.506944in,left,base]{\color{textcolor}\fontsize{10.000000}{12.000000}\selectfont \textsc{h-alpha} (\(\displaystyle k_\mathrm{co}=0\))   (100)}%
\end{pgfscope}%
\begin{pgfscope}%
\pgfsetbuttcap%
\pgfsetroundjoin%
\pgfsetlinewidth{1.505625pt}%
\definecolor{currentstroke}{rgb}{0.000000,0.501961,0.000000}%
\pgfsetstrokecolor{currentstroke}%
\pgfsetdash{{5.550000pt}{2.400000pt}}{0.000000pt}%
\pgfpathmoveto{\pgfqpoint{2.012698in}{0.347222in}}%
\pgfpathlineto{\pgfqpoint{2.290476in}{0.347222in}}%
\pgfusepath{stroke}%
\end{pgfscope}%
\begin{pgfscope}%
\definecolor{textcolor}{rgb}{0.000000,0.000000,0.000000}%
\pgfsetstrokecolor{textcolor}%
\pgfsetfillcolor{textcolor}%
\pgftext[x=2.401587in,y=0.298611in,left,base]{\color{textcolor}\fontsize{10.000000}{12.000000}\selectfont \textsc{h-alpha} (\(\displaystyle k_\mathrm{co}=\infty\))   (74)}%
\end{pgfscope}%
\begin{pgfscope}%
\pgfsetbuttcap%
\pgfsetroundjoin%
\pgfsetlinewidth{1.505625pt}%
\definecolor{currentstroke}{rgb}{1.000000,0.000000,1.000000}%
\pgfsetstrokecolor{currentstroke}%
\pgfsetdash{{9.600000pt}{2.400000pt}{1.500000pt}{2.400000pt}}{0.000000pt}%
\pgfpathmoveto{\pgfqpoint{4.176939in}{0.555556in}}%
\pgfpathlineto{\pgfqpoint{4.454716in}{0.555556in}}%
\pgfusepath{stroke}%
\end{pgfscope}%
\begin{pgfscope}%
\definecolor{textcolor}{rgb}{0.000000,0.000000,0.000000}%
\pgfsetstrokecolor{textcolor}%
\pgfsetfillcolor{textcolor}%
\pgftext[x=4.565827in,y=0.506944in,left,base]{\color{textcolor}\fontsize{10.000000}{12.000000}\selectfont \textsc{h-alpha} (with \textsc{h-clingo} heuristics, \(\displaystyle k_\mathrm{co}=0\))   (46)}%
\end{pgfscope}%
\begin{pgfscope}%
\pgfsetbuttcap%
\pgfsetroundjoin%
\pgfsetlinewidth{1.505625pt}%
\definecolor{currentstroke}{rgb}{1.000000,0.000000,1.000000}%
\pgfsetstrokecolor{currentstroke}%
\pgfsetdash{{1.500000pt}{2.475000pt}}{0.000000pt}%
\pgfpathmoveto{\pgfqpoint{4.176939in}{0.347222in}}%
\pgfpathlineto{\pgfqpoint{4.454716in}{0.347222in}}%
\pgfusepath{stroke}%
\end{pgfscope}%
\begin{pgfscope}%
\definecolor{textcolor}{rgb}{0.000000,0.000000,0.000000}%
\pgfsetstrokecolor{textcolor}%
\pgfsetfillcolor{textcolor}%
\pgftext[x=4.565827in,y=0.298611in,left,base]{\color{textcolor}\fontsize{10.000000}{12.000000}\selectfont \textsc{h-alpha} (with \textsc{h-clingo} heuristics, \(\displaystyle k_\mathrm{co}=\infty\))   (65)}%
\end{pgfscope}%
\begin{pgfscope}%
\pgfsetrectcap%
\pgfsetroundjoin%
\pgfsetlinewidth{1.505625pt}%
\definecolor{currentstroke}{rgb}{1.000000,0.647059,0.000000}%
\pgfsetstrokecolor{currentstroke}%
\pgfsetdash{}{0pt}%
\pgfpathmoveto{\pgfqpoint{7.957150in}{0.555556in}}%
\pgfpathlineto{\pgfqpoint{8.234927in}{0.555556in}}%
\pgfusepath{stroke}%
\end{pgfscope}%
\begin{pgfscope}%
\definecolor{textcolor}{rgb}{0.000000,0.000000,0.000000}%
\pgfsetstrokecolor{textcolor}%
\pgfsetfillcolor{textcolor}%
\pgftext[x=8.346038in,y=0.506944in,left,base]{\color{textcolor}\fontsize{10.000000}{12.000000}\selectfont \textsc{clingo}   (13)}%
\end{pgfscope}%
\begin{pgfscope}%
\pgfsetbuttcap%
\pgfsetroundjoin%
\pgfsetlinewidth{1.505625pt}%
\definecolor{currentstroke}{rgb}{1.000000,0.647059,0.000000}%
\pgfsetstrokecolor{currentstroke}%
\pgfsetdash{{9.600000pt}{2.400000pt}{1.500000pt}{2.400000pt}}{0.000000pt}%
\pgfpathmoveto{\pgfqpoint{7.957150in}{0.347222in}}%
\pgfpathlineto{\pgfqpoint{8.234927in}{0.347222in}}%
\pgfusepath{stroke}%
\end{pgfscope}%
\begin{pgfscope}%
\definecolor{textcolor}{rgb}{0.000000,0.000000,0.000000}%
\pgfsetstrokecolor{textcolor}%
\pgfsetfillcolor{textcolor}%
\pgftext[x=8.346038in,y=0.298611in,left,base]{\color{textcolor}\fontsize{10.000000}{12.000000}\selectfont \textsc{h-clingo}   (23)}%
\end{pgfscope}%
\end{pgfpicture}%
\makeatother%
\endgroup%
}
			\caption{Solver configurations, with numbers of solved instances}
			\label{fig-cactus-pup-simple-legend}
		\end{subfigure}%
		\vspace{\floatsep}
		\begin{subfigure}{.32\textwidth}
			\centering
			\resizebox{\textwidth}{!}{\input{figures/cactus_pup-simple_g.pgf}}
			\caption{\rev{}{Number of guesses}}
			\label{fig-cactus-pup-simple-guesses}
		\end{subfigure}%
		\hspace*{\fill}
		\begin{subfigure}{.32\textwidth}
			\centering
			\resizebox{\textwidth}{!}{\input{figures/cactus_pup-simple_walltime.pgf}}
			\caption{\rev{Accumulated time }{Time} consumption}
			\label{fig-cactus-pup-simple-time}
		\end{subfigure}%
		\hspace*{\fill}
		\begin{subfigure}{.32\textwidth}
			\centering
			\resizebox{\textwidth}{!}{\input{figures/cactus_pup-simple_memory.pgf}}
			\caption{Memory consumption}
			\label{fig-cactus-pup-simple-memory}
		\end{subfigure}
		\caption{\rev{Time and memory }{Resource} consumption for solving each simple PUP instance}
		\label{fig-experimental-results-pup-simple}
	\end{figure}

\rev{}{
	In the experiments, \alphaslv\ used the encoding presented in \cref{sec-pup-heuristics};
	\slv{h-alpha} additionally used the heuristics presented in \cref{sec-pup-heuristics};
	\alphaslv\ with \slv{h-clingo}-like domain-specific heuristics used the same encoding but different heuristics;
	\slv{h-clingo} also used the \alphaslv\ encoding but with different heuristics;
	and \slv{dlv2} and \slv{clingo} used the new ASP competition encoding mentioned in \cref{sec-pup-encodings}.
}

Cactus plots for PUP (\cref{fig-experimental-results-pup-simple,fig-experimental-results-pup}) were generated in the same way as for HRP (cf.\ \cref{sec-applications-hrp-results}).
\rev{}{
	\Cref{fig-experimental-results-pup-simple} shows results for the 100 newly generated instances and \cref{fig-experimental-results-pup} shows result for the 18 competition instances that are polynomially decidable, connected, and satisfiable.
}

\rev{Again, }{Looking at the generated instances (\cref{fig-experimental-results-pup-simple}),} \alphaslv\ with domain-specific heuristics \rev{}{again} solved the highest number of instances (all 100)\rev{. }{\ when grounding constraints strictly.}
\rev{The curves for \slv{h-alpha} ($k_\mathrm{co}=0$) and \slv{h-alpha} ($k_\mathrm{co}=\infty$) are almost indistinguishable, meaning that it did not make a difference in the PUP domain whether constraints were grounded strictly or permissively.}{}
\rev{}{
	When grounding constraints permissively, \alphaslv\ needed much fewer guesses, but consumed more time and memory.
}

On the other extreme, \alphaslv\ without domain-specific heuristic could solve only 7 of the 100 instances\rev{. }{\ when grounding constraints strictly, and 9 instances when grounding constraints permissively.}

The systems \slv{dlv2}, \slv{clingo}, and \slv{h-clingo} performed somewhere \rev{}{in} between those extremes.
\slv{h-clingo} with domain-specific heuristics solved many more instances than \slv{clingo} without domain-specific heuristics.

The largest instance in our instance set contained 300 units.
\slv{h-alpha} was able to solve all these instances.
In contrast, the size of the largest instance that could be solved by any other system, using the given encoding, was only \rev{105.}{75.}
Recall that the time limit for solving each instance was \rev{10 }{15} minutes.
For \rev{11 }{28} instances, \slv{h-clingo} returned an error (\enquote{\rev{}{Value too large for defined data type: }Id out of range}).

\rev{}{
	For each of the 13 instances solved by \slv{clingo}, the system spent between 0\% and 100\% of total solving time in grounding, on average 34\%.
	When using domain-specific heuristics, \slv{h-clingo} solved 23 instances and spent between 63\% and 100\% in grounding, on average 85\% of overall time.
	\slv{dlv2} solved 10 instances and spent between 0\% and 14\% of total solving time in grounding, on average 4\%.
}

	\begin{figure}[t]
		\centering
		\begin{subfigure}{\textwidth}
			\centering
			\resizebox{\textwidth}{!}{
\begingroup%
\makeatletter%
\begin{pgfpicture}%
\pgfpathrectangle{\pgfpointorigin}{\pgfqpoint{9.359771in}{0.666667in}}%
\pgfusepath{use as bounding box, clip}%
\begin{pgfscope}%
\pgfsetbuttcap%
\pgfsetmiterjoin%
\definecolor{currentfill}{rgb}{1.000000,1.000000,1.000000}%
\pgfsetfillcolor{currentfill}%
\pgfsetlinewidth{0.000000pt}%
\definecolor{currentstroke}{rgb}{1.000000,1.000000,1.000000}%
\pgfsetstrokecolor{currentstroke}%
\pgfsetdash{}{0pt}%
\pgfpathmoveto{\pgfqpoint{0.000000in}{0.000000in}}%
\pgfpathlineto{\pgfqpoint{9.359771in}{0.000000in}}%
\pgfpathlineto{\pgfqpoint{9.359771in}{0.666667in}}%
\pgfpathlineto{\pgfqpoint{0.000000in}{0.666667in}}%
\pgfpathclose%
\pgfusepath{fill}%
\end{pgfscope}%
\begin{pgfscope}%
\pgfsetbuttcap%
\pgfsetmiterjoin%
\definecolor{currentfill}{rgb}{1.000000,1.000000,1.000000}%
\pgfsetfillcolor{currentfill}%
\pgfsetfillopacity{0.800000}%
\pgfsetlinewidth{1.003750pt}%
\definecolor{currentstroke}{rgb}{0.800000,0.800000,0.800000}%
\pgfsetstrokecolor{currentstroke}%
\pgfsetstrokeopacity{0.800000}%
\pgfsetdash{}{0pt}%
\pgfpathmoveto{\pgfqpoint{0.027778in}{0.000000in}}%
\pgfpathlineto{\pgfqpoint{9.331993in}{0.000000in}}%
\pgfpathquadraticcurveto{\pgfqpoint{9.359771in}{0.000000in}}{\pgfqpoint{9.359771in}{0.027778in}}%
\pgfpathlineto{\pgfqpoint{9.359771in}{0.638889in}}%
\pgfpathquadraticcurveto{\pgfqpoint{9.359771in}{0.666667in}}{\pgfqpoint{9.331993in}{0.666667in}}%
\pgfpathlineto{\pgfqpoint{0.027778in}{0.666667in}}%
\pgfpathquadraticcurveto{\pgfqpoint{-0.000000in}{0.666667in}}{\pgfqpoint{-0.000000in}{0.638889in}}%
\pgfpathlineto{\pgfqpoint{-0.000000in}{0.027778in}}%
\pgfpathquadraticcurveto{\pgfqpoint{-0.000000in}{0.000000in}}{\pgfqpoint{0.027778in}{0.000000in}}%
\pgfpathclose%
\pgfusepath{stroke,fill}%
\end{pgfscope}%
\begin{pgfscope}%
\pgfsetbuttcap%
\pgfsetroundjoin%
\pgfsetlinewidth{1.505625pt}%
\definecolor{currentstroke}{rgb}{0.000000,0.000000,0.000000}%
\pgfsetstrokecolor{currentstroke}%
\pgfsetdash{{9.600000pt}{2.400000pt}{1.500000pt}{2.400000pt}}{0.000000pt}%
\pgfpathmoveto{\pgfqpoint{0.055556in}{0.555556in}}%
\pgfpathlineto{\pgfqpoint{0.333333in}{0.555556in}}%
\pgfusepath{stroke}%
\end{pgfscope}%
\begin{pgfscope}%
\definecolor{textcolor}{rgb}{0.000000,0.000000,0.000000}%
\pgfsetstrokecolor{textcolor}%
\pgfsetfillcolor{textcolor}%
\pgftext[x=0.444444in,y=0.506944in,left,base]{\color{textcolor}\fontsize{10.000000}{12.000000}\selectfont \textsc{alpha} (\(\displaystyle k_\mathrm{co}=0\))   (15)}%
\end{pgfscope}%
\begin{pgfscope}%
\pgfsetbuttcap%
\pgfsetroundjoin%
\pgfsetlinewidth{1.505625pt}%
\definecolor{currentstroke}{rgb}{0.000000,0.000000,0.000000}%
\pgfsetstrokecolor{currentstroke}%
\pgfsetdash{{1.500000pt}{2.475000pt}}{0.000000pt}%
\pgfpathmoveto{\pgfqpoint{0.055556in}{0.347222in}}%
\pgfpathlineto{\pgfqpoint{0.333333in}{0.347222in}}%
\pgfusepath{stroke}%
\end{pgfscope}%
\begin{pgfscope}%
\definecolor{textcolor}{rgb}{0.000000,0.000000,0.000000}%
\pgfsetstrokecolor{textcolor}%
\pgfsetfillcolor{textcolor}%
\pgftext[x=0.444444in,y=0.298611in,left,base]{\color{textcolor}\fontsize{10.000000}{12.000000}\selectfont \textsc{alpha} (\(\displaystyle k_\mathrm{co}=\infty\))   (15)}%
\end{pgfscope}%
\begin{pgfscope}%
\pgfsetrectcap%
\pgfsetroundjoin%
\pgfsetlinewidth{1.505625pt}%
\definecolor{currentstroke}{rgb}{0.098039,0.098039,0.439216}%
\pgfsetstrokecolor{currentstroke}%
\pgfsetdash{}{0pt}%
\pgfpathmoveto{\pgfqpoint{0.055556in}{0.138889in}}%
\pgfpathlineto{\pgfqpoint{0.333333in}{0.138889in}}%
\pgfusepath{stroke}%
\end{pgfscope}%
\begin{pgfscope}%
\definecolor{textcolor}{rgb}{0.000000,0.000000,0.000000}%
\pgfsetstrokecolor{textcolor}%
\pgfsetfillcolor{textcolor}%
\pgftext[x=0.444444in,y=0.090278in,left,base]{\color{textcolor}\fontsize{10.000000}{12.000000}\selectfont \textsc{dlv2}   (15)}%
\end{pgfscope}%
\begin{pgfscope}%
\pgfsetrectcap%
\pgfsetroundjoin%
\pgfsetlinewidth{1.505625pt}%
\definecolor{currentstroke}{rgb}{0.000000,0.501961,0.000000}%
\pgfsetstrokecolor{currentstroke}%
\pgfsetdash{}{0pt}%
\pgfpathmoveto{\pgfqpoint{2.082142in}{0.555556in}}%
\pgfpathlineto{\pgfqpoint{2.359920in}{0.555556in}}%
\pgfusepath{stroke}%
\end{pgfscope}%
\begin{pgfscope}%
\definecolor{textcolor}{rgb}{0.000000,0.000000,0.000000}%
\pgfsetstrokecolor{textcolor}%
\pgfsetfillcolor{textcolor}%
\pgftext[x=2.471031in,y=0.506944in,left,base]{\color{textcolor}\fontsize{10.000000}{12.000000}\selectfont \textsc{h-alpha} (\(\displaystyle k_\mathrm{co}=0\))   (17)}%
\end{pgfscope}%
\begin{pgfscope}%
\pgfsetbuttcap%
\pgfsetroundjoin%
\pgfsetlinewidth{1.505625pt}%
\definecolor{currentstroke}{rgb}{0.000000,0.501961,0.000000}%
\pgfsetstrokecolor{currentstroke}%
\pgfsetdash{{5.550000pt}{2.400000pt}}{0.000000pt}%
\pgfpathmoveto{\pgfqpoint{2.082142in}{0.347222in}}%
\pgfpathlineto{\pgfqpoint{2.359920in}{0.347222in}}%
\pgfusepath{stroke}%
\end{pgfscope}%
\begin{pgfscope}%
\definecolor{textcolor}{rgb}{0.000000,0.000000,0.000000}%
\pgfsetstrokecolor{textcolor}%
\pgfsetfillcolor{textcolor}%
\pgftext[x=2.471031in,y=0.298611in,left,base]{\color{textcolor}\fontsize{10.000000}{12.000000}\selectfont \textsc{h-alpha} (\(\displaystyle k_\mathrm{co}=\infty\))   (17)}%
\end{pgfscope}%
\begin{pgfscope}%
\pgfsetbuttcap%
\pgfsetroundjoin%
\pgfsetlinewidth{1.505625pt}%
\definecolor{currentstroke}{rgb}{1.000000,0.000000,1.000000}%
\pgfsetstrokecolor{currentstroke}%
\pgfsetdash{{9.600000pt}{2.400000pt}{1.500000pt}{2.400000pt}}{0.000000pt}%
\pgfpathmoveto{\pgfqpoint{4.246383in}{0.555556in}}%
\pgfpathlineto{\pgfqpoint{4.524161in}{0.555556in}}%
\pgfusepath{stroke}%
\end{pgfscope}%
\begin{pgfscope}%
\definecolor{textcolor}{rgb}{0.000000,0.000000,0.000000}%
\pgfsetstrokecolor{textcolor}%
\pgfsetfillcolor{textcolor}%
\pgftext[x=4.635272in,y=0.506944in,left,base]{\color{textcolor}\fontsize{10.000000}{12.000000}\selectfont \textsc{h-alpha} (with \textsc{h-clingo} heuristics, \(\displaystyle k_\mathrm{co}=0\))   (17)}%
\end{pgfscope}%
\begin{pgfscope}%
\pgfsetbuttcap%
\pgfsetroundjoin%
\pgfsetlinewidth{1.505625pt}%
\definecolor{currentstroke}{rgb}{1.000000,0.000000,1.000000}%
\pgfsetstrokecolor{currentstroke}%
\pgfsetdash{{1.500000pt}{2.475000pt}}{0.000000pt}%
\pgfpathmoveto{\pgfqpoint{4.246383in}{0.347222in}}%
\pgfpathlineto{\pgfqpoint{4.524161in}{0.347222in}}%
\pgfusepath{stroke}%
\end{pgfscope}%
\begin{pgfscope}%
\definecolor{textcolor}{rgb}{0.000000,0.000000,0.000000}%
\pgfsetstrokecolor{textcolor}%
\pgfsetfillcolor{textcolor}%
\pgftext[x=4.635272in,y=0.298611in,left,base]{\color{textcolor}\fontsize{10.000000}{12.000000}\selectfont \textsc{h-alpha} (with \textsc{h-clingo} heuristics, \(\displaystyle k_\mathrm{co}=\infty\))   (17)}%
\end{pgfscope}%
\begin{pgfscope}%
\pgfsetrectcap%
\pgfsetroundjoin%
\pgfsetlinewidth{1.505625pt}%
\definecolor{currentstroke}{rgb}{1.000000,0.647059,0.000000}%
\pgfsetstrokecolor{currentstroke}%
\pgfsetdash{}{0pt}%
\pgfpathmoveto{\pgfqpoint{8.026594in}{0.555556in}}%
\pgfpathlineto{\pgfqpoint{8.304372in}{0.555556in}}%
\pgfusepath{stroke}%
\end{pgfscope}%
\begin{pgfscope}%
\definecolor{textcolor}{rgb}{0.000000,0.000000,0.000000}%
\pgfsetstrokecolor{textcolor}%
\pgfsetfillcolor{textcolor}%
\pgftext[x=8.415483in,y=0.506944in,left,base]{\color{textcolor}\fontsize{10.000000}{12.000000}\selectfont \textsc{clingo}   (18)}%
\end{pgfscope}%
\begin{pgfscope}%
\pgfsetbuttcap%
\pgfsetroundjoin%
\pgfsetlinewidth{1.505625pt}%
\definecolor{currentstroke}{rgb}{1.000000,0.647059,0.000000}%
\pgfsetstrokecolor{currentstroke}%
\pgfsetdash{{9.600000pt}{2.400000pt}{1.500000pt}{2.400000pt}}{0.000000pt}%
\pgfpathmoveto{\pgfqpoint{8.026594in}{0.347222in}}%
\pgfpathlineto{\pgfqpoint{8.304372in}{0.347222in}}%
\pgfusepath{stroke}%
\end{pgfscope}%
\begin{pgfscope}%
\definecolor{textcolor}{rgb}{0.000000,0.000000,0.000000}%
\pgfsetstrokecolor{textcolor}%
\pgfsetfillcolor{textcolor}%
\pgftext[x=8.415483in,y=0.298611in,left,base]{\color{textcolor}\fontsize{10.000000}{12.000000}\selectfont \textsc{h-clingo}   (16)}%
\end{pgfscope}%
\end{pgfpicture}%
\makeatother%
\endgroup%
}
			\caption{Solver configurations, with numbers of solved instances}
			\label{fig-cactus-pup-legend}
		\end{subfigure}%
		\vspace{\floatsep}
		\begin{subfigure}{.32\textwidth}
			\centering
			\resizebox{\textwidth}{!}{\input{figures/cactus_pup_g.pgf}}
			\caption{\rev{}{Number of guesses}}
			\label{fig-cactus-pup-guesses}
		\end{subfigure}%
		\hspace*{\fill}
		\begin{subfigure}{.32\textwidth}
			\centering
			\resizebox{\textwidth}{!}{\input{figures/cactus_pup_walltime.pgf}}
			\caption{\rev{Accumulated time }{Time} consumption}
			\label{fig-cactus-pup-time}
		\end{subfigure}%
		\hspace*{\fill}
		\begin{subfigure}{.32\textwidth}
			\centering
			\resizebox{\textwidth}{!}{\input{figures/cactus_pup_memory.pgf}}
			\caption{Memory consumption}
			\label{fig-cactus-pup-memory}
		\end{subfigure}
		\caption{\rev{Time and memory }{Resource} consumption for solving each competition PUP instance}
		\label{fig-experimental-results-pup}
	\end{figure}

\rev{}{
	To solve the competition instances (\cref{fig-experimental-results-pup}), all systems used the same encoding as for the generated instances.
	The number of guesses needed to find the answer set is significantly reduced by our heuristics (\cref{fig-cactus-pup-guesses}).
	The effect on time and memory consumption is much lower, however (\cref{fig-cactus-pup-time,fig-cactus-pup-memory}).
	Overall, \slv{clingo} without domain-specific heuristics outperforms all other systems and solves all 18 instances.
	
	Again, domain-specific heuristics raised \slv{clingo}'s grounding efforts:
	For each of the 18 instances solved by \slv{clingo}, the system spent between 0\% and 50\% of total solving time in grounding, on average 7\%.
	When using domain-specific heuristics, \slv{h-clingo} solved 16 instances and spent between 2\% and 89\% in grounding, on average 40\% of overall time.
	\slv{dlv2} solved 15 instances and spent between 0\% and 10\% of total solving time in grounding, on average 2\%.
	
	For comparison, we also ran \slv{clingo} with \alphaslv's encoding without domain-specific heuristics.
	The results (which are not shown in the figures) revealed that \slv{clingo} performed very poorly with this encoding, solving only 5 of the simple instances and only 7 competition instances.
}

\subsection{Case Study 3: State-Space Search with A*}
\label{sec-applications-astar}

A* is a form of \emph{best-first search} that searches a weighted graph for the optimal path from a given start node to a given goal node.
Best-first search strategies expand the most promising \rev{of }{among} all the nodes encountered so far.
The promise of a node $n$ is estimated numerically by a heuristic evaluation function, traditionally denoted $f(n)$  \cite{DBLP:journals/tssc/HartNR68,aima,DBLP:books/daglib/0068933}.

\rev{A* evaluates nodes by combining $g(n)$, the known cost to reach the node, and $h(n)$, the estimated cost to get from the node to the goal \cite{aima}:}{A* evaluates nodes by combining $g(n)$, the path-cost to reach node $n$ from the start node, and $h(n)$, the estimated cost of the cheapest path from $n$ to the goal \cite{aima}:}
\begin{align*}
	f(n) = g(n) + h(n).
\end{align*}
A* is \emph{complete} on finite graphs with non-negative edge weights.
The graph-search version of A*\!, in which no state is explored repeatedly, is \emph{optimal} if $h$ is consistent.
Heuristics discussed in this section are consistent; for details, see \citeA{aima}.

\subsubsection{Problem Definition}

\begin{figure}
	\centering
	\begin{tikzpicture}[scale=0.55,node distance=0.4cm,>=latex]
	
	\foreach \i in {0,...,5} 
	{
		\draw [-] (0,\i) -- (6,\i);
	}

	\foreach \i in {0,...,6} 
	{
		\draw [-] (\i,0) -- (\i,5);
	}
	
	\foreach \i in {0,...,5} 
	{
		\node[anchor=west] at (\i+.1,-.5) {\i} ;
	}

	\foreach \i in {0,...,4} 
	{
		\node[anchor=south] at (-.5,\i) {\i} ;
	}

	\node[anchor=center] at (4.5,2.5) {S};
	\node[anchor=center] at (1.5,3.5) {G};
	
	\path[draw, fill=black] (2,1) rectangle (3,2) ; 
	\path[draw, fill=black] (3,3) rectangle (4,4) ; 
\end{tikzpicture}
	\caption{A sample Pathfinding instance (repeated from \cref{fig-pathfinding})}
	\label{fig-pathfinding-2}
\end{figure}
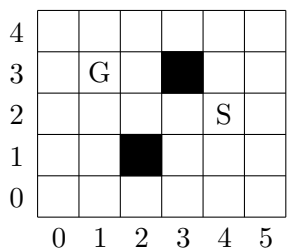

The formal specification of search problems will now be discussed.
The \emph{Pathfinding} problem introduced in \cref{ex-pathfinding-clingo,ex-pathfinding-alpha} will be used to give examples.\footnote{Note that we disregard time steps in the formulation of Pathfinding as a search problem.}
We repeat \cref{fig-pathfinding} in \cref{fig-pathfinding-2} for easier reference.

\begin{definition}[Search Problem]
	\label{def:problem}
A \emph{problem} to be solved by a search algorithm like A* can be defined formally by five components \cite{aima}:
\begin{itemize}
	\item The \textbf{initial state}, e.g., $\mathrm{at}(4,2)$.
	\item The \textbf{actions} available.
			Given a state $s$, $\func{actions}(s)$ returns the set of actions that can be executed in $s$.\\
			For example, $\func{actions}(\mathrm{at}(4,2)) = \{ \mathrm{move}(4,3), \mathrm{move}(5,2), \mathrm{move}(4,1), \mathrm{move}(3,2)\}$.
	\item The \textbf{transition model} specifying the results of each action.\\
			For example, $\func{result}(\mathrm{at}(4,2),\mathrm{move}(3,2)) = \mathrm{at}(3,2)$.\\
			The \textbf{state-space} of the problem, the set of all states reachable from the initial state by any sequence of actions, is implicitly defined by the initial state, the actions, and the transition model.
			The state-space forms a graph in which the nodes are states, and the edges are actions.
	\item The \textbf{goal test}, which determines whether a given state is a goal state.
			The set of goal states can be defined implicitly or explicitly; in our example, it contains the single element $\mathrm{at}(1,3)$.
	\item A \textbf{cost} function that assigns a numerical cost to each edge in the graph.
			The step cost of reaching state $s'$ from $s$ by applying action $a$ is denoted by $c(s,a,s')$.
			In our example, the cost of each step is the same, e.g., $c(\mathrm{at}(4,2),\mathrm{move}(3,2),\mathrm{at}(3,2)) = 1$.
			In practical routing problems, step cost might be defined as the length of a road or the time needed to travel between two cities.
\end{itemize}
\end{definition}

\rev{}{A solution for the search problem is a path, representing a sequence of actions, from the initial state to a goal state. An optimal solution is a solution with lowest path cost among all solutions.}

\Cref{alg:a-star} shows an informal specification of the A* algorithm \cite<adapted from Figures 3.7 and 3.14 by>{aima}.
For details, consult \citeA{DBLP:journals/tssc/HartNR68,aima,DBLP:books/daglib/0068933}.

\begin{algorithm}[t]
	\caption{Graph Search with A* \cite<adapted from Figures 3.7 and 3.14 by>{aima}}
	\label{alg:a-star}
	\DontPrintSemicolon
	
	\KwIn{$\astarproblem$, a problem defined by the components specified in \cref{def:problem}.
	}
	\KwOut{A solution, or failure.}
	
	\smallskip
	Initialize frontier $\astarfrontier$, a priority queue ordered by $f(\cdot)$, initially containing the initial state of $\astarproblem$\;
	Initialize explored set $\astarexplored$ to be empty\;
	\While{$\astarfrontier$ is not empty}{
		Pop node $n$ with least $f(n)$ from $F$\;
		\If{$n$ contains a goal state}{
			\Return the corresponding solution\;
		}
		\rev{Add $n$ to $E$}{Add state of $n$ to $E$}\;
		\ForEach{action $a$ available in the state of $n$}{
			\rev{}{Generate child node by applying action $a$ on node $n$ \;}
			\If{the child state is neither in $\astarfrontier$ nor in $\astarexplored$}{
				Insert the child \rev{state}{node} into $\astarfrontier$\;
			}
			\ElseIf{the child state is already in $\astarfrontier$, but with a higher path cost}{
				\rev{Update the path cost of the child state in $\astarfrontier$}{Replace that node in $\astarfrontier$ with child} \;
			}
		}
	}
	\Return failure\;
\end{algorithm}

\subsubsection{Instances, Encodings, and Heuristics}

We now present our ASP-based encoding of the A* search algorithm \rev{}{ based on graph-search}, which is the first of its kind to the best of our knowledge. \rev{}{As we apply graph-search we assume the consistency of the heuristic evaluation function $h(n)$. Moreover, we require that actions are deterministic, i.e. the result of an action is exactly one successor state.}
To facilitate using our encoding with different search problems, we distinguish between the encoding of A* itself and a problem-specific encoding that can be used with different problem instances.
\rev{Our presentation will move from most specific (problem instances for Pathfinding) to most general (A*). }{
We will demonstrate the usage of our A* encoding with two search problems: Pathfinding known from \cref{ex-pathfinding-clingo}, and routing on a real-world street network.
We will first present A* with the Pathfinding problem and move from most specific (problem instances) to most general (A* encoding).
Afterwards, we will present the routing application. State-based search, where the problem's solution is a path, deals with pathfinding. Therefore we focus our evaluation on pathfinding and routing, showing the benefits of lazy-grounding combined with heuristic search.}

\paragraph{Problem Instances for Pathfinding.}
Instances are defined as stated in \cref{ex-pathfinding-clingo}, i.e., they consist of atoms over the \rev{}{following} predicates\rev{}{:}
\rev{$\mathrm{xmin}/1$, $\mathrm{xmax}/1$, $\mathrm{ymin}/1$, $\mathrm{ymax}/1$, $\mathrm{start}/2$, $\mathrm{goal}/2$, and $\mathrm{obstacle}/2$. }
{
\begin{itemize}
	\item $\mathrm{xmin}/1$,
	\item $\mathrm{xmax}/1$,
	\item $\mathrm{ymin}/1$,
	\item $\mathrm{ymax}/1$,
	\item $\mathrm{start}/2$,
	\item $\mathrm{goal}/2$, and
	\item $\mathrm{obstacle}/2$.
\end{itemize}
}

For our experiments, we have generated instances as follows.
Each instance represents a square plane of width $w$.
For each $w \in \{ 5, 10, \dots, 500 \}$ there exists one instance, so there are 100 instances in total.
The start is located in the top-left corner, and the goal is located in the bottom-right corner.
Each square in the plane, except for the start and the goal, is an obstacle with a probability of $0.2$.
Additionally, two long vertical walls cut through the plane, one starting at the bottom and one at the top of the plane.
The $x$-positions of these walls, as well as their lengths, are randomly determined within reasonable ranges.

\paragraph{Problem-Specific Encoding for Pathfinding.}
The problem-specific encoding defines some intermediate predicates:
\rev{$\mathrm{neighbour}/4$ defines which squares are adjacent to each other, and $\mathrm{manhattan}/3$ defines the Manhattan distance from each square to the goal. }
{
\begin{itemize}
	\item $\mathrm{neighbour}/4$ defines which squares are adjacent to each other, and 
	\item $\mathrm{manhattan}/3$ defines the Manhattan distance from each square to the goal.
\end{itemize}
}
For example, $\mathrm{neighbour}(4,2,3,2)$ denotes that $(4,2)$ is adjacent to $(3,2)$, and $\mathrm{manhattan}(3,2,3)$ denotes that the Manhattan distance from $(3,2)$ to the goal at $(1,3)$ is 3.

As input to the problem-independent A* encoding, the following predicates have to be defined:
\rev{$\mathrm{init}/1$ defines the initial state;
$\mathrm{goal}/1$ defines the goal state;
$\mathrm{c}/4$ defines the cost function;
$\mathrm{h}/2$ defines the heuristic evaluation function $h$ (which is here the Manhattan distance to the goal). }
{
\begin{itemize}
	\item $\mathrm{init}/1$ defines the initial state;
	\item $\mathrm{goal}/1$ defines the goal state;
	\item $\mathrm{c}/4$ defines the cost function; and
	\item $\mathrm{h}/2$ defines the heuristic evaluation function $h$ (which is here the Manhattan distance to the goal).
\end{itemize}
}
Actions and transition model need not be explicitly defined because this information is implicitly contained in $\mathrm{c}/4$.

The input to A* is realised in the problem-specific Pathfinding encoding by the following rules:
\begin{align*}
	\mathrm{init(at(X,Y))} \leftarrow\; &\mathrm{start(X,Y)}.	\\
	\mathrm{goal(at(X,Y))} \leftarrow\; &\mathrm{goal(X,Y)}.	\\
	\mathrm{c(at(AtX,AtY)}, \mathrm{move(X,Y)}, \mathrm{at(X,Y),StepCost)} \leftarrow\;
												&\mathrm{neighbour(AtX,AtY,X,Y)},\\
												&\naf \mathrm{obstacle(X,Y)},\\
												&\mathrm{StepCost}=1.	\\
	\mathrm{h(at(X,Y),H)} \leftarrow\; &\mathrm{manhattan(X,Y,H)}.
\end{align*}

\paragraph{Problem-Independent A* Encoding.}
We will now present our ASP-based A* encoding step by step.
This problem-independent encoding stays the same for any search problem to which A* is applied.

The first step is to retrieve the information on the $\func{actions}$ and $\func{result}$ functions from the problem definition, implicitly contained in $\mathrm{c}/4$.
\begin{align*}
	\mathrm{result(ParentState,Action,ChildState)} \leftarrow\; &\mathrm{c(ParentState,Action,ChildState,StepCost)}.	\\
	\mathrm{action(ParentState,Action)} \leftarrow\; &\mathrm{result(ParentState,Action,ChildState)}.
\end{align*}

The $\mathrm{n}/4$ predicate is used to represent each node $n$ in the searched graph by state $S$, parent state $\mi{PS}$ (which generated $n$), action $A$ (applied to generate $n$), and path cost $G = g(n)$.
There also exists an $\mathrm{n}/1$ predicate that is used when only the state of a node is needed.
Recall that $\mathrm{init}/1$ is defined in the problem-specific encoding.
The $\mathrm{\rev{choose}{explore}}/2$ predicate will be defined further below; it is used for state-action pairs that have already been explored.
\begin{align*}
	\mathrm{n(S,null,null,0)} \leftarrow\; &\mathrm{init(S)}.	\\
	\mathrm{n(S,PS,A,G)} \leftarrow\; &\mathrm{result(PS,A,S)}, \mathrm{\rev{choose}{explore}(PS,A)}, \mathrm{g(PS,A,G)}.	\\
	\mathrm{n(S)} \leftarrow\; &\mathrm{n(S,PS,A,G)}.
\end{align*}

\rev{}{Note that if $h(n)$ is consistent, a property of graph-search is that nodes are only explored if their smallest $g$-value was found \cite{aima}. Based on this property and the determinism of actions it follows that 
$\mi{PS}$ and $A$ are uniquely identifying a tuple of $\mathrm{n}/4$}.

The $f$-value for each node is the sum of the node's $g$-value (cost from the initial state to the node) and its $h$-value (heuristically estimated cost from the node to the \rev{}{nearest} goal).
Recall that $\mathrm{h}/2$ is defined in the problem-specific encoding.
\begin{align*}
	\mathrm{g(S,G)} \leftarrow\; &\mathrm{n(S,P,A,G)}.	\\
	\mathrm{f(S,G+H)} \leftarrow\; &\mathrm{g(S,G)}, \mathrm{h(S,H)}.
\end{align*}

Nodes are only generated in our approach when they are explored.
However, $f(n)$ is already needed before $n$ is explored to decide which nodes to explore.
Therefore, we define $f$ also for state-action pairs.
\begin{align*}
	\mathrm{g(PS,A,ParentG+StepCost)} \leftarrow\; &\mathrm{g(PS,ParentG)}, \mathrm{c(PS,A,S,StepCost)}.	\\
	\mathrm{h(PS,A,H)} \leftarrow\; &\mathrm{result(PS,A,S)}, \mathrm{h(S,H)}.	\\
	\mathrm{f(S,A,G+H)} \leftarrow\; &\mathrm{g(S,A,G)}, \mathrm{h(S,A,H)}.
\end{align*}

Now we can encode elements of A* (cf.\ \cref{alg:a-star}) as rules.
There is no one-to-one correspondence between statements in the algorithm and rules in our ASP encodings, so we present ASP rules in a logical order.

The \rev{}{union of} frontier \rev{}{and explored set} is represented by $\mathrm{frontier\rev{}{\_and\_explored\_states}}/2$.
The predicate consists of state-action pairs instead of nodes because nodes are not generated before they are explored.
Since an atom's truth value cannot change from true to false without backtracking, atoms of the $\mathrm{frontier\rev{}{\_and\_explored\_states}}/2$ predicate also include nodes that have already been removed from the frontier. \rev{}{These removed nodes are in the explored set, i.e.\ nodes represented by $\mathrm{n}/4$ were generated.}
The following rule adds an explored node's children to \rev{the frontier}{$\mathrm{frontier\_and\_explored\_states}/2$}:
\begin{align*}
	\mathrm{frontier\rev{}{\_and\_explored\_states}(S,A)} \leftarrow\; &\mathrm{n(S)}, \mathrm{action(S,A)}, \naf \mathrm{goal(S)}.
\end{align*}

The explored set $\astarexplored$ is already represented by predicates $\mathrm{n}/4$ and $\mathrm{n}/1$ because atoms of these predicates represent nodes that are only generated when explored.
However, to avoid exploring the same state repeatedly, the $\mathrm{explored\_by\_other\_step}/3$ predicate stores actions that may not be used because the resulting children have already been explored.
\begin{align*}
	\mathrm{explored\_by\_other\_step(S,PS,A)} \leftarrow\; &\mathrm{n(S,PS',A',G)}, \mathrm{result(PS,A,S)}, \mathrm{PS}' \neq \mathrm{PS}.	\\
	\mathrm{explored\_by\_other\_step(S,PS,A)} \leftarrow\; &\mathrm{n(S,PS',A',G)}, \mathrm{result(PS,A,S)}, \mathrm{A'} \neq \mathrm{A}.
\end{align*}

When a child state is already explored, the corresponding state-action pair is still added to the frontier but marked as suboptimal.
\begin{align*}
	\mathrm{suboptimal\_step(PS,A)} \leftarrow\; &\mathrm{frontier\rev{}{\_and\_explored\_states}(PS,A)}, \mathrm{result(PS,A,S)},\\
		&\mathrm{explored\_by\_other\_step(S,PS,A)}.
\end{align*}

Similarly, when a cheaper path to a child state is discovered, the existing state-action pair in the frontier is marked as suboptimal.
\begin{align*}
	\mathrm{suboptimal\_step(PS,A)} \leftarrow\; &\mathrm{frontier\rev{}{\_and\_explored\_states}(PS,A)}, \mathrm{result(PS,A,S)},\\
		&\mathrm{frontier\rev{}{\_and\_explored\_states}(PS',A')}, \mathrm{result(PS',A',S)},\\
		&\mathrm{f(PS,A,F)}, \mathrm{f(PS',A',F')}, \mathrm{F} > \mathrm{F'}.
\end{align*}

The following two rules are here to recognise when a goal state is explored.
\begin{align*}
	\mathrm{goal\_found(S)} \leftarrow\; &\mathrm{n(S)}, \mathrm{goal(S)}. \\
	\mathrm{goal\_found} \leftarrow\; &\mathrm{goal\_found(S)}.
\end{align*}

A found goal is not accepted if a node with a lower $f$-value is still on the frontier.
This restriction ensures optimality of found solutions even without heuristic directives.\footnote{\rev{}{Using the heuristics presented below, the problem can be solved without backtracking, which means that constraints are not needed when using these heuristics. We still include constraints in the encoding to render it a correct declarative problem specification when viewed without heuristics.}}
\begin{align*}
	\leftarrow\; &\mathrm{goal\_found(S)}, \mathrm{\rev{choose}{explore}(PS,A)}, \mathrm{result(PS,A,S)}, \mathrm{f(S,F)},\\
	&\mathrm{frontier\rev{}{\_and\_explored\_states}(S',A')},\\
	&\naf \mathrm{\rev{choose}{explore}(S',A')}, \naf \mathrm{suboptimal\_step(S',A')},\\
	&\mathrm{f(S',A',F')}, \mathrm{F'} < \mathrm{F}.
\end{align*}

The encoding contains a single choice rule, which realises choosing a leaf node from the frontier.
Without heuristic directives, which will be defined later, this rule does not differentiate between nodes in the frontier.
\begin{align*}
	\{~ \mathrm{\rev{choose}{explore}(S,A)} ~\} \leftarrow\; &\mathrm{frontier\rev{}{\_and\_explored\_states}(S,A)}, \naf \mathrm{suboptimal\_step(S,A)}.
\end{align*}

\rev{}{Note that by the consistency property of $h(n)$ nodes in graph-search are only selected for exploration if the minimum $g$-value for this node was found. We assure this property by the heuristic formulated for $\mathrm{explore/2}$ (see below). $S,A$ tuples of the frontier with smallest $f$-value are explored first. Consequently, if there are two distinct state/action pairs $(s,a)$ and $(s',a')$ which were not explored and which lead to the same successor state, then our heuristic will select the state/action pair for exploaration which is on an optimal path.}

Failure is derived when the frontier is empty, and no goal has been found.
Our encoding of A* is always satisfiable. An answer set contains a solution if the search problem has a solution, and an answer set contains the $\mathrm{failure}$ atom if there is no solution.
We have decided this way because the alternative, having a constraint that $\mathrm{goal\_found}$ must be true in an answer set, makes solving performance deteriorate sharply for instances without a solution.
\begin{align*}
	\mathrm{frontier\_nonempty\rev{(S,A)}{}} \leftarrow\; &\mathrm{frontier\rev{}{\_and\_explored\_states}(S,A)},\\
		&\naf \mathrm{suboptimal\_step(S,A)}, \naf \mathrm{\rev{choose}{explore}(S,A)}. \\
	\mathrm{failure} \leftarrow\; &\naf \mathrm{frontier\_nonempty}, \naf \mathrm{goal\_found}. \\
	\leftarrow\; &\naf \mathrm{failure}, \naf \mathrm{goal\_found}. \\
\end{align*}

When the goal has been found, the path from the initial state to the goal state can be computed along with corresponding path costs.
\begin{align*}
	\mathrm{path\_to\_goal(PS,A,S)} \leftarrow\; &\mathrm{\rev{choose}{explore}(PS,A)}, \mathrm{result(PS,A,S)}, \mathrm{goal\_found(S)}. \\
	\mathrm{path\_to\_goal(PS,A,S)} \leftarrow\; &\mathrm{\rev{choose}{explore}(PS,A)}, \mathrm{result(PS,A,S)}, \mathrm{path\_to\_goal(S,\rev{A }{A'},S')}. \\
	\mathrm{cost\_to\_goal(G)} \leftarrow\; &\mathrm{goal\_found(GoalState)}, \mathrm{g(GoalState,G)}.
\end{align*}

This subsequent computation is necessary because A* may choose to apply more than one action on the same state if one of those actions turns out to lead to a more costly path to the goal.
Our approach does not need backtracking, so the answer set includes all $\mathrm{\rev{choose}{explore}}/2$ atoms tried, and we need to reconstruct the path that leads to the goal. 

The result computed by A* is encoded by the predicates
\rev{$\mathrm{path\_to\_goal}/3$, $\mathrm{cost\_to\_goal}/1$, and $\mathrm{failure}/0$, }
{
\begin{itemize}
	\item $\mathrm{path\_to\_goal}/3$,
	\item $\mathrm{cost\_to\_goal}/1$, and
	\item $\mathrm{failure}/0$,
\end{itemize}} so an answer set solver can be instructed to display only atoms of those predicates in an answer set.
For \alphaslv, this can be done with the command-line arguments \texttt{-{}-filter path\_to\_goal -{}-filter cost\_to\_goal -{}-filter failure}.

Of course, our A* encoding also needs heuristic directives to work correctly.
Without heuristics, the encoding would just search arbitrarily through the state-space and ignore the $f$-value of nodes.
Therefore, our primary heuristic directive prefers to choose those nodes from the frontier whose $f$-value is the lowest.
\begin{align*}
	\heudirstmt\ \mathrm{\rev{choose}{explore}(PS,A)} :\; &\mathrm{frontier\rev{}{\_and\_explored\_states}(PS,A)},\\
		&\naf \mathrm{suboptimal\_step(PS,A)},\\
		&\mathrm{f(PS,A,PathCost)}, \naf \heusignT\ \mathrm{goal\_found}. \;[-\mathrm{PathCost}@3]
\end{align*}

A second heuristic directive closes unassigned choice points when the goal is found.
\begin{align*}
	\heudirstmt\ \heusignF\ \mathrm{\rev{choose}{explore}(PS,A)} :\; &\mathrm{frontier\rev{}{\_and\_explored\_states}(PS,A)},\\
		&\heusignT\ \mathrm{goal\_found}. \;[1@2]
\end{align*}

\rev{
Finally, to eliminate any remaining choice points, two heuristics either determine failure or that the frontier is non-empty, whichever of the two is applicable.
\begin{align*}
	&\heudirstmt\ \mathrm{failure}. \;[1@1] \\
	&\heudirstmt\ \mathrm{frontier\_nonempty(S,A)} :\; \mathrm{frontier\rev{}{\_and\_explored\_states}(S,A)}. \;[1@1]
\end{align*}
}{}

The heuristics we created for \alphaslv\ cannot be used with \slv{clingo} due to the usage of $\heusignT$ and default negation.
To assess the performance of \slv{clingo}, we have also created an encoding omitting unsupported literals in heuristic conditions.
\rev{}{
	Furthermore, an \alphaslv\ encoding with heuristics without our novel features, compliant with those employed by \slv{clingo}, has been created.
}

\paragraph{Generating States on Demand.}
In many practical domains, the number of states is vast, thus prohibiting the upfront generation of all states.
For example, in the well-known 8-puzzle, $181,440$ distinct states are reachable.
While this is a manageable number, the corresponding number for the 15-puzzle is already about $10^{13}$ \cite{aima}.

Since only a small fraction of all states is usually explored by A*\rev{\!}{\ (depending on the quality of the heuristic function)}, a crucial memory-saving feature is to generate only the states that A* needs.
Our approach can generate states on demand using the interface between problem-specific encoding and A* encoding bidirectionally instead of just passing information from the problem-specific encoding to A*\!.

As described above, predicates $\mathrm{init}/1$, $\mathrm{goal}/1$, $\mathrm{c}/4$, and $\mathrm{h}/2$ are defined in the problem-specific encoding and accessed by the A* encoding.
In the other direction, the problem-specific encoding can access the $\mathrm{frontier\rev{}{\_and\_explored\_states}}/2$ predicate defined in the problem-independent A* encoding to generate states on demand, for example, by introducing a new $\mathrm{state}/1$ predicate:
\begin{align*}
	\mathrm{state(InitState)} \leftarrow\; &\mathrm{init(InitState)}. \\
	\mathrm{state(GoalState)} \leftarrow\; &\mathrm{goal(GoalState)}. \\
	\mathrm{state(ChildState)} \leftarrow\; &\mathrm{frontier\rev{}{\_and\_explored\_states}(ParentState,Action)},\\
			&\mathrm{result(ParentState,Action,ChildState)}.
\end{align*}
This new $\mathrm{state}/1$ predicate can then be used in the body of rules deriving $\mathrm{c}/4$ and $\mathrm{h}/2$ to derive also information on step cost and heuristic values only on-demand, utilising lazy grounding.
Deriving this information on-demand is not possible with ground-and-solve.

\rev{}{
	\paragraph{Real-World Routing.}
	To test our A* encoding on real world routing problems, we obtained a graph representation of the walkable street network for the first district of Vienna, Austria, from OpenStreetMap,\footnote{\url{https://www.openstreetmap.org}} employing the OSMnx Python library \cite{Boeing2017}.\footnote{\url{https://github.com/gboeing/osmnx}}
	We then pre-processed the graph by consolidating intersections (merging nodes located within 15m of each other) and removing parallel edges.
	\cref{fig-innere_stadt} shows the resulting graph, which contains 462 nodes and 1632 edges.
	
	\begin{figure}
		\centering
		\resizebox{.5\textwidth}{!}{\input{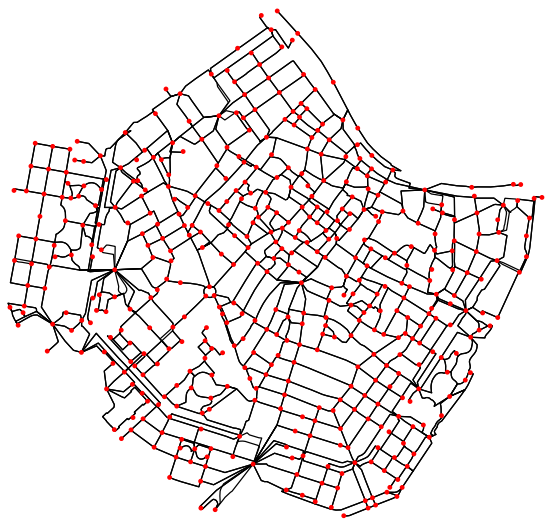}}
		\caption{\rev{}{Vienna's first district}}
		\label{fig-innere_stadt}
	\end{figure}
	
	We have generated 100 instances, each consisting of the following atoms:
	\begin{itemize}
		\item 1632 atoms of the predicate $\mathrm{edge}/3$, in which the first two arguments denote two nodes connected by an edge and the third argument denotes the length of the edge in meters,
		\item $106,953$ atoms\footnote{Since the graph contains $n=462$ nodes, there are $\frac{n \cdot (n+1)}{2} = 106,953$ unique pairs of nodes including reflexive pairs.} of the predicate $\mathrm{distance}/3$, in which the first two arguments denote nodes and the third argument denotes the euclidean distance between them in meters,
		\item one atom of the predicate $\mathrm{origin}/1$ denoting the origin node of the route to be computed, and
		\item one atom of the predicate $\mathrm{destination}/1$ denoting the destination node of the route to be computed.
	\end{itemize}
	The set of $\mathrm{edge}/3$ and $\mathrm{distance}/3$ atoms encodes the street network graph and is the same in each of the 100 instances.
	Origin and destination nodes, however, have been chosen randomly for each instance.
	
	The problem-specific encoding for street network routing then consists of only the following five rules:
	\begin{align*}
		\mathrm{init(at(N))} \leftarrow\; &\mathrm{origin(N)}. \\
		\mathrm{goal(at(N))} \leftarrow\; &\mathrm{destination(N)}. \\
		\mathrm{c(at(X),move(Y),at(Y),Length)} \leftarrow\; &\mathrm{edge(X,Y,Length)}. \\
		\mathrm{h(at(X),Distance)} \leftarrow\; &\mathrm{destination(Y), distance(X,Y,Distance)}. \\
		\mathrm{distance(Y,X,Distance)} \leftarrow\; &\mathrm{distance(X,Y,Distance)}.
	\end{align*}
}

\subsubsection{Results}
\label{sec-applications-astar-results}

	\begin{figure}[t]
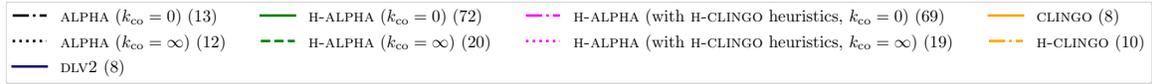
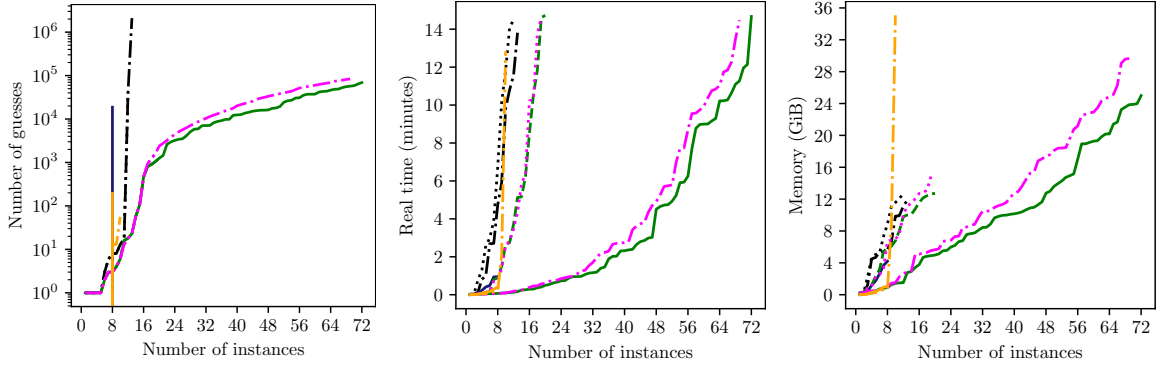

		\centering
		\begin{subfigure}{\textwidth}
			\centering
			\resizebox{\textwidth}{!}{
\begingroup%
\makeatletter%
\begin{pgfpicture}%
\pgfpathrectangle{\pgfpointorigin}{\pgfqpoint{9.359771in}{0.666667in}}%
\pgfusepath{use as bounding box, clip}%
\begin{pgfscope}%
\pgfsetbuttcap%
\pgfsetmiterjoin%
\definecolor{currentfill}{rgb}{1.000000,1.000000,1.000000}%
\pgfsetfillcolor{currentfill}%
\pgfsetlinewidth{0.000000pt}%
\definecolor{currentstroke}{rgb}{1.000000,1.000000,1.000000}%
\pgfsetstrokecolor{currentstroke}%
\pgfsetdash{}{0pt}%
\pgfpathmoveto{\pgfqpoint{0.000000in}{0.000000in}}%
\pgfpathlineto{\pgfqpoint{9.359771in}{0.000000in}}%
\pgfpathlineto{\pgfqpoint{9.359771in}{0.666667in}}%
\pgfpathlineto{\pgfqpoint{0.000000in}{0.666667in}}%
\pgfpathclose%
\pgfusepath{fill}%
\end{pgfscope}%
\begin{pgfscope}%
\pgfsetbuttcap%
\pgfsetmiterjoin%
\definecolor{currentfill}{rgb}{1.000000,1.000000,1.000000}%
\pgfsetfillcolor{currentfill}%
\pgfsetfillopacity{0.800000}%
\pgfsetlinewidth{1.003750pt}%
\definecolor{currentstroke}{rgb}{0.800000,0.800000,0.800000}%
\pgfsetstrokecolor{currentstroke}%
\pgfsetstrokeopacity{0.800000}%
\pgfsetdash{}{0pt}%
\pgfpathmoveto{\pgfqpoint{0.027778in}{0.000000in}}%
\pgfpathlineto{\pgfqpoint{9.331993in}{0.000000in}}%
\pgfpathquadraticcurveto{\pgfqpoint{9.359771in}{0.000000in}}{\pgfqpoint{9.359771in}{0.027778in}}%
\pgfpathlineto{\pgfqpoint{9.359771in}{0.638889in}}%
\pgfpathquadraticcurveto{\pgfqpoint{9.359771in}{0.666667in}}{\pgfqpoint{9.331993in}{0.666667in}}%
\pgfpathlineto{\pgfqpoint{0.027778in}{0.666667in}}%
\pgfpathquadraticcurveto{\pgfqpoint{-0.000000in}{0.666667in}}{\pgfqpoint{-0.000000in}{0.638889in}}%
\pgfpathlineto{\pgfqpoint{-0.000000in}{0.027778in}}%
\pgfpathquadraticcurveto{\pgfqpoint{-0.000000in}{0.000000in}}{\pgfqpoint{0.027778in}{0.000000in}}%
\pgfpathclose%
\pgfusepath{stroke,fill}%
\end{pgfscope}%
\begin{pgfscope}%
\pgfsetbuttcap%
\pgfsetroundjoin%
\pgfsetlinewidth{1.505625pt}%
\definecolor{currentstroke}{rgb}{0.000000,0.000000,0.000000}%
\pgfsetstrokecolor{currentstroke}%
\pgfsetdash{{9.600000pt}{2.400000pt}{1.500000pt}{2.400000pt}}{0.000000pt}%
\pgfpathmoveto{\pgfqpoint{0.055556in}{0.555556in}}%
\pgfpathlineto{\pgfqpoint{0.333333in}{0.555556in}}%
\pgfusepath{stroke}%
\end{pgfscope}%
\begin{pgfscope}%
\definecolor{textcolor}{rgb}{0.000000,0.000000,0.000000}%
\pgfsetstrokecolor{textcolor}%
\pgfsetfillcolor{textcolor}%
\pgftext[x=0.444444in,y=0.506944in,left,base]{\color{textcolor}\fontsize{10.000000}{12.000000}\selectfont \textsc{alpha} (\(\displaystyle k_\mathrm{co}=0\))   (13)}%
\end{pgfscope}%
\begin{pgfscope}%
\pgfsetbuttcap%
\pgfsetroundjoin%
\pgfsetlinewidth{1.505625pt}%
\definecolor{currentstroke}{rgb}{0.000000,0.000000,0.000000}%
\pgfsetstrokecolor{currentstroke}%
\pgfsetdash{{1.500000pt}{2.475000pt}}{0.000000pt}%
\pgfpathmoveto{\pgfqpoint{0.055556in}{0.347222in}}%
\pgfpathlineto{\pgfqpoint{0.333333in}{0.347222in}}%
\pgfusepath{stroke}%
\end{pgfscope}%
\begin{pgfscope}%
\definecolor{textcolor}{rgb}{0.000000,0.000000,0.000000}%
\pgfsetstrokecolor{textcolor}%
\pgfsetfillcolor{textcolor}%
\pgftext[x=0.444444in,y=0.298611in,left,base]{\color{textcolor}\fontsize{10.000000}{12.000000}\selectfont \textsc{alpha} (\(\displaystyle k_\mathrm{co}=\infty\))   (12)}%
\end{pgfscope}%
\begin{pgfscope}%
\pgfsetrectcap%
\pgfsetroundjoin%
\pgfsetlinewidth{1.505625pt}%
\definecolor{currentstroke}{rgb}{0.098039,0.098039,0.439216}%
\pgfsetstrokecolor{currentstroke}%
\pgfsetdash{}{0pt}%
\pgfpathmoveto{\pgfqpoint{0.055556in}{0.138889in}}%
\pgfpathlineto{\pgfqpoint{0.333333in}{0.138889in}}%
\pgfusepath{stroke}%
\end{pgfscope}%
\begin{pgfscope}%
\definecolor{textcolor}{rgb}{0.000000,0.000000,0.000000}%
\pgfsetstrokecolor{textcolor}%
\pgfsetfillcolor{textcolor}%
\pgftext[x=0.444444in,y=0.090278in,left,base]{\color{textcolor}\fontsize{10.000000}{12.000000}\selectfont \textsc{dlv2}   (8)}%
\end{pgfscope}%
\begin{pgfscope}%
\pgfsetrectcap%
\pgfsetroundjoin%
\pgfsetlinewidth{1.505625pt}%
\definecolor{currentstroke}{rgb}{0.000000,0.501961,0.000000}%
\pgfsetstrokecolor{currentstroke}%
\pgfsetdash{}{0pt}%
\pgfpathmoveto{\pgfqpoint{2.082142in}{0.555556in}}%
\pgfpathlineto{\pgfqpoint{2.359920in}{0.555556in}}%
\pgfusepath{stroke}%
\end{pgfscope}%
\begin{pgfscope}%
\definecolor{textcolor}{rgb}{0.000000,0.000000,0.000000}%
\pgfsetstrokecolor{textcolor}%
\pgfsetfillcolor{textcolor}%
\pgftext[x=2.471031in,y=0.506944in,left,base]{\color{textcolor}\fontsize{10.000000}{12.000000}\selectfont \textsc{h-alpha} (\(\displaystyle k_\mathrm{co}=0\))   (72)}%
\end{pgfscope}%
\begin{pgfscope}%
\pgfsetbuttcap%
\pgfsetroundjoin%
\pgfsetlinewidth{1.505625pt}%
\definecolor{currentstroke}{rgb}{0.000000,0.501961,0.000000}%
\pgfsetstrokecolor{currentstroke}%
\pgfsetdash{{5.550000pt}{2.400000pt}}{0.000000pt}%
\pgfpathmoveto{\pgfqpoint{2.082142in}{0.347222in}}%
\pgfpathlineto{\pgfqpoint{2.359920in}{0.347222in}}%
\pgfusepath{stroke}%
\end{pgfscope}%
\begin{pgfscope}%
\definecolor{textcolor}{rgb}{0.000000,0.000000,0.000000}%
\pgfsetstrokecolor{textcolor}%
\pgfsetfillcolor{textcolor}%
\pgftext[x=2.471031in,y=0.298611in,left,base]{\color{textcolor}\fontsize{10.000000}{12.000000}\selectfont \textsc{h-alpha} (\(\displaystyle k_\mathrm{co}=\infty\))   (20)}%
\end{pgfscope}%
\begin{pgfscope}%
\pgfsetbuttcap%
\pgfsetroundjoin%
\pgfsetlinewidth{1.505625pt}%
\definecolor{currentstroke}{rgb}{1.000000,0.000000,1.000000}%
\pgfsetstrokecolor{currentstroke}%
\pgfsetdash{{9.600000pt}{2.400000pt}{1.500000pt}{2.400000pt}}{0.000000pt}%
\pgfpathmoveto{\pgfqpoint{4.246383in}{0.555556in}}%
\pgfpathlineto{\pgfqpoint{4.524161in}{0.555556in}}%
\pgfusepath{stroke}%
\end{pgfscope}%
\begin{pgfscope}%
\definecolor{textcolor}{rgb}{0.000000,0.000000,0.000000}%
\pgfsetstrokecolor{textcolor}%
\pgfsetfillcolor{textcolor}%
\pgftext[x=4.635272in,y=0.506944in,left,base]{\color{textcolor}\fontsize{10.000000}{12.000000}\selectfont \textsc{h-alpha} (with \textsc{h-clingo} heuristics, \(\displaystyle k_\mathrm{co}=0\))   (69)}%
\end{pgfscope}%
\begin{pgfscope}%
\pgfsetbuttcap%
\pgfsetroundjoin%
\pgfsetlinewidth{1.505625pt}%
\definecolor{currentstroke}{rgb}{1.000000,0.000000,1.000000}%
\pgfsetstrokecolor{currentstroke}%
\pgfsetdash{{1.500000pt}{2.475000pt}}{0.000000pt}%
\pgfpathmoveto{\pgfqpoint{4.246383in}{0.347222in}}%
\pgfpathlineto{\pgfqpoint{4.524161in}{0.347222in}}%
\pgfusepath{stroke}%
\end{pgfscope}%
\begin{pgfscope}%
\definecolor{textcolor}{rgb}{0.000000,0.000000,0.000000}%
\pgfsetstrokecolor{textcolor}%
\pgfsetfillcolor{textcolor}%
\pgftext[x=4.635272in,y=0.298611in,left,base]{\color{textcolor}\fontsize{10.000000}{12.000000}\selectfont \textsc{h-alpha} (with \textsc{h-clingo} heuristics, \(\displaystyle k_\mathrm{co}=\infty\))   (19)}%
\end{pgfscope}%
\begin{pgfscope}%
\pgfsetrectcap%
\pgfsetroundjoin%
\pgfsetlinewidth{1.505625pt}%
\definecolor{currentstroke}{rgb}{1.000000,0.647059,0.000000}%
\pgfsetstrokecolor{currentstroke}%
\pgfsetdash{}{0pt}%
\pgfpathmoveto{\pgfqpoint{8.026594in}{0.555556in}}%
\pgfpathlineto{\pgfqpoint{8.304372in}{0.555556in}}%
\pgfusepath{stroke}%
\end{pgfscope}%
\begin{pgfscope}%
\definecolor{textcolor}{rgb}{0.000000,0.000000,0.000000}%
\pgfsetstrokecolor{textcolor}%
\pgfsetfillcolor{textcolor}%
\pgftext[x=8.415483in,y=0.506944in,left,base]{\color{textcolor}\fontsize{10.000000}{12.000000}\selectfont \textsc{clingo}   (8)}%
\end{pgfscope}%
\begin{pgfscope}%
\pgfsetbuttcap%
\pgfsetroundjoin%
\pgfsetlinewidth{1.505625pt}%
\definecolor{currentstroke}{rgb}{1.000000,0.647059,0.000000}%
\pgfsetstrokecolor{currentstroke}%
\pgfsetdash{{9.600000pt}{2.400000pt}{1.500000pt}{2.400000pt}}{0.000000pt}%
\pgfpathmoveto{\pgfqpoint{8.026594in}{0.347222in}}%
\pgfpathlineto{\pgfqpoint{8.304372in}{0.347222in}}%
\pgfusepath{stroke}%
\end{pgfscope}%
\begin{pgfscope}%
\definecolor{textcolor}{rgb}{0.000000,0.000000,0.000000}%
\pgfsetstrokecolor{textcolor}%
\pgfsetfillcolor{textcolor}%
\pgftext[x=8.415483in,y=0.298611in,left,base]{\color{textcolor}\fontsize{10.000000}{12.000000}\selectfont \textsc{h-clingo}   (10)}%
\end{pgfscope}%
\end{pgfpicture}%
\makeatother%
\endgroup%
}
			\caption{Solver configurations, with numbers of solved instances}
			\label{fig-cactus-a-star_pathfinding-legend}
		\end{subfigure}%
		\vspace{\floatsep}
		\begin{subfigure}{.32\textwidth}
			\centering
			\resizebox{\textwidth}{!}{\input{figures/cactus_a-star_pathfinding_g.pgf}}
			\caption{\rev{}{Number of guesses}}
			\label{fig-cactus-a-star_pathfinding-guesses}
		\end{subfigure}%
		\hspace*{\fill}
		\begin{subfigure}{.32\textwidth}
			\centering
			\resizebox{\textwidth}{!}{\input{figures/cactus_a-star_pathfinding_walltime.pgf}}
			\caption{\rev{Accumulated time }{Time} consumption}
			\label{fig-cactus-a-star_pathfinding-time}
		\end{subfigure}%
		\hspace*{\fill}
		\begin{subfigure}{.32\textwidth}
			\centering
			\resizebox{\textwidth}{!}{\input{figures/cactus_a-star_pathfinding_memory.pgf}}
			\caption{Memory consumption}
			\label{fig-cactus-a-star_pathfinding-memory}
		\end{subfigure}
		\caption{\rev{Time and memory }{Resource} consumption for solving each Pathfinding instance with A*}
		\label{fig-experimental-results-a-star_pathfinding}
	\end{figure}

	\begin{figure}[t]
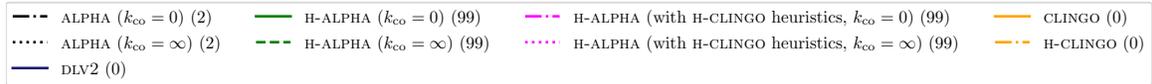
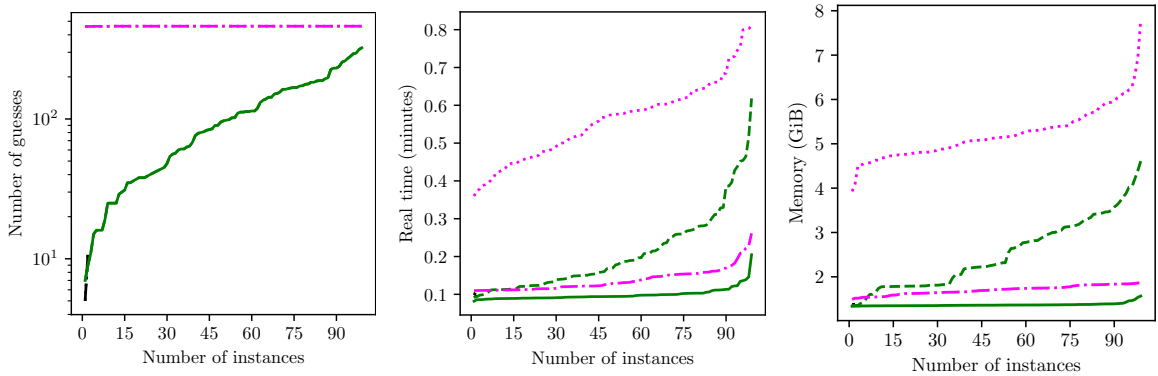

		\centering
		\begin{subfigure}{\textwidth}
			\centering
			\resizebox{\textwidth}{!}{
\begingroup%
\makeatletter%
\begin{pgfpicture}%
\pgfpathrectangle{\pgfpointorigin}{\pgfqpoint{9.220881in}{0.666667in}}%
\pgfusepath{use as bounding box, clip}%
\begin{pgfscope}%
\pgfsetbuttcap%
\pgfsetmiterjoin%
\definecolor{currentfill}{rgb}{1.000000,1.000000,1.000000}%
\pgfsetfillcolor{currentfill}%
\pgfsetlinewidth{0.000000pt}%
\definecolor{currentstroke}{rgb}{1.000000,1.000000,1.000000}%
\pgfsetstrokecolor{currentstroke}%
\pgfsetdash{}{0pt}%
\pgfpathmoveto{\pgfqpoint{0.000000in}{0.000000in}}%
\pgfpathlineto{\pgfqpoint{9.220881in}{0.000000in}}%
\pgfpathlineto{\pgfqpoint{9.220881in}{0.666667in}}%
\pgfpathlineto{\pgfqpoint{0.000000in}{0.666667in}}%
\pgfpathclose%
\pgfusepath{fill}%
\end{pgfscope}%
\begin{pgfscope}%
\pgfsetbuttcap%
\pgfsetmiterjoin%
\definecolor{currentfill}{rgb}{1.000000,1.000000,1.000000}%
\pgfsetfillcolor{currentfill}%
\pgfsetfillopacity{0.800000}%
\pgfsetlinewidth{1.003750pt}%
\definecolor{currentstroke}{rgb}{0.800000,0.800000,0.800000}%
\pgfsetstrokecolor{currentstroke}%
\pgfsetstrokeopacity{0.800000}%
\pgfsetdash{}{0pt}%
\pgfpathmoveto{\pgfqpoint{0.027778in}{0.000000in}}%
\pgfpathlineto{\pgfqpoint{9.193104in}{0.000000in}}%
\pgfpathquadraticcurveto{\pgfqpoint{9.220881in}{0.000000in}}{\pgfqpoint{9.220881in}{0.027778in}}%
\pgfpathlineto{\pgfqpoint{9.220881in}{0.638889in}}%
\pgfpathquadraticcurveto{\pgfqpoint{9.220881in}{0.666667in}}{\pgfqpoint{9.193104in}{0.666667in}}%
\pgfpathlineto{\pgfqpoint{0.027778in}{0.666667in}}%
\pgfpathquadraticcurveto{\pgfqpoint{0.000000in}{0.666667in}}{\pgfqpoint{0.000000in}{0.638889in}}%
\pgfpathlineto{\pgfqpoint{0.000000in}{0.027778in}}%
\pgfpathquadraticcurveto{\pgfqpoint{0.000000in}{0.000000in}}{\pgfqpoint{0.027778in}{0.000000in}}%
\pgfpathclose%
\pgfusepath{stroke,fill}%
\end{pgfscope}%
\begin{pgfscope}%
\pgfsetbuttcap%
\pgfsetroundjoin%
\pgfsetlinewidth{1.505625pt}%
\definecolor{currentstroke}{rgb}{0.000000,0.000000,0.000000}%
\pgfsetstrokecolor{currentstroke}%
\pgfsetdash{{9.600000pt}{2.400000pt}{1.500000pt}{2.400000pt}}{0.000000pt}%
\pgfpathmoveto{\pgfqpoint{0.055556in}{0.555556in}}%
\pgfpathlineto{\pgfqpoint{0.333333in}{0.555556in}}%
\pgfusepath{stroke}%
\end{pgfscope}%
\begin{pgfscope}%
\definecolor{textcolor}{rgb}{0.000000,0.000000,0.000000}%
\pgfsetstrokecolor{textcolor}%
\pgfsetfillcolor{textcolor}%
\pgftext[x=0.444444in,y=0.506944in,left,base]{\color{textcolor}\fontsize{10.000000}{12.000000}\selectfont \textsc{alpha} (\(\displaystyle k_\mathrm{co}=0\))   (2)}%
\end{pgfscope}%
\begin{pgfscope}%
\pgfsetbuttcap%
\pgfsetroundjoin%
\pgfsetlinewidth{1.505625pt}%
\definecolor{currentstroke}{rgb}{0.000000,0.000000,0.000000}%
\pgfsetstrokecolor{currentstroke}%
\pgfsetdash{{1.500000pt}{2.475000pt}}{0.000000pt}%
\pgfpathmoveto{\pgfqpoint{0.055556in}{0.347222in}}%
\pgfpathlineto{\pgfqpoint{0.333333in}{0.347222in}}%
\pgfusepath{stroke}%
\end{pgfscope}%
\begin{pgfscope}%
\definecolor{textcolor}{rgb}{0.000000,0.000000,0.000000}%
\pgfsetstrokecolor{textcolor}%
\pgfsetfillcolor{textcolor}%
\pgftext[x=0.444444in,y=0.298611in,left,base]{\color{textcolor}\fontsize{10.000000}{12.000000}\selectfont \textsc{alpha} (\(\displaystyle k_\mathrm{co}=\infty\))   (2)}%
\end{pgfscope}%
\begin{pgfscope}%
\pgfsetrectcap%
\pgfsetroundjoin%
\pgfsetlinewidth{1.505625pt}%
\definecolor{currentstroke}{rgb}{0.098039,0.098039,0.439216}%
\pgfsetstrokecolor{currentstroke}%
\pgfsetdash{}{0pt}%
\pgfpathmoveto{\pgfqpoint{0.055556in}{0.138889in}}%
\pgfpathlineto{\pgfqpoint{0.333333in}{0.138889in}}%
\pgfusepath{stroke}%
\end{pgfscope}%
\begin{pgfscope}%
\definecolor{textcolor}{rgb}{0.000000,0.000000,0.000000}%
\pgfsetstrokecolor{textcolor}%
\pgfsetfillcolor{textcolor}%
\pgftext[x=0.444444in,y=0.090278in,left,base]{\color{textcolor}\fontsize{10.000000}{12.000000}\selectfont \textsc{dlv2}   (0)}%
\end{pgfscope}%
\begin{pgfscope}%
\pgfsetrectcap%
\pgfsetroundjoin%
\pgfsetlinewidth{1.505625pt}%
\definecolor{currentstroke}{rgb}{0.000000,0.501961,0.000000}%
\pgfsetstrokecolor{currentstroke}%
\pgfsetdash{}{0pt}%
\pgfpathmoveto{\pgfqpoint{2.012698in}{0.555556in}}%
\pgfpathlineto{\pgfqpoint{2.290476in}{0.555556in}}%
\pgfusepath{stroke}%
\end{pgfscope}%
\begin{pgfscope}%
\definecolor{textcolor}{rgb}{0.000000,0.000000,0.000000}%
\pgfsetstrokecolor{textcolor}%
\pgfsetfillcolor{textcolor}%
\pgftext[x=2.401587in,y=0.506944in,left,base]{\color{textcolor}\fontsize{10.000000}{12.000000}\selectfont \textsc{h-alpha} (\(\displaystyle k_\mathrm{co}=0\))   (99)}%
\end{pgfscope}%
\begin{pgfscope}%
\pgfsetbuttcap%
\pgfsetroundjoin%
\pgfsetlinewidth{1.505625pt}%
\definecolor{currentstroke}{rgb}{0.000000,0.501961,0.000000}%
\pgfsetstrokecolor{currentstroke}%
\pgfsetdash{{5.550000pt}{2.400000pt}}{0.000000pt}%
\pgfpathmoveto{\pgfqpoint{2.012698in}{0.347222in}}%
\pgfpathlineto{\pgfqpoint{2.290476in}{0.347222in}}%
\pgfusepath{stroke}%
\end{pgfscope}%
\begin{pgfscope}%
\definecolor{textcolor}{rgb}{0.000000,0.000000,0.000000}%
\pgfsetstrokecolor{textcolor}%
\pgfsetfillcolor{textcolor}%
\pgftext[x=2.401587in,y=0.298611in,left,base]{\color{textcolor}\fontsize{10.000000}{12.000000}\selectfont \textsc{h-alpha} (\(\displaystyle k_\mathrm{co}=\infty\))   (99)}%
\end{pgfscope}%
\begin{pgfscope}%
\pgfsetbuttcap%
\pgfsetroundjoin%
\pgfsetlinewidth{1.505625pt}%
\definecolor{currentstroke}{rgb}{1.000000,0.000000,1.000000}%
\pgfsetstrokecolor{currentstroke}%
\pgfsetdash{{9.600000pt}{2.400000pt}{1.500000pt}{2.400000pt}}{0.000000pt}%
\pgfpathmoveto{\pgfqpoint{4.176938in}{0.555556in}}%
\pgfpathlineto{\pgfqpoint{4.454716in}{0.555556in}}%
\pgfusepath{stroke}%
\end{pgfscope}%
\begin{pgfscope}%
\definecolor{textcolor}{rgb}{0.000000,0.000000,0.000000}%
\pgfsetstrokecolor{textcolor}%
\pgfsetfillcolor{textcolor}%
\pgftext[x=4.565827in,y=0.506944in,left,base]{\color{textcolor}\fontsize{10.000000}{12.000000}\selectfont \textsc{h-alpha} (with \textsc{h-clingo} heuristics, \(\displaystyle k_\mathrm{co}=0\))   (99)}%
\end{pgfscope}%
\begin{pgfscope}%
\pgfsetbuttcap%
\pgfsetroundjoin%
\pgfsetlinewidth{1.505625pt}%
\definecolor{currentstroke}{rgb}{1.000000,0.000000,1.000000}%
\pgfsetstrokecolor{currentstroke}%
\pgfsetdash{{1.500000pt}{2.475000pt}}{0.000000pt}%
\pgfpathmoveto{\pgfqpoint{4.176938in}{0.347222in}}%
\pgfpathlineto{\pgfqpoint{4.454716in}{0.347222in}}%
\pgfusepath{stroke}%
\end{pgfscope}%
\begin{pgfscope}%
\definecolor{textcolor}{rgb}{0.000000,0.000000,0.000000}%
\pgfsetstrokecolor{textcolor}%
\pgfsetfillcolor{textcolor}%
\pgftext[x=4.565827in,y=0.298611in,left,base]{\color{textcolor}\fontsize{10.000000}{12.000000}\selectfont \textsc{h-alpha} (with \textsc{h-clingo} heuristics, \(\displaystyle k_\mathrm{co}=\infty\))   (99)}%
\end{pgfscope}%
\begin{pgfscope}%
\pgfsetrectcap%
\pgfsetroundjoin%
\pgfsetlinewidth{1.505625pt}%
\definecolor{currentstroke}{rgb}{1.000000,0.647059,0.000000}%
\pgfsetstrokecolor{currentstroke}%
\pgfsetdash{}{0pt}%
\pgfpathmoveto{\pgfqpoint{7.957149in}{0.555556in}}%
\pgfpathlineto{\pgfqpoint{8.234927in}{0.555556in}}%
\pgfusepath{stroke}%
\end{pgfscope}%
\begin{pgfscope}%
\definecolor{textcolor}{rgb}{0.000000,0.000000,0.000000}%
\pgfsetstrokecolor{textcolor}%
\pgfsetfillcolor{textcolor}%
\pgftext[x=8.346038in,y=0.506944in,left,base]{\color{textcolor}\fontsize{10.000000}{12.000000}\selectfont \textsc{clingo}   (0)}%
\end{pgfscope}%
\begin{pgfscope}%
\pgfsetbuttcap%
\pgfsetroundjoin%
\pgfsetlinewidth{1.505625pt}%
\definecolor{currentstroke}{rgb}{1.000000,0.647059,0.000000}%
\pgfsetstrokecolor{currentstroke}%
\pgfsetdash{{9.600000pt}{2.400000pt}{1.500000pt}{2.400000pt}}{0.000000pt}%
\pgfpathmoveto{\pgfqpoint{7.957149in}{0.347222in}}%
\pgfpathlineto{\pgfqpoint{8.234927in}{0.347222in}}%
\pgfusepath{stroke}%
\end{pgfscope}%
\begin{pgfscope}%
\definecolor{textcolor}{rgb}{0.000000,0.000000,0.000000}%
\pgfsetstrokecolor{textcolor}%
\pgfsetfillcolor{textcolor}%
\pgftext[x=8.346038in,y=0.298611in,left,base]{\color{textcolor}\fontsize{10.000000}{12.000000}\selectfont \textsc{h-clingo}   (0)}%
\end{pgfscope}%
\end{pgfpicture}%
\makeatother%
\endgroup%
}
			\caption{Solver configurations, with numbers of solved instances}
			\label{fig-cactus-a-star_osmnx-legend}
		\end{subfigure}%
		\vspace{\floatsep}
		\begin{subfigure}{.32\textwidth}
			\centering
			\resizebox{\textwidth}{!}{\input{figures/cactus_a-star_osmnx_g.pgf}}
			\caption{\rev{}{Number of guesses}}
			\label{fig-cactus-a-star_osmnx-guesses}
		\end{subfigure}%
		\hspace*{\fill}
		\begin{subfigure}{.32\textwidth}
			\centering
			\resizebox{\textwidth}{!}{\input{figures/cactus_a-star_osmnx_walltime.pgf}}
			\caption{\rev{Accumulated time }{Time} consumption}
			\label{fig-cactus-a-star_osmnx-time}
		\end{subfigure}%
		\hspace*{\fill}
		\begin{subfigure}{.32\textwidth}
			\centering
			\resizebox{\textwidth}{!}{\input{figures/cactus_a-star_osmnx_memory.pgf}}
			\caption{Memory consumption}
			\label{fig-cactus-a-star_osmnx-memory}
		\end{subfigure}
		\caption{\rev{Time and memory }{Resource} consumption for solving each Routing instance with A*}
		\label{fig-experimental-results-a-star_osmnx}
	\end{figure}

Cactus plots for Pathfinding with A* (\cref{fig-experimental-results-a-star_pathfinding}) \rev{}{and Routing with A* (\cref{fig-experimental-results-a-star_osmnx})} were generated in the same way as for HRP (cf.\ \cref{sec-applications-hrp-results}).

\rev{Again }{In the case of Pathfinding (\cref{fig-experimental-results-a-star_pathfinding})}, \alphaslv\ with domain-specific heuristics (\slv{h-alpha}) solved the highest number of instances (\rev{63 }{72} out of 100).
Like with HRP, this performance could only be achieved when grounding constraints strictly ($k_\mathrm{co}=0$), while permissive grounding of constraints ($k_\mathrm{co}=\infty$) resulted in lower performance, solving only \rev{16 }{20} instances.
\rev{}{Performance of \alphaslv\ with \slv{h-clingo}-like heuristics was a bit lower.}

On the other extreme, \alphaslv\ without domain-specific heuristic\rev{}{s} could solve only \rev{5 }{13} of the 100 instances.

The systems \slv{dlv2}, \slv{clingo}, and \slv{h-clingo} performed somewhere in between those extremes.
\rev{With the exception of the smallest instance, \slv{clingo} and \slv{h-clingo} could only solve instances without a solution (i.e., the solver derived failure). }{%
	Of the 72 instances solved by \slv{h-alpha}, 18 are unsatisfiable, and of the 10 instances solved by \slv{h-clingo}, 9 are unsatisfiable.
}%
The largest instance solved by \slv{h-alpha} had a width of \rev{315 }{400}, and
the largest instance for which \slv{h-clingo} could derive failure had a width of 470.
\rev{}{Note that our heuristics are designed to find solutions, not to efficiently prove unsatisfiability.}

\rev{}{
	In the Routing domain (\cref{fig-experimental-results-a-star_osmnx}), \alphaslv\ with domain-specific heuristics (\slv{h-alpha}) solved 99 of the 100 instances, while \alphaslv\ without domain-specific heuristics could solve only two instances, and the other systems solved no instances at all.
	\alphaslv\ with \slv{h-clingo}-like heuristics solved the same number of instances, but consumed more time and memory in doing so.
}

\rev{}{
	For each of the 8 Pathfinding instances solved by \slv{clingo}, the system spent between 88\% and 100\% of total solving time in grounding, on average 95\%.
	When using domain-specific heuristics, \slv{h-clingo} solved 10 Pathfinding instances and spent between 96\% and 100\% in grounding, on average more than 99\% of overall time.
	\slv{dlv2} solved 8 instances and spent between 15\% and 55\% of total solving time in grounding, on average 43\%.
	Neither \slv{clingo} nor \slv{h-clingo} nor \slv{dlv2} finished grounding for any of the Routing instances.
}

\rev{}{
	Note some peculiarities in \cref{fig-experimental-results-a-star_pathfinding,fig-experimental-results-a-star_osmnx}:
	In \cref{fig-cactus-a-star_pathfinding-guesses}, there is a vertical line for \slv{clingo} and \slv{dlv2}.
	This is because for seven instances, these two systems were able to derive failure without doing any guesses.
	\Cref{fig-cactus-a-star_osmnx-guesses}, on the other hand, shows a horizontal line for \slv{h-alpha} with \slv{h-clingo} heuristics.
	This is because these heuristics do not represent A* correctly and try to visit all 461 non-start nodes in almost every instance.
	In \cref{fig-cactus-a-star_osmnx-guesses}, all lines for $k_\mathrm{co}=\infty$ are hidden by their respective $k_\mathrm{co}=0$ counterparts because numbers of guesses coincide.
}

\subsection{Discussion}

Our results show that we have extended the application area of ASP.
By combining our novel approach to domain-specific heuristics with lazy-grounding answer set solving, we could solve large-scale problem instances that are out of reach for conventional ASP systems.
This finding supports our initial hypothesis that both lazy grounding and domain-specific heuristics are crucial for solving large-scale industrial problems.

Our approach extends the one by \citeA{DBLP:conf/aaai/GebserKROSW13}, the first extension of ASP's input language by a declarative framework for domain-specific heuristics.
\citeA{DBLP:conf/aaai/GebserKROSW13} have provided the first experimental evidence that such an approach can improve ASP solving performance and also reported grounding issues with some instances.

Our advancement consists of novel syntax and semantics for heuristic directives that make it possible to reason about the current partial assignment, facilitating heuristics based on what has or has \emph{not yet} been decided by the solver.
Although the approach by \citeA{DBLP:conf/aaai/GebserKROSW13} has worked very well on planning problems, it seems that more flexibility in the definition of heuristics, supported by the novel features of our approach, is necessary to represent heuristics for other kinds of problems.

Our results undeniably show that domain-specific heuristics improve solving performance for the domains under consideration.
This is not only true for \alphaslv\ but also for \slv{clingo}.
However, domain-specific heuristics usually increase \slv{clingo}'s memory consumption, thus exacerbating the grounding bottleneck from which ground-and-solve systems such as \slv{clingo} are suffering.
Domain-specific heuristics for \slv{dlv2} were out of scope because \slv{dlv2} does not support the declarative specification of heuristics.

\rev{
Solution quality is another aspect to keep in mind.
Since domain-specific heuristics lead the solver towards \enquote{better} solutions, the quality of answer sets computed by \slv{h-alpha} or \slv{h-clingo} is higher than the quality of answer sets computed by \alphaslv, \slv{clingo}, or \slv{dlv2}.
In the case of A*\!, the encoding even enforces optimal solutions; domain-specific heuristics help the solver find an optimal solution more quickly.
}{}

However, we do not claim that heuristics based on partial assignments are always beneficial.
Our findings cannot reject the possibility that \slv{h-clingo} might outperform \slv{h-alpha} when other encodings or other heuristics are used since there might be encoding optimisations that we have not thought of.
\rev{}{We even observed this situation with most competition instances of the Partner Units Problem in our experiments.}
Still, we are confident that our approach's novel features make the specification of practical heuristics more intuitive and effortless.
\rev{}{Furthermore, in our experiments, \alphaslv\ usually performed worse when employing \slv{clingo}-like heuristics without our novel features.}

\rev{\slv{clingo}-like heuristics can be approximated in \alphaslv\ by replacing \enquote{\nafsymbol} by \enquote{$\heusignF$}.
Cursory experiments with such encodings suggest that, due to the lack of heuristic conditions exploiting negation as failure to avoid conflicting assignments, \alphaslv\ produces many backtracks and uses much more time and space than with heuristics under our semantics.}{}

Results for HRP (\cref{fig-experimental-results-house}) and \rev{Pathfinding with }{}A* \rev{(\cref{fig-experimental-results-a-star_pathfinding}) }{(\cref{fig-experimental-results-a-star_pathfinding,fig-experimental-results-a-star_osmnx})} indicate that permissive grounding \cite<cf.>{DBLP:conf/lpnmr/TaupeWF19}, i.e., providing the solver with more nogoods representing ground constraints than necessary, can be counterproductive when domain-specific heuristics are used.
We conjecture the reason for this to be that suitable domain-specific heuristics can assist the solver even better than additional constraints while avoiding the overhead of additional nogoods (in terms of space consumption and propagation efforts).
This assumption is supported by the considerable increase in \alphaslv's memory consumption when grounding constraints permissively in those domains.

Closer investigation of this issue revealed that in both \rev{domains,}{} HRP and \rev{Pathfinding with }{}A*, one single constraint was the source of the performance deterioration.
In HRP, the full and very large grounding for one specific constraint was produced before the solver had even made a single choice. In \rev{Pathfinding with }{}A*\!, many ground nogoods were produced for one constraint not necessary to find the optimal solution when domain-specific heuristics are used.

\rev{}{
	The Partner Units Problem (PUP) proved more challenging for our approach than the other domains under consideration.
	While the combination of lazy grounding and domain-specific heuristics was able to solve large instances that are out of reach of ground-and-solve systems, hard competition instances can be solved by ground-and-solve systems even without domain-specific heuristics.
	In these cases, domain-specific heuristics in \alphaslv\ did more harm than good, because they caused a significant overhead and they are obviously not effective when the full grounding and efficient solving techniques are available.
	However, finding efficient encodings or heuristics for PUP is out of scope of this paper.
	Future work should investigate whether another combination of encoding and heuristic directives can help to solve instances that are both large and hard to solve.
}

To sum up, domain-specific heuristics implemented in our novel framework, combined with strict lazy grounding by \alphaslv, outperformed all other tested systems when applied to large instances of the House Reconfiguration Problem, the Partner Units Problem, and \rev{Pathfinding with }{}A* \rev{}{used with two different search problems}.
Applications to other domains should be easy to put into practice and belong to future work.

\section{Conclusions and Future Work}
\label{sec-conclusion}

We have proposed novel syntax and semantics for declarative domain-specific heuristics in ASP that can depend non-monotonically on the partial assignment maintained during solving.
Furthermore, we have demonstrated how to integrate our approach in a lazy-grounding ASP system and presented experimental results obtained with the lazy-grounding solver \alphaslv.

Our semantics has proven beneficial for several practical application domains,
advancing the work by \citeA{DBLP:conf/aaai/GebserKROSW13}.
In experiments, our implementation exhibited convincing time and memory consumption behaviour.
Thus, we extended the application area of ASP by solving large problem instances that conventional ASP systems could not solve.

Our approach's suitability to implement other practice-oriented heuristics should be assessed by the community.
Some real-world domain-specific heuristics will require extensions of our approach, such as by supporting randomness and restarts.
Furthermore, adopting ideas like \texttt{init} and \texttt{factor} modifiers from \slv{clingo}, and investigating the special role of aggregates in heuristic conditions should be addressed in future work.
An adaption to ground-and-solve systems like \slv{clingo} \cite{DBLP:journals/tplp/GebserKKS19} or \slv{dlv2} \cite{DBLP:conf/lpnmr/AlvianoCDFLPRVZ17} should be investigated, also addressing the question of how heuristics interact with non-head-cycle-free disjunction.

Thinking more broadly, the question of how to generate domain-specific heuristics automatically is of great importance since, currently, such heuristics have to be invented by humans familiar with the domain (and partly also with solving technology).

\acks{%
	This work has been conducted in the scope of the research project \textit{DynaCon (FFG-PNr.:\ 861263)}, which was funded by the \rev{Austrian Federal Ministry of Transport, Innovation and Technology (BMVIT) }{Austrian Federal Ministry for Climate Action, Environment, Energy, Mobility, Innovation and Technology (BMK)} under the program \enquote{ICT of the Future} \rev{}{(via the project \emph{DynaCon}, FFG-PNr.:\ 861263)} between 2017 and 2020.\footnote{See \url{https://iktderzukunft.at/en/} for more information.}
	This research was also supported by the Academy of Finland, project 251170, and by EU ECSEL Joint Undertaking under grant agreement no.\ 737459 (project Productive4.0).
	
	We are thankful to Peter Schüller for his contributions to our earlier conference paper on this topic, \rev{}{Martin Gebser for proofreading this article,} Andreas Falkner for his comments on an earlier version of this paper, Stephen Cummings for language editing, and the anonymous reviewers of ICLP 2019 \rev{}{and the Journal of Artificial Intelligence Research} for their helpful reviews.
}

\appendix
\rev{}{
\section{Proofs}
\label{sec-proofs}

\begin{proof}[Proof sketch for \cref{lemma:transformation2}]
	Since heads of heuristic directives are unaffected by the transformation, it suffices to show that $\cond{\heudir}$ is satisfied iff $\exists \heudir' \in \transform{}(\heudir)$ s.t.\ $\cond{\heudir'}$ is satisfied (cf.\ \cref{def-condition-satisfied,def-directive-applicable}).
	Furthermore, it suffices to consider the conditions under which the affected heuristic atoms are satisfied (cf.\ \cref{def-heuat-satisfied}).
	
	Case \eqrefformat{\ref{eq:transform:else}} is trivial.
	
	For \eqrefformat{\ref{eq:transform:1}} and \eqrefformat{\ref{eq:transform:4}}, $\transform{}(\heudir)$ maps $\heudir$ to two new directives, i.e., $\transform{}(\heudir) = \{ \heudir', \heudir'' \}$ and $\heudir' \neq \heudir''$.
	Let $\heusign~a$ be the heuristic atom in $\cond{\heudir}$ affected by the transformation.
	In both cases, \eqrefformat{\ref{eq:transform:1}} and \eqrefformat{\ref{eq:transform:4}}, $\heusign~a$ is removed from $\cond{\heudir}$, one atom is added to (the positive or negative part of) $\cond{\heudir'}$, and one atom is added to (the positive or negative part of) $\cond{\heudir''}$.
	Let us call these new atoms $\heusign'~a$ and $\heusign''~a$.
	
	For \eqrefformat{\ref{eq:transform:1}}, all atoms mentioned occur in the positive part of the condition, i.e., $\heusign~a \in \condpos{\heudir}$, $\heusign'~a \in \condpos{\heudir'}$, and $\heusign''~a \in \condpos{\heudir''}$.
	So we now have to show that $\heusign~a$ is satisfied iff $\heusign'~a$ is satisfied or $\heusign''~a$ is satisfied.
	For example, if $\heusign = \heusignFM$, then $\heusign' = \heusignM$ and $\heusign'' = \heusignF$.
	Since a heuristic atom $\heusign~a$ is satisfied w.r.t.\ an assignment $\assignment$ iff $\assignedtruthof{a} \in \heusign$ (cf.\ \cref{def-heuat-satisfied}), it is obvious that $\heusignFM~a$ is satisfied iff $\heusignM~a$ is satisfied or $\heusignF~a$ is satisfied.
	
	The other cases can be shown in a similar way.
	\end{proof}

\begin{proof}[Proof of \cref{lemma-heu-iff-atom}]

    ``$\Rightarrow$'': Let $\assignedtruthof{h} = \sigT$, resp.~$\assignedtruthof{h} = \sigM$, and note that \alphaslv{} is only guessing the truth value of an atom if that atom represents an applicable rule, i.e., it is of the form $\beta(r,\sigma)$ (cf.~\cref{sec-domspec-lazygrounding-alpha} and \citeA{DBLP:conf/lpnmr/Weinzierl17}). Since $h$ is not of such a form, \alphaslv{} therefore only assigns $h$ to $\sigT$, respectively $\sigM$, if it is propagated.
    Observe that all nogoods from \eqref{nogood-example1}-\eqref{nogood-example-last} are such that the heuristic atom only appears with negative polarity in it, i.e., $h$ appears as $\sigF h \in ng$. Furthermore, since there are no other nogoods that contain $h$, the only way for $\assignedtruthof{h}=\sigT$, respectively $\assignedtruthof{h}=\sigM$, to hold is by $ng$ being unit in $\assignment' = \assignment \setminus \{\sigT h, \sigM h\}$.

    Notice that the polarity of $h$ is negative in all nogoods. Since \alphaslv{} does not guess on $h$, it is impossible for $ng$ to be involved in a conflict, as $h$ is unassigned until $ng$ is unit and then $h$ will be assigned such that $ng$ is not violated and by that all other nogoods containing $h$ can no longer become violated. This implies that $ng$ also is never directly involved in the mechanics of conflict-driven learning, hence learning will not create additional nogoods that contain $h$.

    ``$\Leftarrow$'': Let one nogood $ng$ of the form \eqref{nogood-example1}-\eqref{nogood-example-last} be unit w.r.t.\ $\assignment' = \assignment \setminus \{\sigT h, \sigM h\}$, then unit propagation on $ng$ will lead to an extended assignment where $h$ is assigned $\sigT$, respectively $\sigM$, i.e., $\prop(ng,\assignment') = \{\sigT, \sigM\}$, respectively $\prop(ng,\assignment') = \{\sigM\}$, and the resulting assignment $\assignment$ is such that $\assignedtruthof{h}=\sigT$, respectively $\assignedtruthof{h}=\sigM$, holds.
  
\end{proof}

\begin{proof}[Proof of \cref{lemma-solving-satisfied-iff-cond}]
    ``$\Rightarrow$'': Let $\cond{\heudir}$ be satisfied w.r.t.~$\assignment$ and $\prog$, i.e., every $ha \in \condpos{\heudir}$ is satisfied and no $ha \in \condneg{\heudir}$ is satisfied. We have to show that $\heudir$ is solving-satisfied w.r.t.~$\assignment^{\Delta_\prog}$, i.e., all of the following hold:
    \begin{enumerate}[label=(\arabic*)]
    \item $\assignedtruthof[\assignment^{\Delta_\prog}]{\HeuOnrm{T}(\heudir)} = \sigT$,
    \item $\assignedtruthof[\assignment^{\Delta_\prog}]{\HeuOnrm{MT}(\heudir)} \in \{ \sigM, \sigT \}$,
    \item $\assignedtruthof[\assignment^{\Delta_\prog}]{\HeuOnrm{F}(\heudir)} = \sigT$,
    \item $\assignedtruthof[\assignment^{\Delta_\prog}]{\HeuOffrm{T}(\heudir)} \neq \sigT$,
    \item $\assignedtruthof[\assignment^{\Delta_\prog}]{\HeuOffrm{MT}(\heudir)} \notin \{ \sigM, \sigT \}$, and
    \item $\assignedtruthof[\assignment^{\Delta_\prog}]{\HeuOffrm{F}(\heudir)} \neq \sigT$.
    \end{enumerate}

    \noindent
    \begin{enumerate*}[label=(\arabic*)]
    \item \label{proof:lem-2-case1} Since every $ha \in \condpos{\heudir}$ is satisfied, it holds by \cref{def-heuat-satisfied} that $\assignedtruthof{\fheuat(ha)}$ $\in$ $\fsignset(ha)$. This specifically implies for every $a \in \fheuat(\filterbysigns{\condpos{\heudir}}{\heusignT})$ that $\assignedtruthof{a} = \sigT$. Hence, the nogood $g$ of the form \eqref{nogood-example1} is unit w.r.t.~$\assignment$, i.e., $\sigT~\HeuOnrm{T}(\heudir) \in \prop(g,\assignment)$ and consequently $\assignedtruthof[\assignment^{\Delta_\prog}]{\HeuOnrm{T}(\heudir)} = \sigT$.\\
    \item The proof is analogeous to \ref{proof:lem-2-case1} relying on the nogood of the form \eqref{nogood-generated-2} to show that $\assignedtruthof[\assignment^{\Delta_\prog}]{\HeuOnrm{MT}(\heudir)} \in \{ \sigM, \sigT \}$.\\
    \item The proof is analogeous to \ref{proof:lem-2-case1} relying on the nogood of the form \eqref{nogood-generated-3} to show that $\assignedtruthof[\assignment^{\Delta_\prog}]{\HeuOnrm{F}(\heudir)} = \sigT$.\\
    \item \label{proof:lem-2-case2} Since no $ha \in \condneg{\heudir}$ is satisfied w.r.t.~$\assignment$, it holds for every $\sigT a \in \condneg{\heudir}$ that $\assignedtruthof{a} \neq \sigT$. Consequently, no nogood of the form \eqref{nogood-example2} is unit w.r.t.~$\assignment$. From Lemma~\ref{lemma-heu-iff-atom} it then follows that $\assignedtruthof[\assignment^{\Delta_\prog}]{\HeuOffrm{T}(\heudir)} \neq \sigT$.\\
    \item The proof is analogeous to \ref{proof:lem-2-case2} relying on the nogood(s) of the form \eqref{nogood-generated-5} and Lemma~\ref{lemma-heu-iff-atom} to show that $\assignedtruthof[\assignment^{\Delta_\prog}]{\HeuOffrm{MT}(\heudir)} \notin \{ \sigM, \sigT \}$.
      \\
    \item The proof is analogeous to \ref{proof:lem-2-case2} relying on the nogood(s) of the form \eqref{nogood-example-last} and Lemma~\ref{lemma-heu-iff-atom} to show that $\assignedtruthof[\assignment^{\Delta_\prog}]{\HeuOffrm{F}(\heudir)} \neq \sigT$.
    \end{enumerate*}
    
    From the above follows that $\heudir$ is solving-satisfied w.r.t.~$\assignment^{\Delta_\prog}$.

    ``$\Leftarrow$'': Let $\heudir$ be solving-satisfied w.r.t.~$\assignment^{\Delta_\prog}$, i.e.,
    \begin{enumerate*}[label=(\arabic*)]
    \item $\assignedtruthof[\assignment^{\Delta_\prog}]{\HeuOnrm{T}(\heudir)} = \sigT$,\\
    \item $\assignedtruthof[\assignment^{\Delta_\prog}]{\HeuOnrm{MT}(\heudir)} \in \{ \sigM, \sigT \}$,
    \item $\assignedtruthof[\assignment^{\Delta_\prog}]{\HeuOnrm{F}(\heudir)} = \sigT$,\\
    \item $\assignedtruthof[\assignment^{\Delta_\prog}]{\HeuOffrm{T}(\heudir)} \neq \sigT$,
    \item $\assignedtruthof[\assignment^{\Delta_\prog}]{\HeuOffrm{MT}(\heudir)} \notin \{ \sigM, \sigT \}$, and\\
    \item $\assignedtruthof[\assignment^{\Delta_\prog}]{\HeuOffrm{F}(\heudir)} \neq \sigT$.
    \end{enumerate*}
    From Lemma~\ref{lemma-heu-iff-atom} and $\assignedtruthof[\assignment^{\Delta_\prog}]{\HeuOnrm{T}(\heudir)} = \sigT$ it directly follows that the nogood of the form \eqref{nogood-example1} is unit w.r.t.~$\assignment$.
    Since $\assignment$ is deductively consistent with $\prog$ it holds for every $a \in \fheuat(\filterbysigns{\condpos{\heudir}}{\heusignT})$ that $\assignedtruthof{a} = \sigT$ and consequently, that every $ha \in \filterbysigns{\condpos{\heudir}}{\heusignT}$ is satisfied w.r.t.~$\assignment$.
    By an analogeous argument it also follows that that every $ha \in \filterbysigns{\condpos{\heudir}}{\heusignM\heusignT}$ is satisfied w.r.t.~$\assignment$ and that every $ha \in \filterbysigns{\condpos{\heudir}}{\heusignF}$ is satisfied w.r.t.~$\assignment$.

    Likewise, from $\assignedtruthof[\assignment^{\Delta_\prog}]{\HeuOffrm{T}(\heudir)} \neq \sigT$, Lemma~\ref{lemma-heu-iff-atom}, and every nogood $g$ of the form \eqref{nogood-example2} it also follows that $\heusignT a \in \condneg{\heudir}$ is not assigned true w.r.t.~$\assignment$, i.e., $\assignedtruthof{a} \neq \sigT$. By \cref{def-heuat-satisfied} it therefore follows that no $a \in \fheuat(\filterbysigns{\condneg{\heudir}}{\heusignT})$ is satisfied w.r.t.~$\assignment$. By analogeous arguments it also follows that no $a \in \fheuat(\filterbysigns{\condneg{\heudir}}{\heusignM\heusignT})$ and no $a \in \fheuat(\filterbysigns{\condneg{\heudir}}{\heusignF})$ is satisfied w.r.t.~$\assignment$, i.e., no $ha \in \fheuat(\condneg{\heudir})$ is satisfied w.r.t.~$\assignment$.
    
    Since every $ha \in \condpos{\heudir}$ is satisfied w.r.t.~$\assignment$ and no $ha \in \condneg{\heudir}$ is satisfied w.r.t.~$\assignment$, it follows that $\cond{\heudir}$ is satisfied w.r.t.~$\assignment$.
    
\end{proof}

}

\vskip 0.2in
\bibliography{jair}
\bibliographystyle{theapa}

\end{document}